\documentclass[twocolumn]{article}

\usepackage{template/arxiv_style}

\usepackage[utf8]{inputenc} % allow utf-8 input
\usepackage[T1]{fontenc}    % use 8-bit T1 fontss
\usepackage{url}            % simple URL typesetting
\usepackage{booktabs}       % professional-quality tables
\usepackage{amsfonts}       % blackboard math symbols
\usepackage{nicefrac}       % compact symbols for 1/2, etc.
\usepackage{microtype}      % microtypography
\usepackage{xcolor}         % colors
\usepackage{hyperref}
\usepackage[compress]{natbib}
\usepackage{url}
\usepackage{amsfonts}       % blackboard math symbols
\usepackage{amsmath}        % for math environments
\usepackage{amssymb}        % for math symbols
\usepackage{mathtools}
\usepackage{amsthm}
\usepackage{dsfont}
\usepackage{graphicx}       % for including graphics
\usepackage{bm}             % proper bold notation in math mode
\usepackage{url}            % simple URL typesetting
\usepackage{nicefrac}       % compact symbols for 1/2, etc.
\usepackage{microtype}      % microtypography
\usepackage[symbol]{footmisc}
\usepackage{float}
\usepackage{wrapfig}
\usepackage{algorithm}
\usepackage{algorithmic}
\usepackage{array}          % to make \newcolumntype work
\usepackage{enumitem}       % can remove spaces from itemize
\usepackage{authblk}        % custom author list
\usepackage{adjustbox}
\usepackage{subcaption}
\usepackage[capitalise]{cleveref}
\usepackage{breakurl}
\usepackage{longtable}
\usepackage{lscape}
\usepackage{setspace}
\usepackage[htt]{hyphenat}

\usepackage{svg}
\usepackage{multirow}

\definecolor{cgreen}{RGB}{0, 158, 115}
\definecolor{cred}{RGB}{213, 94, 0}
\usepackage{pifont}
\newcommand{\cmark}{\textcolor{cgreen}{\ding{51}}}%
\newcommand{\xmark}{\textcolor{cred}{\ding{55}}}%

\theoremstyle{plain}
\newtheorem{theorem}{Theorem}[section]
\crefname{theorem}{Thm.}{T}
\newtheorem{proposition}[theorem]{Proposition}
\newtheorem{lemma}[theorem]{Lemma}
\newtheoremstyle{property}
  {\topsep}
  {-1pt}
  {\itshape}
  {}
  {\textsc}
  {}
  {.5em}
  {\textsc{\thmname{#1}\thmnumber{ #2}\thmnote{ (#3)}}}
\theoremstyle{property}
\newtheorem{property}{Property}

\newtheoremstyle{assumption} % Define new style
  {6pt}  % Space above
  {-2pt}  % Space below
  {\itshape}     % Body font (default)
  {}     % Indentation
  {\textsc} % Theorem head font (bold)
  {}    % Punctuation after theorem head
  {.5em} % Space between head and body
  {\textsc{\thmname{#1}\thmnumber{ #2}\thmnote{ (#3)}}}     % Theorem head spec

\theoremstyle{assumption}  % Apply assumption style
\newtheorem{assumption}{Assumption}

\crefname{property}{Prop.}{P}
\crefname{assumption}{Asm.}{A}
\newtheoremstyle{sketch}
  {\topsep}
  {\topsep}
  {}
  {}
  {\textsc}
  {}
  {.5em}
  {\textit{\thmname{#1}}}
\theoremstyle{sketch}

\DeclareMathOperator*{\codom}{\text{codom}}

\makeatletter
\renewcommand{\paragraph}{%
  \@startsection{paragraph}{4}%
  {\z@}{.2ex \@plus 1ex \@minus .2ex}{-1em}%
  {\normalfont\normalsize\bfseries}%
}
\makeatother

\AddToHook{cmd/appendix/before}{\crefalias{section}{appendix}}

\title{\textbf{Formally Exploring\\Time-Series Anomaly Detection Evaluation Metrics}}

\author[1]{\normalsize Dennis~Wagner}
\author[1]{\normalsize Arjun~Nair}
\author[1]{\normalsize Billy~Joe~Franks}
\author[2]{\normalsize Justus~Arweiler}
\author[3]{\normalsize Aparna~Muraleedharan}
\author[2]{\normalsize Indra~Jungjohann}
\author[4]{\normalsize Fabian~Hartung}
\author[1]{\normalsize Mayank~C.~Ahuja}
\author[1]{\normalsize Andriy~Balinskyy}
\author[1]{\normalsize Saurabh~Varshneya}
\author[1]{\normalsize Nabeel~Hussain~Syed}
\author[1]{\normalsize Mayank~Nagda}
\author[1]{\normalsize Phillip~Liznerski}
\author[1]{\normalsize Steffen~Reithermann}
\author[5]{\normalsize Maja~Rudolph}
\author[1,6]{\normalsize Sebastian~Vollmer}
\author[7]{\normalsize Ralf~Schulz}
\author[4]{\normalsize Torsten~Katz}
\author[8]{\normalsize Stephan~Mandt}
\author[9]{\normalsize Michael~Bortz}
\author[10]{\normalsize Heike~Leitte}
\author[11]{\normalsize Daniel~Neider}
\author[3]{\normalsize Jakob~Burger}
\author[2]{\normalsize Fabian~Jirasek}
\author[2]{\normalsize Hans~Hasse}
\author[1]{\normalsize Sophie~Fellenz}
\author[1]{\normalsize Marius~Kloft}
\affil[1]{\normalsize Department of Machine Learning, University of Kaiserslautern-Landau, Kaiserslautern, Germany}
\affil[2]{\normalsize Laboratory of Engineering Thermodynamics, University of Kaiserslautern-Landau, Kaiserslautern, Germany}
\affil[3]{\normalsize Department of Chemical Process Engineering, Technical University of Munich, Straubing, Germany}
\affil[4]{\normalsize BASF SE, Ludwigshafen am Rhein, Germany}
\affil[5]{\normalsize Data Science Institute, University of Wisconsin-Madison, Madison, USA}
\affil[6]{\normalsize German Research Center for Artificial Intelligence (DFKI), Kaiserslautern, Germany}
\affil[7]{\normalsize Department of Natural and Environmental Science, University of Kaiserslautern-Landau, Landau, Germany}
\affil[8]{\normalsize Department of Computer Science, University of California, Irvine, USA}
\affil[9]{\normalsize Department of Optimization, Fraunhofer Institute for Industrial Mathematics, Kaiserslautern, Germany}
\affil[10]{\normalsize Department of Computer Science, University of Kaiserslautern-Landau, Kaiserslautern, Germany}
\affil[11]{\normalsize University Alliance Ruhr, TU Dortmund University, Germany}

\date{}

\begin{document}

\maketitle

\begin{abstract}
Undetected anomalies in time series can trigger catastrophic failures in safety-critical systems, such as chemical plant explosions or power grid outages. Although many detection methods have been proposed, their performance remains unclear because current metrics capture only narrow aspects of the task and often yield misleading results. We address this issue by introducing verifiable properties that formalize essential requirements for evaluating time-series anomaly detection. These properties enable a theoretical framework that supports principled evaluations and reliable comparisons. Analyzing 37 widely used metrics, we show that most satisfy only a few properties, and none satisfy all, explaining persistent inconsistencies in prior results. To close this gap, we propose LARM, a flexible metric that provably satisfies all properties, and extend it to ALARM, an advanced variant meeting stricter requirements.
\end{abstract}

\section{Introduction}\label{sec:introduction}
Many modern applications require the processing of data while it is generated. This includes monitoring patients, servers, or industrial production processes \citep{zamanzadeh2024deep}.
We call data collected at regular intervals from one or more sensors \emph{time series}.
A fundamental task on time-series data is the \emph{detection of anomalies (TSAD)} -- rare yet significant, temporally bounded deviations from the norm \citep{ruff2021unifying}.
TSAD plays a critical role in preventing catastrophic failures, increasing productivity, reducing losses, or ensuring safety. 
Therefore, the performance of different TSAD methods needs to be thoroughly evaluated.
\begin{figure}[t]
    \begin{center}
       \centerline{\includesvg[inkscapelatex=false,width=\linewidth]{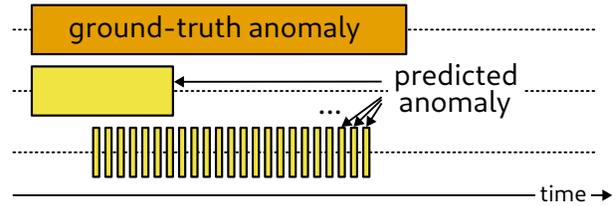}}
        \caption{
        Illustration of a ground-truth anomaly (orange) and two prediction patterns (yellow). 
        The upper prediction is timely and concise; the bottom prediction is delayed and oscillating, producing redundant alarms. Although both overlap equally with the true anomaly, they differ substantially in latency and redundancy—differences that many evaluation metrics ignore, often assigning them the same score.
        }
        \label{fig:first_example}
    \end{center}
    \vspace{-20pt}
\end{figure}
\vspace{-15pt}

However, TSAD methods are often evaluated with inconsistent and incompatible metrics \citep{lai2024nominality, wang2024drift, kim2024model}. 
Multiple works have shown that widely used metrics suffer from critical flaws, for example, significantly overestimating the performance of random models \citep{doshi2022tisat, gim2023evaluation, kim2022towards}, and several alternatives have been introduced to address such flaws \citep{wagner2023timesead, bhattacharya2025towards, si2024timeseriesbench}. 
Yet, most metrics are tailored to specific applications and lack a formal foundation, making it difficult to compare or generalize their behavior across different settings \citep{garg2021evaluation, abdulaal2021practical, gim2023evaluation, sorbo2024navigating}.

To enable trustworthy evaluations, metrics should behave in predictable ways that align with real-world needs. 
For example, a model that detects an anomaly is generally preferred over ones that fail to do so. 
Surprisingly, some commonly used metrics do not meet such requirements (\cref{sec:analysis}, \cref{tab:metrics-properties}).
his motivates a more systematic perspective: while definitive rankings of methods are generally difficult, meaningful comparisons often arise when methods differ in one key aspect, such as detection delay or alarm redundancy.
From such pairwise comparisons, we derive general, verifiable properties that characterize the behavior of evaluation metrics.
This characterization provides a useful tool for understanding the strengths and weaknesses of existing TSAD evaluation metrics while allowing for task-specific refinements.

This paper makes the following contributions to the study of TSAD evaluation metrics:
\begin{itemize}
\vspace{-6pt}\setlength\itemsep{1pt}
\item \textbf{A formal framework} for evaluating TSAD metrics, introducing two sets of nine verifiable properties that capture relevant aspects of evaluation metrics such as detection accuracy, alarm redundancy, and timing.
\item \textbf{A systematic theoretical study} of 37 existing TSAD metrics, revealing that no metric satisfies all properties, thus explaining persistent inconsistencies in TSAD evaluations.
\item \textbf{Two novel evaluation metrics}—LARM and ALARM—that provably satisfy all proposed properties, allow flexible control over application-specific trade-offs, and yield intuitive, consistent model rankings in practice.
\end{itemize}

\section{What makes a good metric?}\label{sec:motivation}
Metrics shape our perception of performance—and flawed metrics can distort it. This raises a fundamental question: what makes a good metric for TSAD?

Choosing the right method for a TSAD application can be overwhelming. The number of available methods is huge and evaluations often vary considerably. 
Suppose we assemble a list of candidate methods and select an appropriate benchmark dataset on which we train our methods. 
Each produces predictions on the test set that we can compare with the ground truth. 
At this point, the central difficulty emerges: Which metric should we use?
\begin{figure}[t]
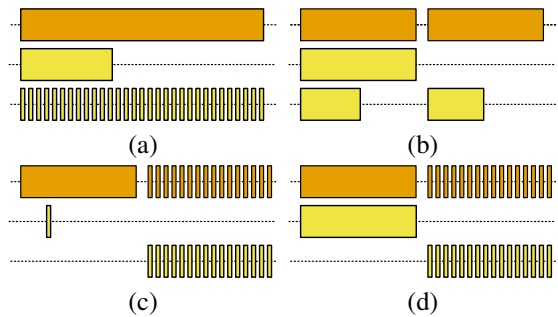

    \begin{center}
        \begin{tabular}{cc}
            \includesvg[inkscapelatex=false,scale=0.4]{figures/examples/example_precision_recall} & \includesvg[inkscapelatex=false,scale=0.4]{figures/examples/example_events_precision_recall} \\
            (a)\label{example:precision_recall} & (b)\label{example:precision_recall_pa} \\
            \includesvg[inkscapelatex=false,scale=0.4]{figures/examples/example_pa} & \includesvg[inkscapelatex=false,scale=0.4]{figures/examples/example_applied_pa} \\
            (c)\label{example:pa} & (d)\label{example:pa_applied} \\
        \end{tabular}
        \caption{Challenging examples of prediction patterns (yellow) for different ground truths (orange) that common evaluation metrics struggle to order.} 
        \label{fig:concrete_examples}
    \end{center}
    \vspace{-10pt}
\end{figure}

Popular choices in the literature are point-wise Precision (the fraction of correctly predicted anomalous time steps among all predicted anomalous time steps) and point-wise Recall (the fraction of correctly identified anomalous time steps among all time steps labeled anomalous).
By their point-wise nature, these metrics are unable to distinguish any patterns in the predictions such as depicted in \cref{fig:first_example}.
Consider the example in \cref{fig:concrete_examples}(a). 
Precision does not indicate a difference between the two methods, while Recall favors the second. 
In practice, however, we do not have access to the ground truth, and every alarm raised by the second method would require investigation, since they could be caused by distinct anomalies as illustrated in \cref{fig:concrete_examples}(b), where neither Precision nor Recall indicate a difference in performance. 

A common adjustment treats any prediction within an anomaly window as correct if at least one time step in the same window is predicted correctly \citep{xu2018unsupervised}.
With this adjustment, Recall favors the second method.
However, this introduces further problems. 
Consider the example in \cref{fig:concrete_examples}(c) and the adjusted predictions in \cref{fig:concrete_examples}(d). 
Increasing the leftmost anomaly window makes the first method eventually appear better according to Recall, while Precision remains unchanged. 
However, the first prediction could be a single lucky guess, which becomes increasingly likely as the size of the anomaly window increases \cite{doshi2022tisat, kim2022towards, liu2024elephant}.

\begin{figure}[ht!]
    \centering
    \includegraphics[width=\linewidth]{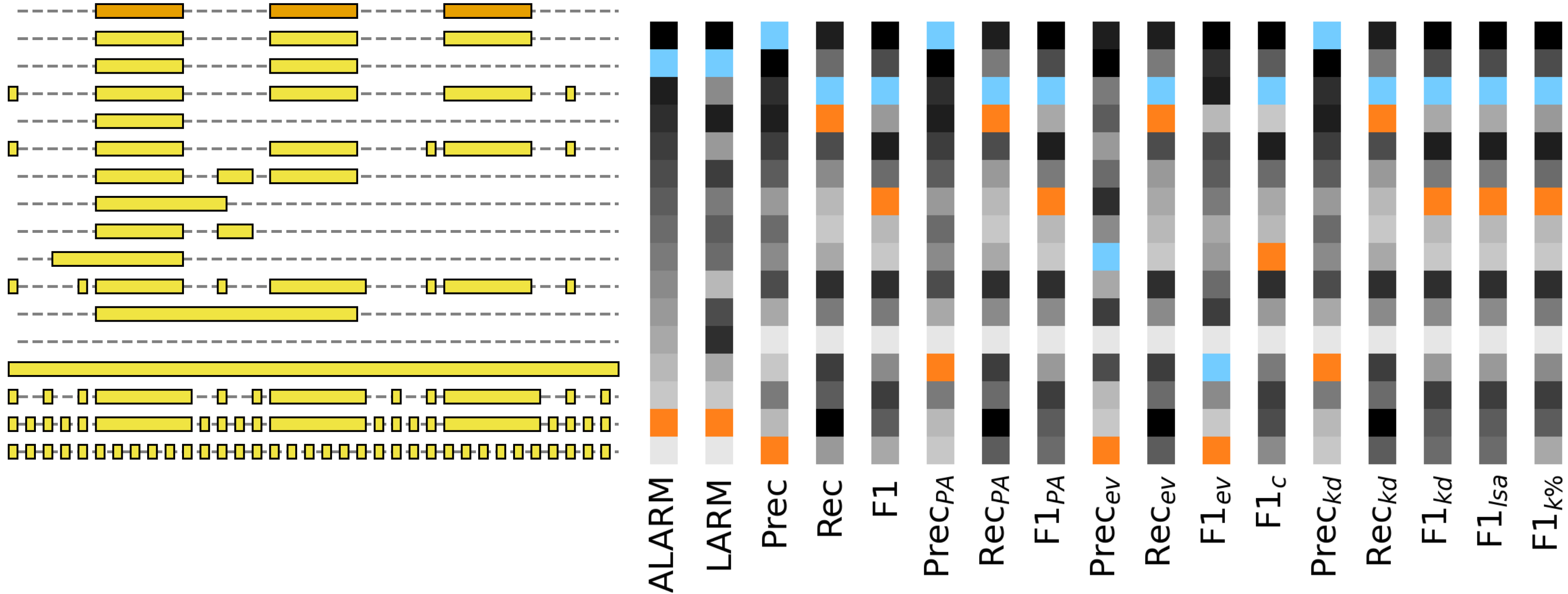}
    \caption{The rankings of 16 synthetic methods assigned an individual color each from best (top) to worst (bottom) by the ALARM score (\cref{theorem:advanced}) (left) compared to rankings of 15 other metrics (right), reveal that no two metrics agree on a ranking.
    }
    \label{fig:comparison_toy}
\end{figure}

To illustrate the variation in rankings of different metrics, we compare the rankings of 16 predictions with 17 metrics in \cref{fig:comparison_toy}. 
While the ranking on the left appropriately places perfect predictions first and effectively random predictions last, most metrics disagree considerably.

\paragraph{These observations lead to our main hypothesis:}
Many TSAD evaluation metrics have flaws that are easily overlooked and counterintuitive.
This highlights the need for a formal characterization of TSAD evaluation metrics and new evaluation setups that increase the transparency of evaluations.

\section{Related Work}\label{sec:related_work}

\paragraph{Point-adjusted metrics}
To remedy some of the unwanted effects of point-wise metrics \citep{paparrizos2022volume, wagner2023timesead, bhattacharya2025towards}, \cite{xu2018unsupervised} treat all missed anomalous time points as correct if at least one anomalous time point was predicted correctly in the same window.
Several works have investigated these \emph{point-adjusted} methods. 
Some conclude that point-adjustment introduces even worse unwanted effects \citep{doshi2022tisat, gim2023evaluation, kim2022towards} such as a preference for random predictions. 
Others have iterated on the idea and introduced their own variations of point-adjusted metrics \citep{hundman2018detecting, ren2019time, abdulaal2021practical, kim2022towards, gim2023evaluation, si2024timeseriesbench, bhattacharya2025towards}.

\paragraph{Range-based metrics}
Point-adjusted metrics tend to share one major drawback with point-wise metrics: the inability to distinguish patterns of true positives.
To address this flaw, \cite{tatbul2018precision} extend the classical notion of Precision and Recall to ranges by considering continuous intervals of anomalies and predictions.
Building on this idea, \cite{wagner2023timesead} propose extended versions addressing some of their flaws.
These metrics retain almost unparalleled degrees of freedom, making them highly flexible but difficult to apply in practice.

\paragraph{Fuzzy metrics}
Some metrics consider false positives approximately correct, as long as they are close enough to a ground-truth anomaly.
Some only include predictions after ground-truth anomalies \citep{hwang2019time, lavin2015evaluating}, while others include predictions on both sides of anomalies \citep{scharwachter2020statistical,huet2022local}.
A significant drawback of these metrics is the dependence of their parameters on the dataset. 
For example, without the expectation of noise in the ground truth, it should be impossible to detect anomalies early. 
Without extensive domain knowledge, the existence and severity of such noise are impossible to discern based on the data alone.

\paragraph{Detection-delay metrics}
Most evaluation metrics struggle to consider the delay until an anomaly is detected.
\cite{xu2018unsupervised} specifically account for this delay by calculating the average delay until a ground-truth anomaly is detected.
Similarly, \cite{kovács2019evaluation} consider the temporal distance to the next prediction after the start of a ground-truth anomaly.
Both studies propose to use these metrics in addition to other evaluation metrics, which introduces another trade-off to consider for evaluations.

\paragraph{Volume-under-surface metrics}
Anomaly detection methods usually compute a scalar score before applying a threshold to obtain binary predictions.
Thus, each method can potentially produce a wide range of predictions depending on the threshold.
When benchmarking, we only have a finite number of possible scores.
Thus, we can consider all evaluations simultaneously by computing the area under some curve that is parameterized by the threshold \citep{xu2018unsupervised, paparrizos2022volume, doshi2022tisat, ghorbani2024pate, si2024timeseriesbench}.
In this study, we exclude any metrics that are computed on scores directly, as they are much more complex, not directly compatible with our analysis, and are much less prevalent in TSAD evaluations \citep{xu2018unsupervised}.

\paragraph{Characterization of evaluation metrics}
When choosing an evaluation metric, we implicitly specify the desired behavior of a model.
Domain experts have the most insight into the desired behavior for their application.
Thus, metrics or at least specifications of the desired behavior should be provided alongside datasets \citep{hundman2018detecting, si2024timeseriesbench}.
Consequently, it is equally important to specify intended and actual behavior of a metric to match the expectations of a given task. 
Some works explicitly state that metrics should reward predictions when predicting at least one true positive for a ground-truth anomaly \citep{garg2021evaluation, jacob2020exathlon, lavin2015evaluating, gim2023evaluation, si2024timeseriesbench}, predicting true positives earlier in a ground-truth anomaly window \citep{abdulaal2021practical, doshi2022tisat, garg2021evaluation, jacob2020exathlon, lavin2015evaluating, gim2023evaluation, ghorbani2024pate, ren2019time, si2024timeseriesbench}, minimizing false positives \citep{abdulaal2021practical, garg2021evaluation, lavin2015evaluating}, in particular false alarms \citep{gim2023evaluation, si2024timeseriesbench}, maximizing true positives \citep{abdulaal2021practical, garg2021evaluation, jacob2020exathlon, ghorbani2024pate}, or minimizing redundant alarms \citep{doshi2022tisat, jacob2020exathlon}.
While providing useful intuitions and arguably useful goals, previous specifications lack formal rigor and are therefore not verifiable, making rigorous comparison and analysis impossible.

\paragraph{Evaluating evaluation metrics}
\cite{sorbo2024navigating} and \cite{liu2024elephant} compare the rankings of two or more predictions with respect to several evaluation metrics on concrete examples that cover timing of predictions, effect of anomaly length, coverage of ground-truth anomalies, or tolerance towards label noise \citep{sorbo2024navigating}, or timing of predictions or ranking of random predictions \citep{liu2024elephant}. 
While such examples can provide intuition and reveal unintuitive behavior \citep{liu2024elephant}, they can only provide definitive statements for the examples evaluated, which is not sufficient to support general implications, as small changes to an example can significantly affect the rankings (see \cref{sec:motivation}). 
Therefore, this study introduces the first formal framework to rigorously specify and evaluate general concepts, providing explanations for such unintuitive behavior.

\section{Formalizing Properties} \label{sec:properties}
In this section, we define the problem setting, explain and motivate the core assumptions, and finally define and discuss properties of evaluation metrics.

\subsection{Formal Problem Setting}
An evaluation metric $m$ is a function that compares a ground truth $g$ with predictions $p$.
Ground truth and predictions are binary sequences of equal length.
$0$ at any time step indicates normal behavior, while $1$ indicates an anomaly.
We can think of any sequence of length $n$ as a function on the domain $[n] = \{1, \ldots, n\}$.
Lastly, we define the set of alarms raised in a binary sequence $s \colon [n] \rightarrow \{0,1\}$. 
An alarm is a maximal continuous subsequence of $1$'s.
The set of all alarms in a sequence $s$ is given by 
$I_v(s) = \{[l,u] \subset [n] \;|\; s_{[l,u]} = v \land \nexists [l,u] \subsetneq [\hat{l}, \hat{u}] \subset [n] \colon s_{[\hat{l},\hat{u}]} = v\}$.
Then the number of alarms raised in $s$ is given by $|I_1(s)|$.

\subsection{Assumptions} \label{sec:assumptions}
Labeling data requires extensive domain knowledge.
Thus, any uncertainties in the labels is best resolved during the creation of a dataset.
With this in mind, we assume the following.
\begin{assumption}\label{ass:correct_labels}
    The labels are correct.
\end{assumption}
The assumption implies that alarms should be raised where the ground truth is labeled anomalous, and alarms should never be raised where the ground truth is labeled normal. 
The vast majority of metrics and evaluations are based on this assumption.

In real applications, we generally do not have access to the ground truth and thus have to rely entirely on the predictions.
This means that the system must act on its predictions as they occur, without additional context or corrections.
For example, once an alarm is raised, it is treated as a genuine indication of anomalous behavior, prompting intervention.
This clearly differentiates Anomaly Detection from Event Detection, where only the detection of anomalous events as a whole is of interest.
This leads us to the next assumption.
\begin{assumption}\label{ass:predictions} Predictions are taken at face value. \end{assumption}
This assumption is in line with the standard practice in the literature while emphasizing the need for precise, reliable predictions to minimize costly or impractical interventions. 
The main implication of this assumption is that all alarms generate costs.

In general, anomalies are rare, temporally bounded deviations from normal behavior.
It is impossible to definitively quantify how rare anomalies are, but to simplify the discussions in this paper, we assume that they are at least rare enough not to overlap.
\begin{assumption}\label{ass:ad}
    Anomalies are rare and do not overlap.
\end{assumption}
In particular, this assumption implies that multiple alarms raised in a single ground-truth anomaly window are always considered redundant.

To simplify the notation, we assume that $g$ is a ground truth of length $n$, and $p$ and $q$ are predictions of the same length, and are thus comparable by a metric. 
In general, a metric produces a scalar value that can either increase or decrease with improved performance. 
Without loss of generality, we assume that a higher value of $m$ corresponds to better performance.
\newlength{\figuresize}
\setlength{\figuresize}{0.3\textwidth}
\newlength{\horizontalbetweenfigures}
\setlength{\horizontalbetweenfigures}{10pt} %spacing between horizontally aligned subfigures
\newlength{\verticalbetweenfigures}
\setlength{\verticalbetweenfigures}{4pt} %spacing between vertically aligned subfigures
\begin{figure*}[ht!]
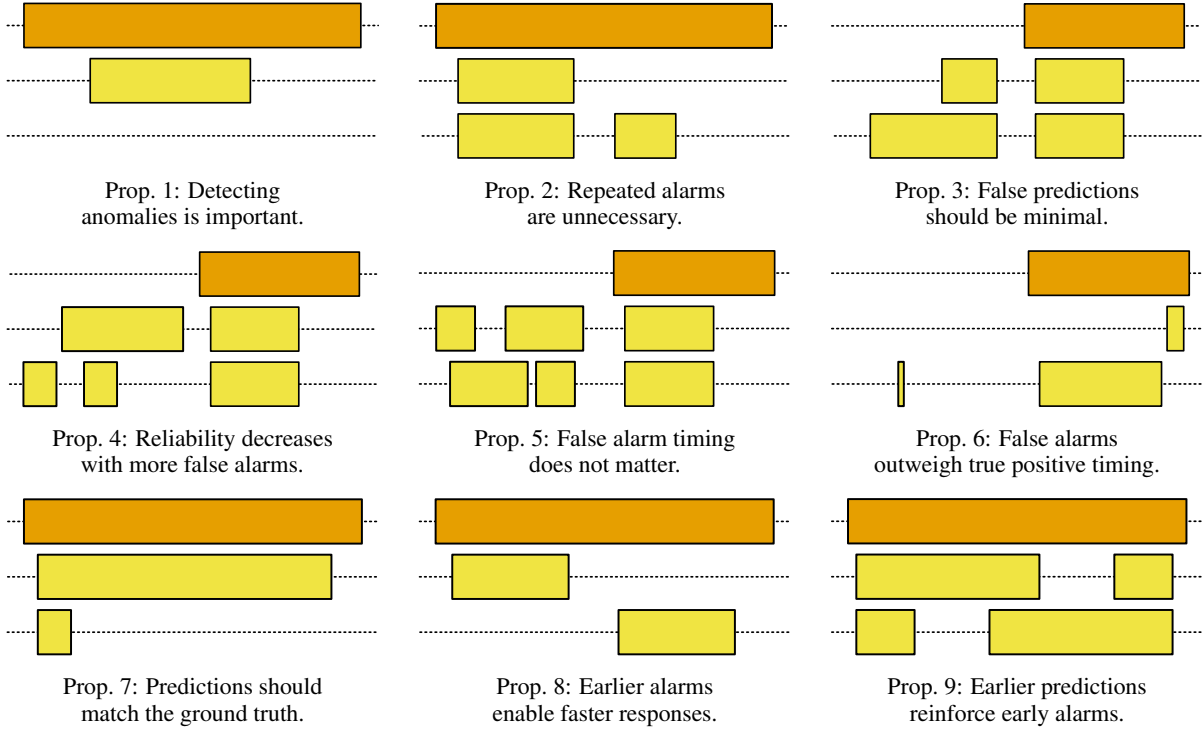

    \centering
    \begin{subfigure}{\figuresize}
    \includesvg[inkscapelatex=false,width=\figuresize]{figures/properties/prop_detection}
    \caption*{\centering\cref{prop:detection}: Detecting\\anomalies is important.}
    \end{subfigure}
    \hspace{\horizontalbetweenfigures}
    \begin{subfigure}{\figuresize}
    \includesvg[inkscapelatex=false,width=\figuresize]{figures/properties/prop_raised_alarms}
    \caption*{\centering\cref{prop:alarms}: Repeated alarms\\are unnecessary.}
    \end{subfigure}
    \hspace{\horizontalbetweenfigures}
    \begin{subfigure}{\figuresize}
    \includesvg[inkscapelatex=false,width=\figuresize]{figures/properties/prop_fp_size}
    \caption*{\centering\cref{prop:false_positives}: False predictions\\should be minimal.}
    \end{subfigure}
    
    \vspace{\verticalbetweenfigures}
    \begin{subfigure}{\figuresize}
    \includesvg[inkscapelatex=false,width=\figuresize]{figures/properties/prop_false_alarm_monotonicity}
    \caption*{\centering\cref{prop:false_positive_alarms}: Reliability decreases\\with more false alarms.}
    \end{subfigure}
    \hspace{\horizontalbetweenfigures}
    \begin{subfigure}{\figuresize}
    \includesvg[inkscapelatex=false,width=\figuresize]{figures/properties/prop_permutation}
    \caption*{\centering\cref{prop:permutations}: False alarm timing\\does not matter.}
    \end{subfigure}
    \hspace{\horizontalbetweenfigures}
    \begin{subfigure}{\figuresize}
    \includesvg[inkscapelatex=false,width=\figuresize]{figures/properties/prop_false_alarms}
    \caption*{\centering\cref{prop:trust}: False alarms\\outweigh true positive timing.}
    \end{subfigure}

    \vspace{\verticalbetweenfigures}
    \begin{subfigure}{\figuresize}
    \includesvg[inkscapelatex=false,width=\figuresize]{figures/properties/prop_prediction_size}
    \caption*{\centering\cref{prop:prediction_size}: Predictions should\\match the ground truth.}
    \end{subfigure}
    \hspace{\horizontalbetweenfigures}
    \begin{subfigure}{\figuresize}
    \includesvg[inkscapelatex=false,width=\figuresize]{figures/properties/prop_temporal_shift}
    \caption*{\centering\cref{prop:temporal_order}: Earlier alarms\\enable faster responses.}
    \end{subfigure}
    \hspace{\horizontalbetweenfigures}
    \begin{subfigure}{\figuresize}
    \includesvg[inkscapelatex=false,width=\figuresize]{figures/properties/prop_early_bias}
    \caption*{\centering\cref{prop:early_bias}: Earlier predictions\\reinforce early alarms.}
    \end{subfigure}  
    \caption{
    Visual motivation for the formalization of properties of TSAD evaluation metrics.
    Each panel shows a ground-truth anomaly (orange) and two prediction patterns (yellow), designed to isolate specific evaluation principles. 
    With the exception of \cref{prop:permutations}, each property formalizes the notion that a metric would prefer the top prediction, when it 
    %we prefer the top prediction: it 
    detects anomalies more reliably (Props. \ref{prop:detection}, \ref{prop:prediction_size}), avoids redundancy (\cref{prop:alarms}), is more precise (\cref{prop:false_positives}), raises fewer false alarms (Props. \ref{prop:false_positive_alarms}, \ref{prop:trust}), or responds earlier (Props. \ref{prop:temporal_order}, \ref{prop:early_bias}). 
}
    \label{fig:examples}
\end{figure*}

\subsection{Formalizing Properties}\label{sec:properties_definition}
Based on the specified problem setting and assumptions, we define nine properties that reflect different aspects of TSAD. 
While each aspect is important in its own right, different settings might consider some aspects more important than others.
In this section, we present one adaptation and discuss how each individual aspect and property can be relaxed or tightened depending on the specific requirements.
For each property, we provide a formal definition, explain the motivation, and visualize the intuition with motivating examples shown in \cref{fig:examples}.

\paragraph{Detecting anomalies}
Consider almost identical predictions that differ only in the fact that one raises at least one alarm on some anomaly window while the other does not. 
The method raising the alarm would definitively detect one anomaly the other does not.
The following property formalizes the notion that a metric prefers the former.
\begin{property}[Detection of Anomalies]\label{prop:detection}
    Let $A \in I_1(g)$, $p_{[n] \setminus A} = q_{[n] \setminus A}$, and $|I_1(p_A)| > |I_1(q_{A})| = 0$.
    Then $m(g,p) > m(g,q)$.
\end{property}
\paragraph{Additional alarms}
Any excess alarms raised after an anomaly has been detected first result in additional and unnecessary costs, and, potentially, loss of the users trust in the reliability of the predictions. 
\begin{property}[redundant alarms]\label{prop:alarms}
    Let $A \in I_1(g)$, $q(i) = \mathds{1}_{i\in I}+p(i)\mathds{1}_{i\not\in I}$
    for some $I \subset A$ with $\min I > \max \{i \in A\colon p(i) = 1\}$ and $p_I = 0$, and $|I_1(q_A)| = |I_1(p_A)| + 1$.
    Then $m(g, p) > m(g, \hat{p})$.
\end{property}
\paragraph{Falsely predicting anomalies}
Making false positive predictions, time steps where a method incorrectly predicts an anomaly where there is none, decreases the quality of the prediction. 
However, false predictions can appear in different places and can even affect the number of false alarms, introducing a trade-off. 
Let us first consider the case where the number of alarms remains unchanged.
\begin{property}[minimizing false positives]\label{prop:false_positives}
    Let $i^* \in N \in I_0(g)$, $p_{[n] \setminus \{i^*\}} = q_{[n] \setminus \{i^*\}}$, $q(i^*) = 1 \neq 0 = p(i^*)$, and $|I_1(p_N)| = |I_1(q_N)|$.
    Then $m(g,p) > m(g,q)$.
\end{property}
Generally, more impactful than the total number of false positives is the number of false alarms raised as a result.
No matter how many false positives are predicted, a false alarm significantly decreases the reliability of the underlying method.
Too many false alarms render a method unusable for many applications.
\begin{property}[minimizing false alarms]\label{prop:false_positive_alarms}
    Let $N \in I_0(g)$, $p_{[n] \setminus N} = q_{[n] \setminus N}$, and $|I_1(p_N)| < |I_1(q_N)|$.
    Then $m(g,p) > m(g,q)$.
\end{property}
Since a false alarm does not correspond to an underlying anomaly (\cref{ass:correct_labels}), the exact time at which it is raised is irrelevant. 
The following property formalizes the indifference of a metric to the timing of false alarms.
\begin{property}[invariance under permutation of false positives]\label{prop:permutations}
    Let $N \in I_0(g)$, $p_{[n] \setminus N} = q_{[n] \setminus N}$, $|q^{-1}(1)| = |p^{-1}(1)|$, and $|I_1(p_N)| = |I_1(q_N)|$.
    Then $m(g, p) = m(g, q)$.
\end{property}
Each false alarm severely reduces the user confidence in the underlying method.
If the trust of the user erodes enough, they will no longer want to use the method.
Thus, predictions with more false alarms are often worse, even if they are otherwise comparable.
\begin{property}[Keeping user's trust]\label{prop:trust}
    Let $A \in I_1(g)$, $N \in I_0(g)$, $p_{[n] \setminus (A \cup N)} = q_{[n] \setminus (A \cup N)}$, $|I_1(p_{A})| = |I_1(q_{A})|$, $p_{N} = 0$, and $q_{N} = \mathds{1}_{\{i^*\}}$ for some $i^* \in N$.
    Then $m(g,p) > m(g,q)$.
\end{property}
\paragraph{Preciseness of predictions}
Intuitively, the more anomalous time steps a method detects, the better.
However, more such true positives can increase the number of alarms, resulting in worse performance according to \cref{prop:alarms}.
\begin{property}[maximizing true positives]\label{prop:prediction_size}
    Let $A \in I_1(g)$, $p_{[n] \setminus \{i^*\}} = q_{[n] \setminus \{i^*\}}$ for some $i^* \in A$, $p(i^*) = 1 \neq 0 = q(i^*)$, and $|I_1(p_A)| \leq |I_1(q_A)|$.
    Then $m(g, p) > m(g, q)$.
\end{property}
\paragraph{Timing of alarms}
Predicting more true positives could decrease the time until an anomaly is first detected.
\cref{ass:predictions} clearly implies a preference for early alarms.
Intentionally avoiding the trade-off with the number of alarms and number of true positives, we assume that both are constant for this comparison.
\begin{property}[alarm timing]\label{prop:temporal_order}
    Let $A \in I_1(g)$, $p_{[n] \setminus A} = q_{[n] \setminus A}$, $|I_1(p_A)| = |I_1(q_A)|$, $|p^{-1}(1)| = |q^{-1}(1)|$, and $\min \{i \in A\colon p(i) = 1\} < \min \{i \in A\colon q(i) = 1\}$. 
    Then $m(g, p) > m(g, q)$.
\end{property}
If the true positives of a prediction are aligned early in a ground-truth anomaly, more time is spent with earlier alarms, leaving more time to address the anomaly and potentially avoid later alarms entirely.
\begin{property}[Early Bias]\label{prop:early_bias}
    Let $A \in I_1(g)$, $p_{[n] \setminus \{i^*, i^{**}\}} = q_{[n] \setminus \{i^*, i^{**}\}}$ for some $i^{*} < i^{**}\in A$ with $p(i^{*}) = 1 \neq 0 = q(i^{*})$ and $p(i^{**}) = 0 \neq 1 = q(i^{**})$ such that $|I_1(p_A)| \leq |I_1(q_A)|$. 
    Then $m(g, p) > m(g, q)$.   
\end{property}

\subsection{Are these properties sufficient?}\label{sec:advanced_props}

Each property formalizes individual aspects of evaluations, which allows us to rigorously verify which properties each metric satisfies (\cref{sec:analysis}) and explicitly design metrics that satisfy a given set of such properties (\cref{sec:discussion}). 
However, the above properties rely on some simplifications. 
\begin{wrapfigure}{r}{0.4\linewidth}
    \centering
    \vspace{-0.2cm}
    \includesvg[width=\linewidth]{figures/discussion/detection_example.svg}
    \vspace{-0.7cm}
    \label{fig:detection_example}
\end{wrapfigure}
Firstly, we could question whether a ground-truth anomaly is actually detected, if it contains a true positive.
Consider the example where a single predicted window overlaps with two distinct ground-truth anomalies, which do not overlap with other predicted windows.
Since only one alarm is raised, we could reasonably argue that only the first anomaly is detected.

Changing this definition changes the notion of false alarms. 
So far, all windows of false positives are equally important to false alarms.
However, predictions that partially overlap with anomalies technically do not produce additional alarms.
Therefore, predictions that are not overlapping with any anomaly windows---true false alarms---could be worse than early alarms or late alarms. 
However, considering \cref{ass:predictions} and \cref{ass:correct_labels}, a response system or operator might mistakenly dismiss an early alarm as a false alarm, since there are no detectable effects in the data for some time after it was raised.
\begin{wrapfigure}{r}{0.4\linewidth}
    \centering
    \vspace{-0.2cm}
    \includesvg[width=\linewidth]{figures/discussion/example_false_alarms_wide.svg}
    \vspace{-0.7cm}
    \label{fig:false_alarms_example}
\end{wrapfigure}
Thus, we could end up with a differentiated ranking such as the one shown on the right.
These restrictions on false alarms can have strong implications. 
For example, a metric that scores a method that produces any false positives lower than the method that produces no predictions could easily satisfy these properties (\cref{theorem:main}).
Additionally, \cref{prop:trust} imposes very strong restrictions on false alarms which might not apply to all applications equally. 
However, we can control the influence of false alarms explicitly by introducing a tolerance (\cref{theorem:advanced}).

Considering these details introduces additional complexity into the definitions and proofs.
We provide a set of advanced properties that incorporate all the points raised in this discussion in \cref{appendix:advanced_props}.
There, we illustrate how we can adapt our analysis to such changing requirements and how most results translate to this setting without much effort. 
For simplicity, we will work with the simple properties introduced in \cref{sec:properties} unless explicitly stated otherwise.
In the following section, we will show that no existing TSAD metric satisfies all simple properties, in fact, most only satisfy few if any.

\section{Where evaluation metrics fail}\label{sec:analysis}
\setlength{\tabcolsep}{5pt}
\begin{table*}[h!t]
\caption{None of the considered metrics fulfill all properties introduced in \cref{sec:properties}. 
Each entry shows whether a metric satisfies (\cmark~) a property or not (\xmark~).
(\xmark)~( (\cmark) )indicates that, while for most parameters there are examples where the property does not hold (hold), the method in question has parameters for which the property holds (does not hold).
All proofs can be found in \cref{appendix:proofs}.}
\label{tab:metrics-properties}
\begin{center}
\resizebox{\textwidth}{!}{
\begin{tabular}{lllccccccccc}
\toprule
\multicolumn{3}{c}{\textbf{Metric$\downarrow$~\textbackslash~Property$\rightarrow$}}
 & \ref{prop:detection}
 & \ref{prop:alarms}
 & \ref{prop:false_positives}
 & \ref{prop:false_positive_alarms}
 & \ref{prop:permutations}
 & \ref{prop:trust}
 & \ref{prop:prediction_size}
 & \ref{prop:temporal_order}
 & \ref{prop:early_bias} \\
\midrule
\multirow{3}{*}{\rotatebox[origin=c]{90}{\parbox{0.7cm}{point- wise}}} & Precision & (\cref{proof:point-wise_precision}) & \xmark & \xmark & \xmark & \xmark & \cmark & \xmark & \xmark & \xmark & \xmark \\
& Recall & (\cref{proof:point-wise_recall}) & \cmark & \xmark & \xmark & \xmark & \cmark & \xmark & \cmark & \xmark & \xmark \\
& F1 & (\cref{proof:point-wise_f1-score}) & \cmark & \xmark & \xmark & \xmark & \cmark & \xmark & \cmark & \xmark & \xmark \\
\hline
\multirow{14}{*}{\rotatebox[origin=c]{90}{point-adjusted}} & Precision$_{PA}$ \citep{xu2018unsupervised} & (\cref{proof:pa-precision}) & \xmark & \xmark & \xmark & \xmark & \cmark & \cmark & \xmark & \xmark & \xmark \\
&Recall$_{PA}$ \citep{xu2018unsupervised} & (\cref{proof:pa-recall}) & \cmark & \xmark & \xmark & \xmark & \cmark & \xmark & \xmark & \xmark & \xmark \\
&F1$_{PA}$ \citep{xu2018unsupervised} & (\cref{proof:pa-f1-score}) & \cmark & \xmark & \xmark & \xmark & \cmark & \cmark & \xmark & \xmark & \xmark \\
&Precision$_{event}$ \citep{hundman2018detecting} & (\cref{proof:pa-precision-event}) & \xmark & \xmark & \xmark & \xmark & \xmark & \xmark & \xmark & \xmark & \xmark \\
&Recall$_{event}$ \citep{hundman2018detecting} & (\cref{proof:pa-recall-event}) & \cmark & \xmark & \xmark & \xmark & \cmark & \xmark & \xmark & \xmark & \xmark \\
&F1$_{event}$ \citep{hundman2018detecting} & (\cref{proof:pa-f1-event}) & \cmark & \xmark & \xmark & \xmark & \xmark & \xmark & \xmark & \xmark & \xmark \\
& F$_{c1}$ & (\cref{proof:point-wise_composite_f1-score}) & \cmark & \xmark & \xmark & \xmark & \cmark & \cmark & \xmark & \xmark & \xmark \\
&Precision$_{k-delay}$ \citep{ren2019time} & (\cref{proof:pa-precision-k-delay}) & \xmark & \xmark & \xmark & \xmark & \cmark & \xmark & \xmark & \xmark & \xmark \\
&Recall$_{k-delay}$ \citep{ren2019time} & (\cref{proof:pa-recall-k-delay}) & \xmark & \xmark & \xmark & \xmark & \cmark & \xmark & \xmark & \xmark & \xmark \\
&F1$_{k-delay}$ \citep{ren2019time} & (\cref{proof:pa-f1-k-delay}) & \xmark & \xmark & \xmark & \xmark & \cmark & \xmark & \xmark & \xmark & \xmark \\
&F1$_{lsa}$(\citealp{abdulaal2021practical}) & (\cref{proof:pa-f1-lsa}) & \xmark & \xmark & \xmark & \xmark & \xmark & \xmark & \xmark & (\xmark) & \xmark \\
&F1$_{PA\% K}$  \citep{kim2022towards} & (\cref{proof:pa-f1-k-percent}) & \cmark & \xmark & \xmark & \xmark & \cmark & (\xmark) & (\xmark) & \xmark & \xmark \\
&$\int_K$F1$_{PA\% K}$  \citep{kim2022towards} & (\cref{proof:pa-f1-k-percent-int}) & \cmark & \xmark & \xmark & \xmark & \cmark & \xmark & \cmark & \xmark & \xmark \\
&F1$_{PADF}$  \citep{gim2023evaluation} & (\cref{proof:pa-f1-df}) & \cmark & \xmark & \xmark & \xmark & \cmark & (\xmark) & \xmark & (\cmark) & \xmark \\
&F1$_{reduced-length}$  \citep{si2024timeseriesbench} & (\cref{proof:pa-f1-reduced-length}) & \cmark & \xmark & \xmark & \xmark & \cmark & \cmark & \xmark & \xmark & \xmark \\
& F1$_{BA}$ \citep{bhattacharya2025towards} & (\cref{proof:pa-f1-ba}) & \xmark & \xmark & \xmark & \xmark & \xmark & \xmark & \xmark & \xmark & \xmark \\
\hline
\multirow{6}{*}{\rotatebox[origin=c]{90}{range-based}}&Precision$_{TS}$ \citep{tatbul2018precision} & (\cref{proof:ts-precision}) & \xmark & \xmark & \xmark & \xmark & \xmark & \xmark & \xmark & \xmark & \xmark \\
&Recall$_{TS}$ \citep{tatbul2018precision} & (\cref{proof:ts-recall}) & \cmark & \xmark  & \xmark & \xmark &  \cmark & \xmark & \cmark & \xmark & \cmark \\
&F1$_{TS}$ \citep{tatbul2018precision} & (\cref{proof:ts-f1})& \xmark & \xmark & \xmark & \xmark & \xmark & \xmark & \cmark & \xmark & \xmark \\
&TPrec \citep{wagner2023timesead} & (\cref{proof:timesead-precision}) & \xmark & \xmark & \xmark & \xmark & \xmark & \xmark & \xmark & \xmark & \xmark \\
&TRec \citep{wagner2023timesead} & (\cref{proof:timesead-recall}) & \cmark & \xmark  & \xmark & \xmark &  \cmark & \xmark & \cmark & \xmark & \cmark \\
&TF1 \citep{wagner2023timesead} & (\cref{proof:timesead-f1}) & \xmark & \xmark & \xmark & \xmark & \xmark & \xmark & \cmark & \xmark & \xmark \\
\hline
\multirow{8}{*}{\rotatebox[origin=c]{90}{fuzzy labels}} &NAB \citep{lavin2015evaluating} & (\cref{proof:nab}) & (\cmark) & \xmark & (\cmark) & \xmark & (\xmark) & \xmark & \xmark & \cmark & \xmark \\
& TaP \citep{hwang2019time} & (\cref{proof:tap}) & \xmark & \xmark & \xmark & \xmark & \xmark & \xmark & \xmark & \xmark & \xmark \\
&TaR \citep{hwang2019time} & (\cref{proof:tar}) & (\cmark) & \xmark & \xmark & \xmark & \xmark & \xmark & (\cmark) & \xmark & \xmark  \\
& $P_\delta$ \citep{scharwachter2020statistical} & (\cref{proof:p_delta}) & \xmark & \xmark & \xmark & \xmark & (\xmark) & \xmark & \xmark & \xmark & \xmark \\
& $R_\delta$ \citep{scharwachter2020statistical} & (\cref{proof:r_delta}) & (\xmark) & \xmark & \xmark & \xmark & (\xmark) & \xmark & (\xmark) & \xmark & \xmark \\
&P$_{Precision}$ \citep{huet2022local} & (\cref{proof:p_precision}) & \xmark & \xmark & \xmark & \xmark & \xmark & \xmark & \xmark & \xmark & \xmark \\
&P$_{Recall}$ \citep{huet2022local} & (\cref{proof:p_recall}) & \xmark & \xmark & \xmark & \xmark & \xmark & \xmark & \xmark & \xmark & \xmark \\
&F1$_{affiliation}$ \citep{huet2022local} & (\cref{proof:f1_affiliation}) &\xmark  &\xmark  &\xmark  &\xmark  &\xmark  &\xmark  &\xmark  &\xmark &\xmark  \\
\hline
\multirow{4}{*}{\rotatebox[origin=c]{90}{other}}& AAD \citep{xu2018unsupervised} & (\cref{proof:average_alert_delay}) & \xmark & \xmark & \xmark & \xmark & \cmark & \xmark & \xmark & \cmark & \xmark \\
&TD (\citealp{kovács2019evaluation}) & (\cref{proof:temporal_distance}) & \cmark & \xmark & \xmark & \xmark & \xmark & \xmark & \cmark & \xmark & \xmark \\
& eTaP \citep{hwang2022you} & (\cref{proof:sTaP}) & \xmark & \xmark & \xmark & \xmark & \cmark & \xmark & \xmark & \xmark & \xmark  \\
& eTaR \citep{hwang2022you} & (\cref{proof:eTaR}) & \xmark & \xmark & \xmark & \xmark & \cmark & \xmark & \xmark & \xmark & \xmark  \\
\bottomrule
\end{tabular}
}
\end{center}
\vspace{-7pt}
\end{table*}
In this section, we investigate 37 evaluation metrics with respect to the properties defined in \cref{sec:properties_definition}.
The following theorem summarizes the results of our analysis\footnote{We provide detailed definitions for all metrics and proofs for all properties in \cref{appendix:proofs}.}.
\begin{theorem}\label{theorem:summary}
    No existing TSAD evaluation metric satisfies all properties \cref{prop:detection} - \cref{prop:early_bias}.
\end{theorem}
\cref{tab:metrics-properties} summarizes the results of our formal analysis in detail.
While one could argue that individual properties, for example \cref{prop:trust}, are too specific to apply in a general setting, our analysis reveals that no metric satisfies a majority of these properties. 

By definition, point-wise metrics ignore any temporal information in the data and as a result are unable to satisfy \cref{prop:temporal_order} and \cref{prop:early_bias} which entirely depend on the temporal information in the data.
With few exceptions, point-adjusted metrics generally struggle with the same properties, since most adjustment protocols intentionally remove any temporal information to emphasize detection capabilities. 
As a consequence, it is no surprise that they also struggle with \cref{prop:alarms}, \cref{prop:false_positives}, and \cref{prop:false_positive_alarms}.
On the contrary, range-based metrics are explicitly designed to incorporate temporal information through an explicit weighting scheme inside ground-truth anomaly windows.
For this analysis, we focus on parametrizations which are recommended to deal with this temporal information and leave the general case to the interested reader.

Realizing that a single metric is often insufficient to completely explain the performance of different methods, most evaluations rely on a combination of Precision and Recall and their variations.
Interestingly, many of these variations do not satisfy \cref{prop:detection} which in some cases can lead to situations where perfect predictions are outperformed by objectively worse methods (for example, see \cref{proof:p_precision}).
Taking the prediction order properties (\cref{prop:temporal_order} and \cref{prop:early_bias}) aside, most metrics only satisfy three properties at most.
It goes to show that many metrics often reward predictions counter-intuitively and thus produce unexpected results.

All metrics have their use, but their weaknesses and biases should be actively considered.
Our analysis raises the question how applicable many metrics are for the general evaluation of TSAD methods.
To solve this question, we introduce new metrics that satisfy all properties in the next section.

\begin{figure*}[ht!]
    \centering
    \begin{subfigure}{0.3\textwidth}
    \includegraphics[width=\textwidth]{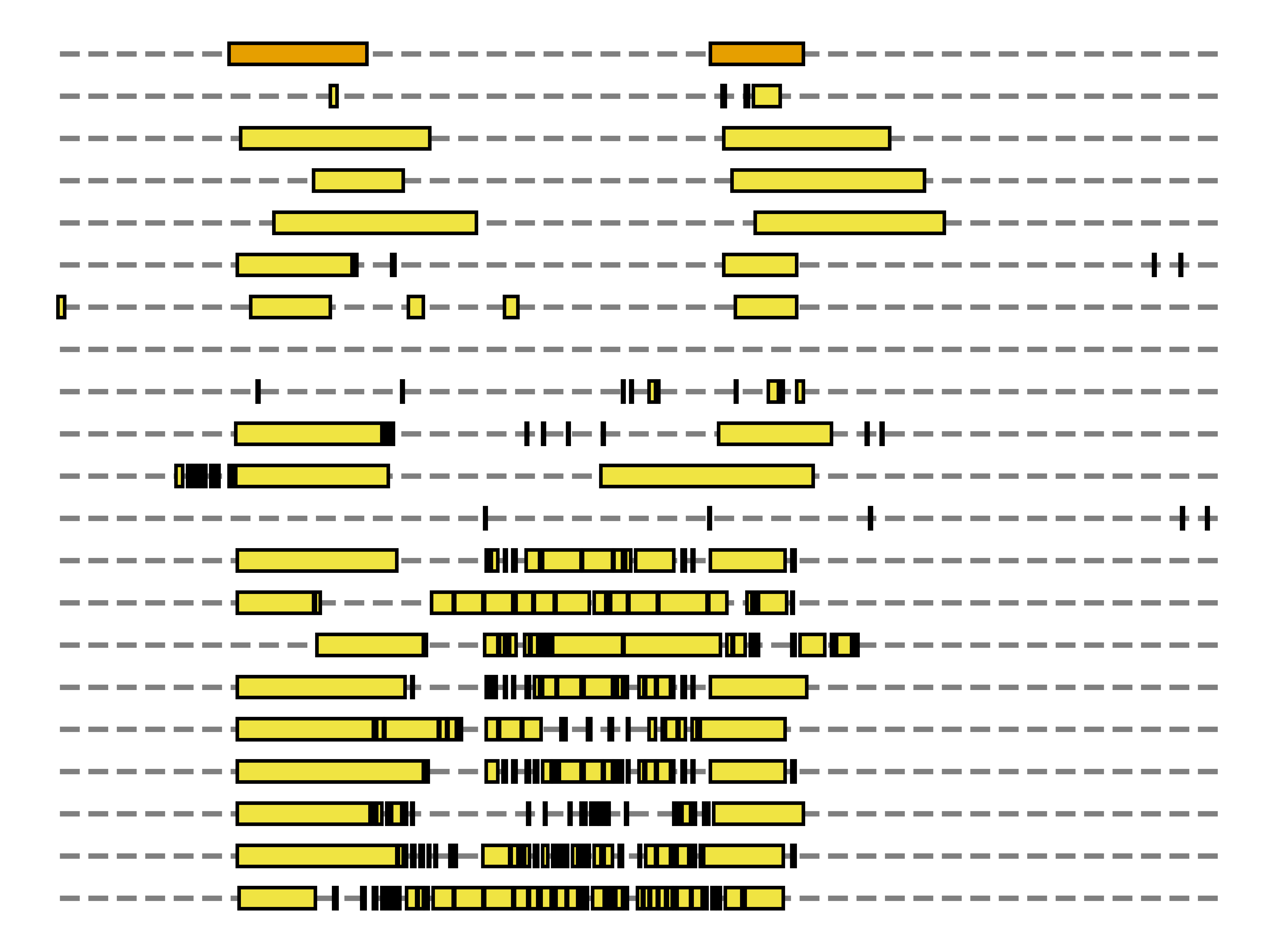}
    \caption*{\centering ALARM score}
    \end{subfigure}
    \begin{subfigure}{0.3\textwidth}
    \includegraphics[width=\textwidth]{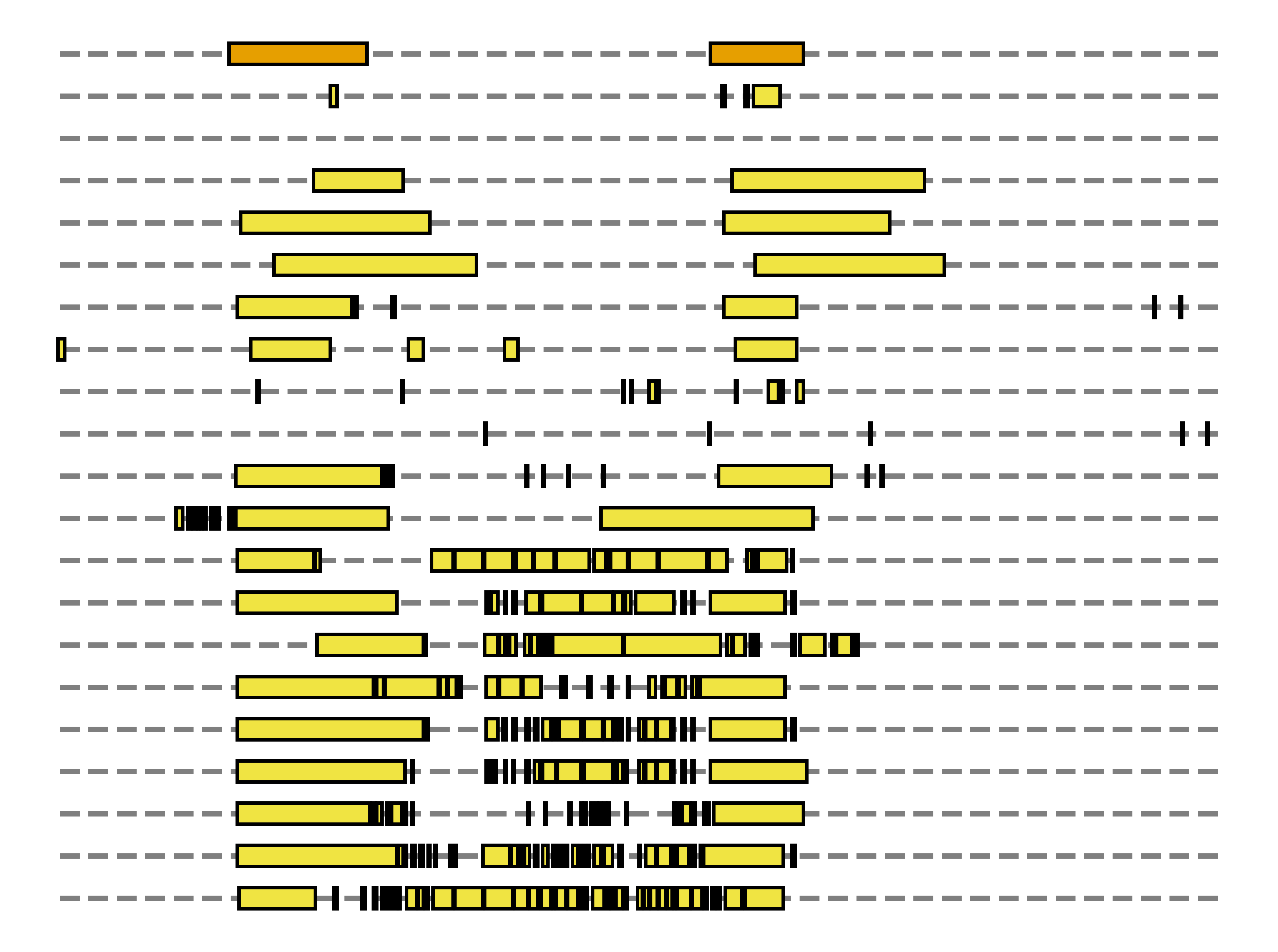}
    \caption*{\centering LARM score}
    \end{subfigure}
    \begin{subfigure}{0.3\textwidth}
    \includegraphics[width=\textwidth]{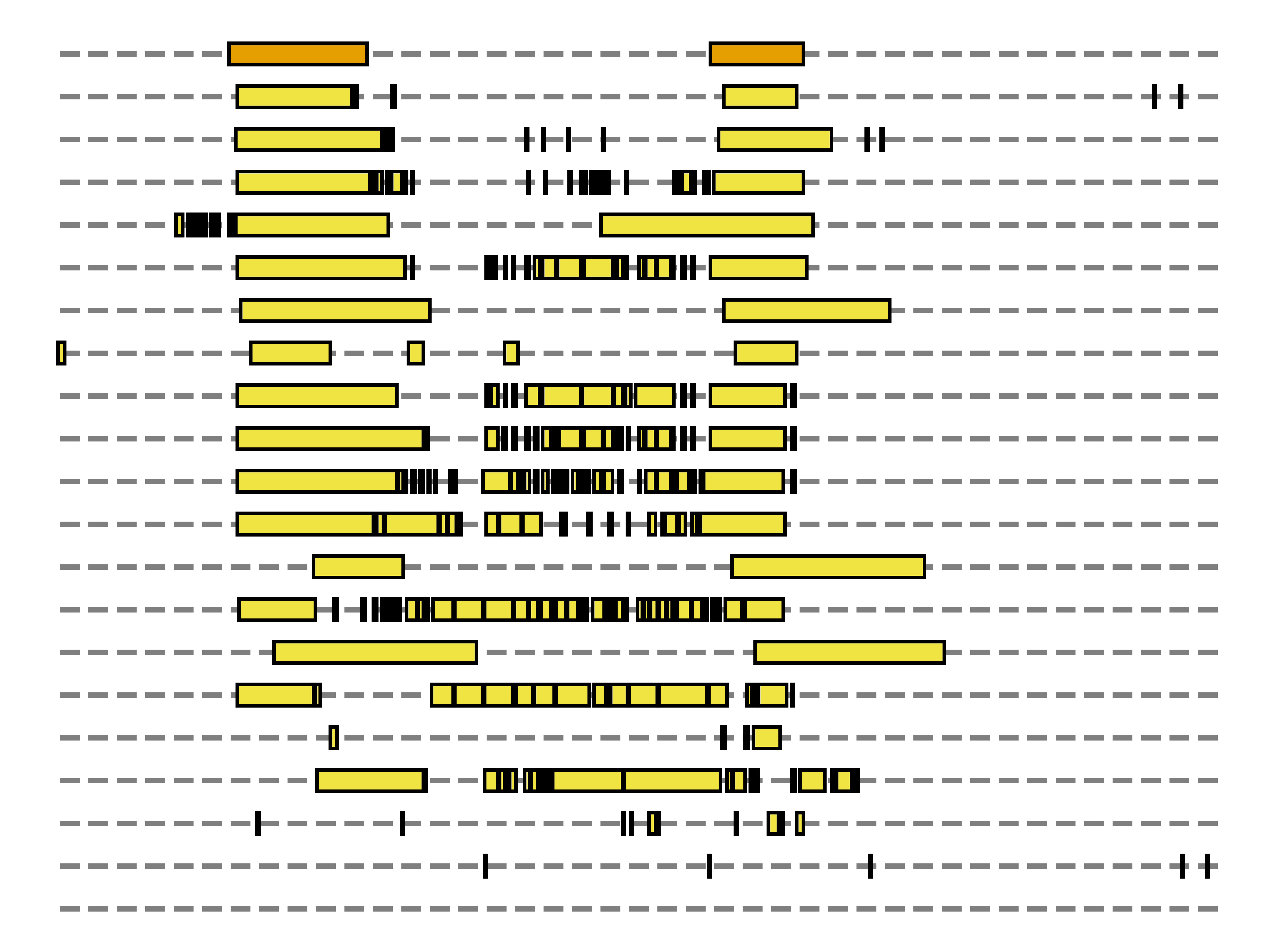}
    \caption*{\centering point-wise F1 score}
    \end{subfigure}
    \hfill
    \caption{Rankings produced on a subset of SMD8 \citep{su2019robust} of 20 methods \citep{wagner2023timesead} by different metrics, illustrate their different priorities. Assuming the labels are correct, most methods produce a staggering number of false alarms almost approaching random predictions, which is severely penalized by both LARM and ALARM. Notably, both group predictions with similar behavior (e.g. rows 2,3, and 4 in the left figure) with similar scores, while classic metrics often fail to do so (see e.g. right figure).}
    \label{fig:comparison}
\end{figure*}
\section{Towards Trustworthy Evaluations}\label{sec:discussion}
The properties defined in \cref{sec:properties} formalize several relevant aspects of TSAD, leading to the question of whether a single metric can simultaneously satisfy all of them.
To answer this question, let us consider the effects of each property on a metric.
The properties \cref{prop:detection}, \cref{prop:alarms}, \cref{prop:prediction_size}, \cref{prop:temporal_order}, and \cref{prop:early_bias} deal with true positives and are therefore not affected by changes in false positives. 
Similarly, we can group properties that deal with false alarms (\cref{prop:false_positive_alarms}, \cref{prop:permutations}, and \cref{prop:trust}), and properties that deal with false positives (\cref{prop:false_positives}).
While not represented by one of the simple properties explicitly, we can add another group dealing with detected anomalies to account for the discussion in \cref{sec:advanced_props}. 
We can now define a metric to satisfy all properties simultaneously by separately addressing the effects of each group as long as they do not overlap. 
For example, by introducing tiers for each anomaly, we can increase or decrease a predictions tier for each detected anomaly or each false alarm correspondingly. 
As long as the changes induced by properties dealing with patterns, stay within a single tier, we can disentagle the individual effects.
The following theorem introduces a variation of such a metric that satisfies all simple properties introduced in \cref{sec:properties}.
\begin{theorem}\label{theorem:main}
    The a\textbf{L}ignment \& \textbf{A}ccu\textbf{R}acy \textbf{M}etric (LARM) defined as
    \begin{align*}
        &\frac{1}{|I_{1}(g)|}\sum_{\stackrel{A\in I_1(g)\colon}{|I_1(p_{A})| > 0}} \frac{\alpha(p_A) + 1}{2^{|I_{1}(p_A)|}}\\
    &- 2\sum_{A \in I_{0}(g)} |I_{1}(p_A)|
    - \sum_{A \in I_{0}(g)} \beta(|p_{A}^{-1}(1)|)
    \end{align*}
    satisfies all properties \cref{prop:detection} - \cref{prop:early_bias}, where $\alpha$ is monotonically increasing with $|p_{A}^{-1}(1)|$, satisfies early bias (in the sense of \cref{prop:early_bias}), and $\codom (\alpha) \subset [0,1)$, and $\beta$ is monotonically increasing in $|p_{A}^{-1}(1)|$ with $\codom (\beta) \subset [0,1]$.
\end{theorem}
For example, $\alpha(p_A) = \sum_{i \in A} p_A(i) \cdot 2^{-i}$ and $\beta(x) = 1 - x^{-1}$ satisfy the requirements of \cref{theorem:main}, which we will use as default choice. 
By separating the metric into individual terms that address specific subsets of properties, we only need to show how the respective terms are affected when comparing methods.
The complete proofs can be found in \cref{appendix:metric}.

Using the same methodology, the following theorem introduces a metric that satisfies the advanced properties discussed in \cref{sec:advanced_props} and \cref{appendix:advanced_props}.
\begin{theorem}\label{theorem:advanced}
    The \textbf{A}dvanced a\textbf{L}ignment \& \textbf{A}ccu\textbf{R}acy \textbf{M}etric (ALARM)
    \begin{align*}
        &|DA(p,g)| 
    + \frac{\sum\limits_{A \in DA(p,g)} \frac{\alpha(p_{A}) + 1}{2^{|I_{1}(p_A)|}}}{|DA(p,g)|}
    - \beta(|p_{g=0}^{-1}(1)|)\\
    &-  t^{-1}\cdot\left(|TA(p,g)| + \frac{3}{2}|EA(p,g)| + \frac{1}{2}|LA(p,g)|\right)
    \end{align*}
    satisfies all advanced properties from \cref{sec:advanced_props} and \cref{appendix:advanced_props}, where $\alpha$ and $\beta$ are defined as in \cref{theorem:main}, $DA(p,g)$ is the set of detected anomalies, $TA(p,g)$ the set of true false alarms, $LA(p,g)$ the set of late false alarms, $EA(p,g)$ the set of early alarms (formal definitions in \cref{appendix:advanced_props}.), and $t \in \mathbb{N}_{>0}$ the alarm tolerance.
\end{theorem}
The ALARM\footnote{\url{https://github.com/wagner-d/tsadm}.} score explicitly differentiates alarms that partially overlap with ground-truth anomalies to the left (early) and to the right (late), and false alarms that do not overlap with ground-truth anomalies (true). 
We provide formal definitions, a detailed discussion, and complete proofs in \cref{appendix:advanced_props}.

Notably, the constant prediction $p=0$ achieves a score of 0 with both scores. 
With the ALARM score, any predictions with a lower score contain at least one true false alarm and, in general, at least $t$ times as many false alarms as detected anomalies. 
The alarm tolerance $t$ is the main parameter of the ALARM score. 
It specifies how many false alarms each detected anomaly compensates for, intuitively reflecting the system robustness with respect to false alarms. 
By default, we set this parameter to two.

To provide intuition for the capabilities of these metrics, we compare the rankings produced by these metrics with the point-wise F1 score in \cref{fig:comparison}.
The comparison clearly shows the substantial impact of false alarms in the LARM metric, which is significantly reduced by the alarm tolerance introduced in \cref{theorem:advanced}. We provide a detailed analysis of this comparison and additional evaluations on more datasets in \cref{appendix:evaluation}.

\section{Conclusion}\label{sec:conclusion}
In this paper, we study evaluation metrics for time-series anomaly detection (TSAD). 
We illustrate that common metrics produce vastly different rankings of methods and that these can be quite counterintuitive.
We formalize several key aspects of TSAD through nine simple and nine advanced properties of evaluation metrics and formally prove that no existing metric satisfies all simple properties explaining this counterintuitive behavior.
Finally, we introduce two new general metrics (A)LARM that provably satisfy all nine (advanced) properties.

\newpage
\paragraph{Acknowledgments}
The main part of this work was conducted within the DFG Research Unit FOR 5359 (ID 459419731) on Deep Learning on Sparse Chemical Process Data. MK and SF further acknowledge support by the DFG through TRR 375 (ID 511263698), SPP 2298 (Id 441826958), and SPP 2331 (441958259), by the Carl-Zeiss Foundation through the initiatives AI-Care and Process Engineering 4.0, and by the BMFTR award 01IS24071A.
{\small
\bibliographystyle{abbrvnat}
\bibliography{bibliography}
}

\newpage
\appendix
\newpage
\onecolumn

\section{ALignment \& Accu\textbf{R}acy \textbf{M}etric (LARM) }\label{appendix:metric}
\begin{theorem}
    There exist metrics that satisfy all properties \cref{prop:detection} - \cref{prop:early_bias}.
    In particular, the a\textbf{L}ignment \& \textbf{A}ccu\textbf{R}acy \textbf{M}etric (LARM) defined as
    \begin{align*}
        \frac{1}{|I_{1}(g)|}\sum_{\stackrel{A\in I_1(g)\colon}{|I_1(p_{A})| > 0}} \frac{\alpha(p_A) + 1}{2^{|I_{1}(p_A)|}}
    - 2\sum_{A \in I_{0}(g)} |I_{1}(p_A)|
    - \sum_{A \in I_{0}(g)} \beta(|p_{A}^{-1}(1)|),
    \end{align*}
    satisfies all properties \cref{prop:detection} - \cref{prop:early_bias}, where $\alpha$ is monotonically increasing with $|p_{A}^{-1}(1)|$, satisfies early bias (in the sense of \cref{prop:early_bias}) and $\codom (\alpha) \subset [0,1)$, and $\beta$ is monotonically increasing in $|p_{A}^{-1}(1)|$ with $\codom (\beta) \subset [0,1]$.
\end{theorem}
\begin{proof}
    \item
    Let $g\colon [n] \rightarrow \{0,1\}^n$ be the ground truth and $p,q \colon [n] \rightarrow \{0,1\}^n$ predictions.
    \paragraph{\cref{prop:detection}} 
    Let $W \in I_1(g)$, $q_{[n] \setminus W} = p_{[n] \setminus W}$, $p_W = 0$, and $|q_W^{-1}(1)| > 0$.
    Then, $m(g, q)$ and $m(g,p)$ differ only in the term 
    \[
        \sum_{\stackrel{W_1\in I_1(g)\colon}{|I_1(p_{W_1})| > 0}} \frac{\alpha(q_{W_1}) + 1}{2^{|I_1(q_{W_1})|}} = \sum_{\stackrel{W_1\in I_1(g)\colon}{|I_1(p_{W_1})| > 0}} \frac{\alpha(p_{W_1}) + 1}{2^{|I_1(p_{W_1})|}} + 
        \underbrace{\frac{\alpha(p_{W}) + 1}{2^{|I_1(p_{W})|}}}_{> 0}.
    \]
    Thus, it holds $m(g, p) < m(g, q)$.
    
    \paragraph{\cref{prop:alarms}} 
    Let $W \in I_1(g)$, and let $p_W(i) = 0$ for all $i \in I \subset W$, $|I_1(q_W)| = |I_1(p_W)| + 1$, $\min\limits_{i \in I} i > \max\limits_{j \in w\colon p(j) = 1} j$, and $q(i) = \mathds{1}_{I}(i) + p(i)\mathds{1}_{[n] \setminus I}(i)$.
    Then, $m(g, q)$ and $m(g,p)$ differ only in the term 
    \[
        \frac{\alpha(q_{W}) + 1}{2^{|I_1(q_{W})|}} 
        < \frac{1 + 1}{2^{|I_1(q_{W})|}}
        = \frac{1}{2^{|I_1(p_{W})|}}
        \leq \frac{\alpha(p_{W}) + 1}{2^{|I_1(p_{W})|}}.
    \]
    Thus, it holds $m(g, p) < m(g, q)$
    
    \paragraph{\cref{prop:false_positives}} 
    Let $W \in I_0(g)$, $q_{[n] \setminus \{i^*\}} = p_{[n] \setminus \{i^*\}}$ for some $i^* \in W$, $q(i^*) \neq p(i^*) = 0$, and $|I_1(q_W)| = |I_1(p_W)|$.
    %Let $p_{[n] \setminus \{i^*\}}' = p_{[n] \setminus \{i^*\}}$, $q(i^*) \neq p(i^*) = 0$, and $|I_1(q)| = |I_1(p)|$.
    Then, $m(g, q)$ and $m(g,p)$ differ only in the term
    \[
        \left( 2|I_1(q_{W_0})| + \beta(|q_{W_0}^{-1}(1)|) \right) > \left( 2|I_1(p_{W_0})| + \beta(|p_{W_0}^{-1}(1)|) \right),
    \]
    since $\beta(|q_{W_0}^{-1}(1)|) > \beta(|p_{W_0}^{-1}(1)|)$.
    Thus, it holds $m(g, p) > m(g, q)$.
    
    \paragraph{\cref{prop:false_positive_alarms}}
    Let $W \in I_0(g)$, $q_{[n] \setminus W} = p_{[n] \setminus W}$, and $|I_1(q_W)| < |I_1(p_W)|$.
    Then, $m(g, q)$ and $m(g,p)$ differ only in the term 
    \[
        \left( 2|I_1(q_{W_0})| + \beta(|q_{W_0}^{-1}(1)|)\right) 
        \leq \left( 2|I_1(q_{W_0})| + 1\right)
        < \left( 2(|I_1(q_{W_0})| + 1))\right)
        \leq \left( 2|I_1(p_{W_0})|\right)
        \leq \left( 2|I_1(p_{W_0})| + \beta(|p_{W_0}^{-1}(1)|)\right).
    \]
    Thus, it holds $m(g, p) < m(g, q)$.
    
    \paragraph{\cref{prop:permutations}}
    Let $W \in I_0(g)$, $q_{[n] \setminus W} = p_{[n] \setminus W}$, $|q_W^{-1}(1)| = |p_W^{-1}(1)|$, and $|I_1(q_W)| = |I_1(p_W)|$.
    Then, $m(g, q)$ and $m(g,p)$ can potentially differ only in the term
    \[
        \left( 2|I_1(q_{W})| + \beta(|q_{W}^{-1}(1)|) \right)
    \]
    and thus, since $\beta(|q_{W_0}^{-1}(1)|) = \beta(|p_{W_0}^{-1}(1)|)$, it holds $m(g, p) = m(g, q)$.
    
    \paragraph{\cref{prop:trust}}
    Let $W \in I_1(g)$, $\hat{W} \in I_0(g)$, $|I_1(q_{W})| = |I_1(p_{W})|$, $q_{[n] \setminus (W \cup \hat{W})} = p_{[n] \setminus (W \cup \hat{W})}$, $p_{\hat{W}} = 0$, and $q_{\hat{W}} = \mathds{1}_{\{i^*\}}$ for some $i^* \in \hat{W}$.
    It holds
    \[
        0 < \frac{1}{|I_1(g)|}\sum_{\stackrel{W_1\in I_1(g)\colon}{|I_1(p_{W_1})| > 0}} \frac{\alpha(p_{W_1}) + 1}{2^{|I_1(p_{W_1})|}}
        < \frac{1}{|I_1(g)|}\sum_{\stackrel{W_1\in I_1(g)\colon}{|I_1(p_{W_1})| > 0}} \frac{1 + 1}{2^{|I_1(p_{W_1})|}}
        \leq \frac{1}{|I_1(g)|}\sum_{\stackrel{W_1\in I_1(g)\colon}{|I_1(p_{W_1})| > 0}} 2
        \leq 2.
    \]
    This implies 
    \[
        0 < |\frac{1}{|I_1(g)|}\sum_{\stackrel{W_1\in I_1(g)\colon}{|I_1(p_{W_1})| > 0}} \frac{\alpha(p_{W_1}) + 1}{2^{|I_1(p_{W_1})|}} - \frac{1}{|I_1(g)|}\sum_{\stackrel{W_1\in I_1(g)\colon}{|I_1(q_{W_1})| > 0}} \frac{\alpha(q_{W_1}) + 1}{2^{|I_1(q_{W_1})|}}| < 2.
    \]
    Additionally, it holds $|I_1(p_{\hat{W}})| = 0$ and $|I_1(q_{\hat{W}})| = 1$.
    The second term in the definition only differs in the term 
    \[
        \left( 2|I_1(q_{\hat{W}})| + \beta(1) \right)
        = \left( 2 + \beta(1) \right)
        = 2 + \beta(1)
        > 2 + \beta(0)
%        > B\beta(0)
%        = B\left( 2|I_1(q_{\hat{W}})| + \beta(0) \right).
        = 2 + \left( 2|I_1(q_{\hat{W}})| + \beta(0) \right).
    \]
    This implies 
    \[
        - \sum_{\stackrel{W_0\in I_1(|g-1|)\colon}{|I_1(q_{W_0})| > 0}} \left( 2|I_1(q_{W_0})| + \beta(|q_{W_0}^{-1}(1)|) \right) 
        < - \sum_{\stackrel{W_0\in I_1(|g-1|)\colon}{|I_1(p_{W_0})| > 0}} \left( 2|I_1(p_{W_0})| + \beta(|p_{W_0}^{-1}(1)|) \right) - 2.
    \]
    Therefore it follows that $m(g, p) > m(g, q)$.
    
    \paragraph{\cref{prop:prediction_size}}
    Let $W \in I_1(g)$, $q_{[n] \setminus \{i^*\}} = p_{[n] \setminus \{i^*\}}$ for some $i^* \in W$, $q(i^*) \neq p(i^*) = 0$, and $|I_1(q_W)| \leq |I_1(p_W)|$.
    Then, $m(g, q)$ and $m(g,p)$ differ only in the term 
    \[
        \frac{1}{|I_1(g)|}\frac{\alpha(q_{W}) + 1}{2^{|I_1(q_{W})|}}
        \geq \frac{1}{|I_1(g)|}\frac{\alpha(q_{W}) + 1}{2^{|I_1(p_{W})|}}
        > \frac{1}{|I_1(g)|}\frac{\alpha(p_{W}) + 1}{2^{|I_1(q_{W})|}}
    \]
    Thus, it holds $m(g, p) < m(g, q)$.
    
    \paragraph{\cref{prop:temporal_order}}
    Let $W \in I_1(g)$, $q_{[n] \setminus W} = p_{[n] \setminus W}$, $|I_1(q_W)| = |I_1(p_W)|$, $|q^{-1}(1)| = |p^{-1}(1)|$, and $\min\limits_{i \in W\colon p(i) = 1} i < \min\limits_{i \in W\colon q(i) = 1} i$. 
    Then $m(g, q)$ and $m(g,p)$ differ only in the term 
    \[
        \frac{1}{|I_1(g)|}\frac{\alpha(q_{W}) + 1}{2^{|I_1(q_{W})|}}
        = \frac{1}{|I_1(g)|}\frac{\alpha(q_{W}) + 1}{2^{|I_1(p_{W})|}}
        < \frac{1}{|I_1(g)|}\frac{\alpha(p_{W}) + 1}{2^{|I_1(q_{W})|}}
    \]
    Thus, it holds $m(g, p) > m(g, q)$.
    
    \paragraph{\cref{prop:early_bias}}
    Let $W \in I_1(g)$, $q_{[n] \setminus W} = p_{[n] \setminus W}$, $|I_1(q_W)| \leq |I_1(p_W)|$, let $i^{*}\in W$ with $p(i^*) = 1$, let $i^{**} < i^* \in W$ with $p(i^{**}) = 0$, and $p_W'(i) = \mathds{1}_{\{i^{**}\}}(i) + p(i) \mathds{1}_{W\setminus \{i^*, i^{**}\}}(i)$.
    Then $m(g, q)$ and $m(g,p)$ differ only in the term 
    \[
        \frac{1}{|I_1(g)|}\frac{\alpha(q_{W}) + 1}{2^{|I_1(q_{W})|}}
        \geq \frac{1}{|I_1(g)|}\frac{\alpha(q_{W}) + 1}{2^{|I_1(p_{W})|}}
        > \frac{1}{|I_1(g)|}\frac{\alpha(p_{W}) + 1}{2^{|I_1(p_{W})|}}
    \]
    Thus, it holds $m(g, p) > m(g, q)$.
\end{proof}

\newpage
\section{Advanced Properties}\label{appendix:advanced_props}
In this section, we formalize the additional requirements discussed in \cref{sec:advanced_props} and prove \cref{theorem:advanced}.
Note, that most proofs can be easily adapted from \cref{appendix:metric}.

\subsection{Definitions}

Let $S_{V} = \{s\colon [n] \rightarrow V \;|\; n \in \mathbb{N}\}$ be the set of all sequences of finite length with values in $V \subset \mathbb{Z}$.
We define the set of all maximal intervals in the domain of a sequence $s \in S$ where the sequence takes a value $v \in V$ as the collection $I_{v}(s)$, or formally
$
    I_{v}(s) = \{[l,u]\subset s^{-1}(v) \;|\; s_{[l,u]} = v \land \nexists [l,u] \subsetneq [\hat{l}, \hat{u}] \subset s^{-1}(v) \colon s_{[\hat{l},\hat{u}]} = v\}.$
Furthermore, we define the set of maximal connected intervals where a sequence $s \in S$ takes values $u,v \in V$ respectively as $I_{uv}(s)$, or formally
$
    I_{uv}(s) = \{[l, b] \sqcup [b+1, u] \subset s^{-1}(\{u,v\}) | [l,b] \in I_{u}(s) \land [b+1, u] \in I_{v}(s)\}, 
$
i.e. the images of elements in $I_{uv}(s)$ of $s$ are of the form $[uuuuuuuvvvvvvvv]$.

We consider two binary sequences of the same length $g, s \in S_{\{0,1\}}$.
An element in $I_{1}(s)$ or $I_{1}(g)$ is called an \textbf{alarm} or \textbf{ground-truth alarm}.
An \textbf{early alarm} is an alarm $A \in I_{1}(s)$ that is raised before a ground-truth alarm and persists beyond its onset.
We Define the set of all early alarms as 
$
    EA(s,g) = \{A \in I_{1}(s_{I}) | I \in I_{01}(g) \land g_{A}\text{ surjective}\}.
$
Similarly, a \textbf{late alarm} is an alarm that is raised after the ground-truth alarm begins and remains active beyond its end.
We define the set of late alarms as $
    LA(s,g) = \{A \in I_{1}(s_{I}) | I \in I_{10}(g) \land g_{A}\text{ surjective}\}.
$
A \textbf{true false alarm} does not overlap with any ground-truth alarm, and can be defined as 
$
    TA(s,g) = \{A \in I_{1}(s) | g_A = 0 \land \nexists A' \in LA(s,g) \cup EA(s,g)\colon A \subset A'\}.
$
Lastly, we define the set of all detected anomalies as 
$
    DA(s,g) = \{A \in I_{1}(g) | \exists A_s \in I_{1}(s) \colon A_s \cap A \neq \emptyset \land (\min A_s \in A \lor g_{[\min A_s, \min A)} = 0)\} 
$

\subsection{Advanced Properties}
In this section we revisit the aspects discussed in \cref{sec:properties_definition} using this advanced framework. 
The differentiation of false alarms implies that changes in predictions within ground-truth anomalies can influence the number of false alarms in the predictions. 
As a consequence, we have to be particularly careful when specifying properties.
\paragraph{Detecting anomalies} The general intuition is the same as in \cref{prop:detection}.
There, it suffices to consider changes within one anomaly window.
In this setting, however, changes in the predictions within one ground-truth anomaly window can result in new early or late false alarms.
In such a case, the number of detected anomalies could stay the same, or, if both co-occur, the number of detected anomalies could decrease.
Thus, we make sure to exclude these corner cases with appropriate restrictions.
\begin{property}[Detection of Anomalies]\label{aprop:detection} Let $g$ be the ground truth of length $n$ and $A\in I_1(g)$. 
Let $p, q$ be predictions such that  $p_{[n]\setminus A} = q_{[n]\setminus A}$, $DA(p,g) = DA(q, g) \sqcup \{A\}$, and for all $A_O \in EA(p,g) \cup LA(p,g)\colon A_O \cap A \neq \emptyset$ it holds $A_O \cap g^{-1}(0) \in TA(q,g)$. 
Then $m(p,g) > m(q, g)$.
\end{property}

\paragraph{Additional Alarms} In general, the intuition is the same as in \cref{prop:alarms}. 
Excess alarms after detecting an anomaly result in additional costs and should therefore result in worse predictions. 
However, if an anomaly was only detected by a long early alarm, a response system could easily disregard that alarm as a false alarm. 
In that case, an additional alarm raised inside the anomaly window might be better.
This is a delicate trade-off.
Therefore, we exclude it explicitly in the following property by requiring at least one predicted window inside a ground-truth window to be present in the reference predictions.

\begin{property}[Redundant Alarms]\label{aprop:repeated_alarms} 
Let $g$ be a ground truth of length $n$ and $A\in DA(p,g)$. 
Let $p,q$ be predictions with $DA(p,g) = DA(q,g)$, $q(i) = \mathds{1}_{i\in I}+p(i)\mathds{1}_{i\not\in I}$ for some $I \subset A$ with $\min I > \max \{i \in A\colon p(i) = 1\}$, $p_I = 0$, $|I_1(q_{A})| = |I_1(p_{A})| + 1$, and $\exists B \in I_1(p)$ with $B \subset A$.
Then $m(p,g) > m(q, g)$.
\end{property}

\paragraph{Falsely Predicting Anomalies} 
Making false positive predictions, time steps where a method incorrectly predicts an anomaly where there is none, should generally decrease a method’s ranking as it decreases the quality of the prediction.
Similar to \cref{prop:false_positives}, we first consider the setting, where the number of alarms stays the same. 
The main difference is that we need to consider $I_1(p)$ instead of $I_1(p_N)$, to account for possible early or late alarms.
\begin{property}[Minimizing false positives]\label{aprop:precise} Let $g$ be a ground truth of length $n$ and $N \in I_0(g)$. 
Let $p,q$ be predictions with $p_{[n] \setminus \{i^*\}} = q_{[n] \setminus \{i^*\}}$, $q(i^*) = 1 \neq 0 = p(i^*)$ for some $i^* \in N$, and $|I_{1}(p)| \leq |I_{1}(q)|$.
Then $m(p,g) > m(q, g)$.
\end{property}
Every false alarm incurs an additional cost. 
Therefore, a method that makes the same number of false predictions as another method but raises fewer false alarms should be ranked higher. 
In contrast to \cref{aprop:false_alarms}, we need to consider changes in true false alarms, early alarms, and late alarms, since changing predictions inside a ground-truth normal window could reduce the number of one while increasing the number of another. 
\begin{property}[minimizing false alarms]\label{aprop:false_alarms} 
    Let $g$ be a ground truth of length $n$ and $N \in I_0(g)$. 
    Let $p,q$ be the predictions such that $p_{[n] \setminus N} = q_{[n] \setminus N}$, $DA(p,g) = DA(q, g)$, $|p^{-1}(1)| = |q^{-1}(1)|$, $|TA(p, g)| \leq |TA(q, g)|$, $|EA(p, g)| \leq |EA(q, g)|$, $|LA(p, g)| \leq |LA(q, g)|$, and $|TA(p, g)| + |EA(p, g)| + |LA(p, g)| < |TA(q, g)| + |EA(q, g)| + |LA(q, g)|$.
    Then $m(p,g) > m(q, g)$.
\end{property}
Following the motivation for \cref{prop:permutations}, timing of true false alarms should not impact the ranking. 
However, the same reasoning does not apply to early and late alarms, as they are inherently connected to ground-truth anomaly windows.
\begin{property}[invariance under permutations of true false alarms]\label{aprop:false_positive_permutation} Let $g$ be a ground truth of length $n$ and $p,q$ be the predictions, such that $p_{g=1} = q_{g=1}$, and $|p^{-1}(1)| = |q^{-1}(1)|$, $EA(p, g) = EA(q, g), LA(p, g) = LA(q, g), |TA(p, g)| = |TA(q, g)|$. Then $m(p,g) = m(q, g)$.
\end{property}
\cref{prop:trust} formalizes a trade-off that imposes strong restrictions on false alarms, which might not apply to all settings equally.
Considering the differentiation of alarm types in the advanced setting, we replace this property with another that considers the relative importance of these different types.
On the one hand, early alarms can have detrimental effects on the performance, since a response system could disregard a predicted anomaly as a false positive when enough time transpires until the anomaly actually manifests in the data.
In this case a technically detected anomaly could be easily missed.
Therefore, we argue that an early alarm is worse than a true false alarm.
Note, that we consider a greatly restricted setting, where we generally assume no differences outside of the actual false positives, which additionally results in a perfectly predicted anomaly for the prediction with the true false alarm.
On the other hand, late alarms are inherently connected to an anomaly that was correctly detected.
Therefore, we argue that late alarms are generally preferred over true false alarms, as they raise no additional false alarm that is not related to a ground-truth anomaly.
Further cementing this ranking is the fact that it is often much easier to pin-point the exact start of an anomaly window because of a thorough investigation of its root cause or because it was known in advance, but it is often difficult to identify the exact time when the system has fully recovered from all anomalous effects.
\begin{property}[Weighing different types of alarms]\label{aprop:false_alarm_type} 
Let $g$ be a ground truth of length $n$ and $p,q$ be the predictions, such that $|p^{-1}(1)| = |q^{-1}(1)|$, and  $DA(p,g) = DA(q, g)$. 
Then $m(p,g) > m(q,g)$, if either of the following holds.
\begin{itemize}
    \item[(i)] Let $A \in EA(q,g)$, $A_E = A \cap g^{-1}(0)$, $A_T \in TA(p,g)$ such that $A_E \cap A_T = \emptyset$, $p_{[n] \setminus A_E \cup A_T} = q_{[n] \setminus A_E \cup A_T}$, $p_{A_E} = 0$, and $q_{A_T} = 0$. 
    \item[(ii)] Let $A_T \in TA(q,g)$, $A \in LA(p,g)$, $A_L = A \cap g^{-1}(0)$ such that $A_T \cap A_L = \emptyset$, $p_{[n] \setminus A_T \cup A_L} = q_{[n] \setminus A_T \cup A_L}$, $p_{A_T} = 0$, and $q_{A_L} = 0$. 
\end{itemize}
\end{property}
\paragraph{Preciseness of predictions}
Similar to previous properties, we have to consider the potential trade-off with early alarms when adding more correct predictions when adapting \cref{aprop:monotonicity_true_positives} to the advanced setting. 
Furthermore, we need to consider how the set of detected anomalies could change by introducing an additional early or late alarm.
Excluding both trade-offs results in the following property.
\begin{property}[maximizing true positives]\label{aprop:monotonicity_true_positives}
    Let $A \in DA(p,g) \cap DA(q, g)$, and $i^* \in A$ such that $p_{[n] \setminus \{i^*\}} = q_{[n] \setminus \{i^*\}}$, $p(i^*) = 1 \neq 0 = q(i^*)$, and $EA(p,g) = EA(p,g)$.
    Then $m(p,g) > m(q, g)$.
\end{property}
\paragraph{Timing of alarms}
\cref{prop:temporal_order} and \cref{prop:early_bias} formalize the preference for early predictions. 
In the advanced setting, earlier predictions might result in longer early alarms or even in less detected anomalies.
Therefore, we additionally require that early alarms do not change when adapting \cref{prop:temporal_order} and \cref{prop:early_bias}.
Another trade-off to consider is the potential loss of late alarms.
\begin{property}[alarm timing]\label{aprop:temporal_order}
Let $A \in DA(p,g) \cap DA(q, g)$ such that $p_{[n] \setminus A} = q_{[n] \setminus A}$, $|I_1(p_{A})| = |I_1(q_{A})|$, $|p^{-1}(1)| = |q^{-1}(1)|$, $EA(p,g) = EA(q, g)$, and $|LA(p,g)| = |LA(q, g)|$. For some $i \in A$ it holds that $\min \{i \colon p(i) = 1\} < \min \{i \colon q(i) = 1\}$.
Then $m(p,g) > m(q, g)$.
\end{property}
\begin{property}[Early bias]\label{aprop:early_bias}
    Let $g$ be a ground truth of length $n$ and $p,q$ be predictions such that $EA(p,g) = EA(q, g)$ and  $|LA(p,g)| = |LA(q, g)|$.
    Let $A \in DA(p,g) \cap DA(q, g)$ such that  $|I_1(p_A)| \leq |I_1(q_A)|$ and for some $i^{*} < i^{**}\in A$ it holds $q(i^{**}) = p(i^{*}) = 1 \neq 0 = q(i^{*}) = p(i^{**})$, $p_{[n] \setminus \{i^*, i^{**}\}} = q_{[n] \setminus \{i^*, i^{**}\}}$.
    Then $m(p,g) > m(q, g)$.
\end{property}

\subsection{Proof of \cref{theorem:advanced}}

\begin{lemma}\label{lemma:sum}
    Let $x \in [0,n)$, $y \in [0,1)$, and $n \in \mathbb{N}_{> 0}$. Then it holds
    \[
        \frac{x + y}{n + 1} - \frac{x}{n} > -\frac{1}{2}
    \]
\end{lemma}
\begin{proof}
    \begin{align*}
        \frac{x + y}{n + 1} - \frac{x}{n} 
        \geq \frac{nx}{n(n + 1)} - \frac{x(n + 1)}{n(n + 1)} 
        = \frac{- x}{n(n + 1)} 
        > \frac{-n}{n(n + 1)}
        = \frac{-1}{(n + 1)} \geq -\frac{1}{2}
    \end{align*}
\end{proof}

\begin{theorem}
    The ALARM score (\cref{theorem:advanced}) satisfies the properties \cref{aprop:detection} - \cref{aprop:early_bias}.
\end{theorem}
\begin{proof}
    \item
    \paragraph{\cref{aprop:detection}}
    By definition it holds $|DA(p,g)| = |DA(q,g)| + 1$.
    Since $p_{[n]\setminus A} = q_{[n]\setminus A}$, it holds that $\beta(|p_{g=0}^{-1}(1)|) = \beta(|q_{g=0}^{-1}(1)|)$.
    By the third assumption it follows that the number of late alarms and the number of early alarms can change at most by 1.
    With every increase, the number of true false alarms decreases.
    Thus, it follows that
    \begin{align*}
        |TA(q,g)| - 1 + \frac{3}{2}|EA(q,g)| + \frac{1}{2}(|LA(q,g)| + 1)
        &\leq |TA(p,g)| + \frac{3}{2}|EA(p,g)| + \frac{1}{2}|LA(p,g)|\\ 
        &\leq |TA(q,g)| - 1 + \frac{3}{2}(|EA(q,g)| + 1) + \frac{1}{2}|LA(q,g)|
    \end{align*}
    Thus, using \cref{lemma:sum} to bound the difference in the second term, it holds
    \begin{align*}
        m(p,g) - m(q, g) > 1 - \frac{1}{2} - \frac{1}{2t} \geq 0
    \end{align*}

    \paragraph{\cref{aprop:repeated_alarms}}
    By definition, it holds $|DA(p,g)| = |DA(q,g)|$, $\beta(|p_{g=0}^{-1}(1)|) = \beta(|q_{g=0}^{-1}(1)|)$, and $EA(p,g) = EA(q, g)$.
    Furthermore, it holds
    \[
        |TA(p,g)| - 1 + \frac{1}{2}(|LA(p,g)| + 1) \leq |TA(q,g)| + \frac{1}{2}|LA(q,g)| \leq |TA(p,g)| + \frac{1}{2}(|LA(p,g)| +1)
    \]
    and
    \[
        \frac{1 + \alpha(q_{A})}{2^{|I_{1}(q_A)|}} < \frac{2}{2^{|I_{1}(q_A)|}} = \frac{1}{2^{|I_{1}(p_A)|}} \leq \frac{1 + \alpha(p_{A})}{2^{|I_{1}(p_A)|}}
    \]
    and, thus, it holds
    \[
        m(p,g) - m(q, g) > 1 - \frac{1}{2t} \geq 0
    \]
    \paragraph{\cref{aprop:precise}}
    By definition, it holds $DA(p,g) = DA(q, g)$, and $\frac{1 + \alpha(q_{A})}{2^{|I_{1}(q_A)|}} = \frac{1 + \alpha(p_{A})}{2^{|I_{1}(p_A)|}}$ for all $A\in DA(p,g)$.
    From the second assumption, it follows that $\beta(|q_{g=0}^{-1}(1)|) > \beta(|p_{g=0}^{-1}(1)|)$ and from the last assumption it follows that $|TA(q,g)| \geq |TA(p,g)|$.
    Furthermore, it holds $EA(q, g) \geq EA(p, g)$ and $LA(q, g) \geq LA(p, g)$.
    Therefore, it follows that $m(p,g) > m(q, g)$.

    \paragraph{\cref{aprop:false_alarms}}
    By definition, the scores for $p$ and $q$ only differ in the term $m_{FA}$.
    There, it holds by the assumptions on the sum of alarms that this term of the score of $q$ is strictly less than the corresponding term in the score of $p$ due to the last assumption, resulting in the desired ordering.

    \paragraph{\cref{aprop:false_positive_permutation}}
    By definition, the first, second, and third to fourth terms are not affected by the changes between $p$ and $q$.
    Since
    \[
        |p^{-1}(1)| = |p^{-1}(1) \cap g^{-1}(0)| + |p^{-1}(1) \cap g^{-1}(1)| 
        = |p^{-1}(1) \cap g^{-1}(0)| + |q^{-1}(1) \cap g^{-1}(1)|
        = |q^{-1}(1)|
    \]
    it holds that $|p^{-1}(1) \cap g^{-1}(0)| = |q^{-1}(1) \cap g^{-1}(0)|$ and, therefore,
    $\beta(|q_{g=0}^{-1}(1)|) = \beta(|p_{g=0}^{-1}(1)|)$

    \paragraph{\cref{aprop:false_alarm_type} (i)}
    By definition, the first two terms and the last term are unaffected by the differences between $p$ and $q$.
    Furthermore, it holds $|EA(p,g)| + 1 = |EA(q,g)|$, $|TA(p,g)| = |TA(q,g)| + 1$, and $|LA(p,g)| \leq |LA(q,g)|$
    Therefore, it holds
    \begin{align*}
        m(p,g) - m(q, g) = &\frac{1}{t}\underbrace{(|TA(q,g)| - |TA(p,g)|)}_{= -1} + \frac{3}{2t}\underbrace{(|EA(q,g)| - |EA(p,g)|)}_{= 1}\\
        &+ \frac{1}{2t}\underbrace{(|LA(q,g)| - |LA(p,g)|)}_{\geq 0} \geq \frac{1}{2t} > 0
    \end{align*}

    \paragraph{\cref{aprop:false_alarm_type} (ii)}
    By definition, the first two terms, and the last term are unaffected by the differences between $p$ and $q$.
    By the same argument as for (i), it holds
    \begin{align*}
        m(p,g) - m(q, g) = &\frac{1}{t}\underbrace{(|TA(q,g)| - |TA(p,g)|)}_{= 1} + \frac{3}{2t}\underbrace{(|EA(q,g)| - |EA(p,g)|)}_{\geq 0}\\
        &+ \frac{1}{2t}\underbrace{(|LA(q,g)| - |LA(p,g)|)}_{= -1} \geq \frac{1}{2t} > 0
    \end{align*}

    \paragraph{\cref{aprop:monotonicity_true_positives}}
    By definition, only one term in the sum of the second term is affected by the differences between $p$ and $q$.
    Because of the monotonicity of $\alpha$, it holds,
    \[
        \frac{1 + \alpha(p_{A})}{2^{|I_{1}(p_A)|}} > \frac{1 + \alpha(q_A)}{2^{|I_{1} (p_A)|}} \geq \frac{1 + \alpha(q_A)}{2^{|I_{1} (q_A)|}}
    \]
    and, therefore, it holds $m(p,g) > m(q, g)$.

    \paragraph{\cref{aprop:temporal_order}}
    By definition, only one term in the sum of the second term is affected by the differences between $p$ and $q$.
    By the early bias assumption on $\alpha$, it holds
    \[
        \frac{1 + \alpha(p_{A})}{2^{|I_{1}(p_A)|}} > \frac{1 + \alpha(q_{A})}{2^{|I_{1}(q_A)|}}
    \]
    and, therefore, $m(p,g) > m(q, g)$.

    \paragraph{\cref{aprop:early_bias}}
    By definition, only one term in the sum of the second term is affected by the differences between $p$ and $q$.
    By the early bias assumption on $\alpha$, it holds
    \[
        \frac{1 + \alpha(p_{A})}{2^{|I_{1}(p_A)|}} > \frac{1 + \alpha(q_{A})}{2^{|I_{1}(q_A)|}}
    \]
    and, therefore, $m(p,g) > m(q, g)$.
    
\end{proof}

\newpage
\section{Evaluation}\label{appendix:evaluation}
In this section we present additional evaluations of (A)LARM illustrating the differences to other metrics and providing more intuition on the requirements discussed in the main paper. 
We use the SMD dataset \cite{su2019robust} and models trained thereon 
\cite{wagner2023timesead} as a basis for our experiments. 
Due to the differences in the conditions during the collection process, the SMD dataset can be considered a collection of 28 datasets \cite{wagner2023timesead}. 
Since there is no quantitative measure to quantify the quality of evaluation metrics, we discuss a selection of the full evaluation here in detail to illustrate the differences between evaluation metrics.  

\subsection{Ranking Variance}\label{appendix:ranking_variance}
\begin{figure}[h!]
    \centering
    \begin{subfigure}{0.47\textwidth}
    \includegraphics[width=\textwidth]{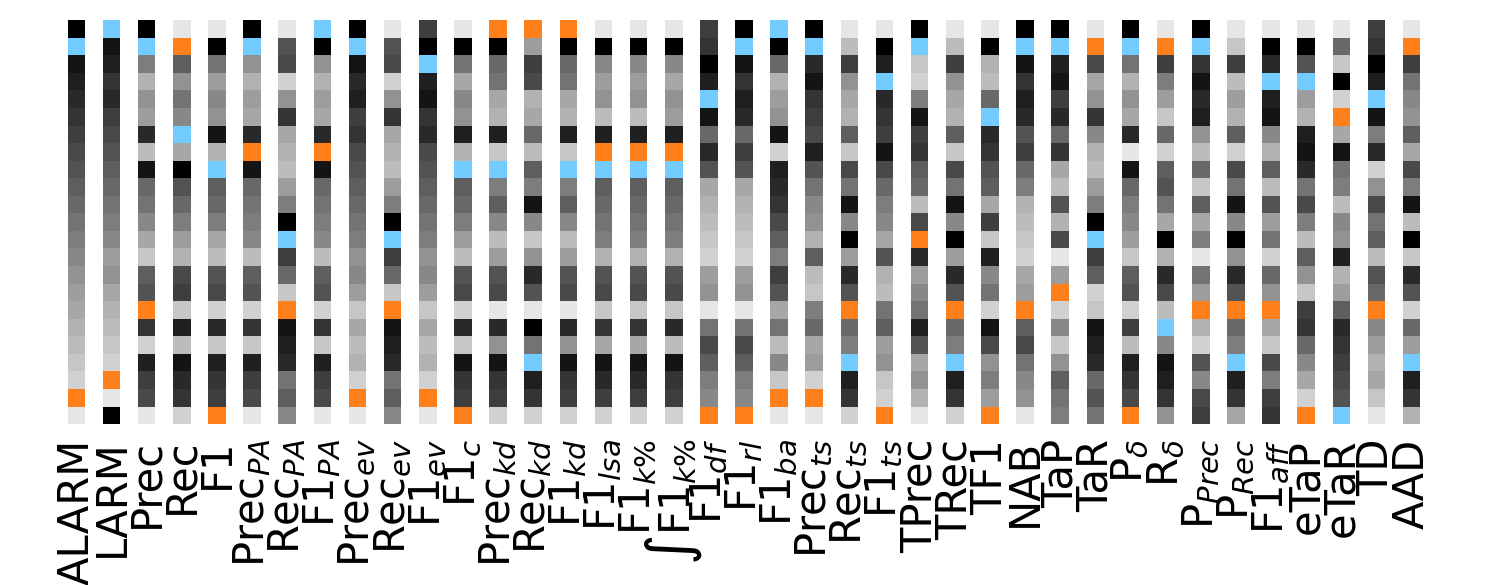}
    \caption*{\centering SMD1}
    \end{subfigure}
    \hspace{\horizontalbetweenfigures}
    \begin{subfigure}{0.47\textwidth}
    \includegraphics[width=\textwidth]{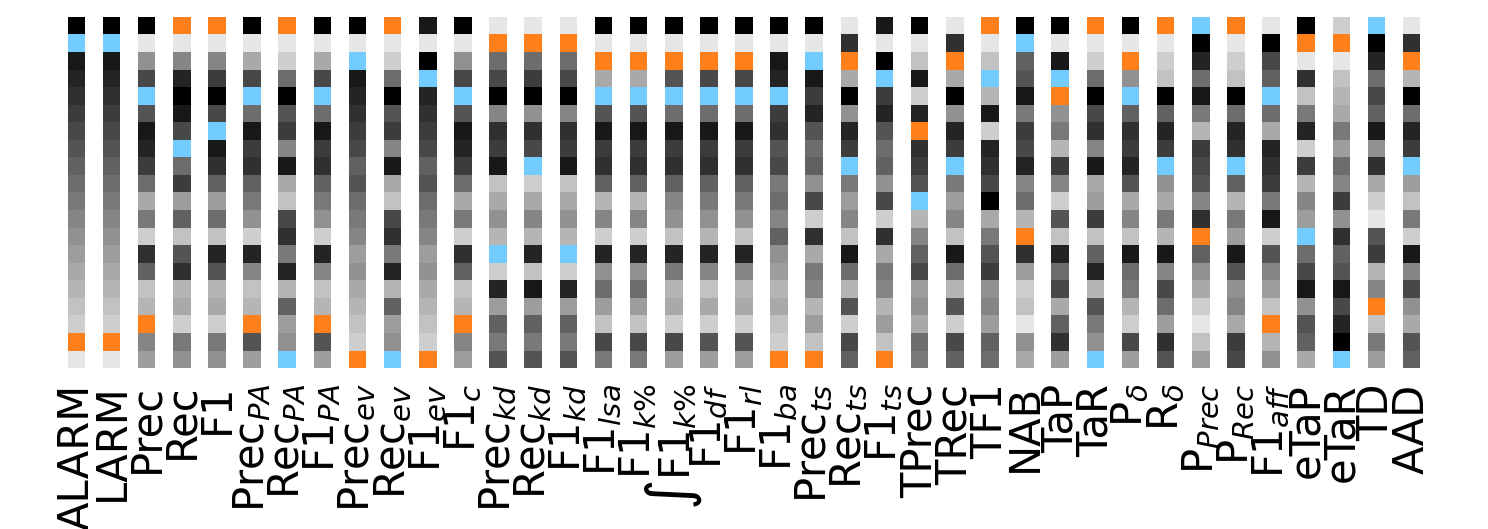}
    \caption*{\centering SMD2}
    \end{subfigure}
    \vspace{\verticalbetweenfigures}

    \begin{subfigure}{0.47\textwidth}
    \includegraphics[width=\textwidth]{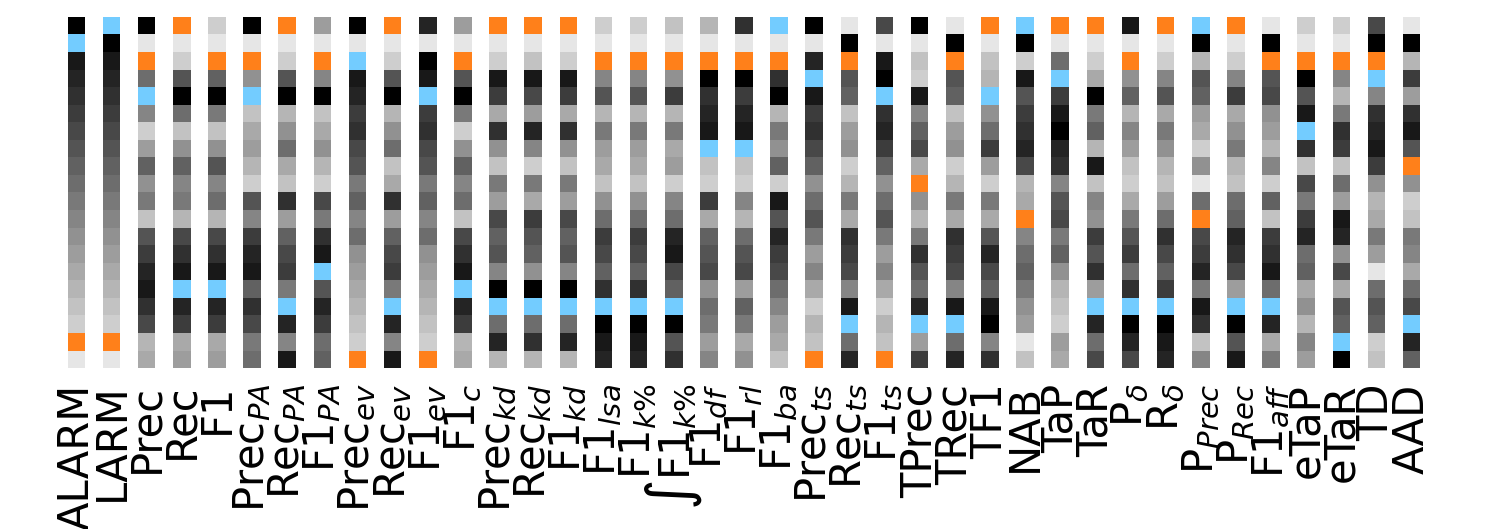}
    \caption*{\centering SMD3}
    \end{subfigure}
    \hspace{\horizontalbetweenfigures}
    \begin{subfigure}{0.47\textwidth}
    \includegraphics[width=\textwidth]{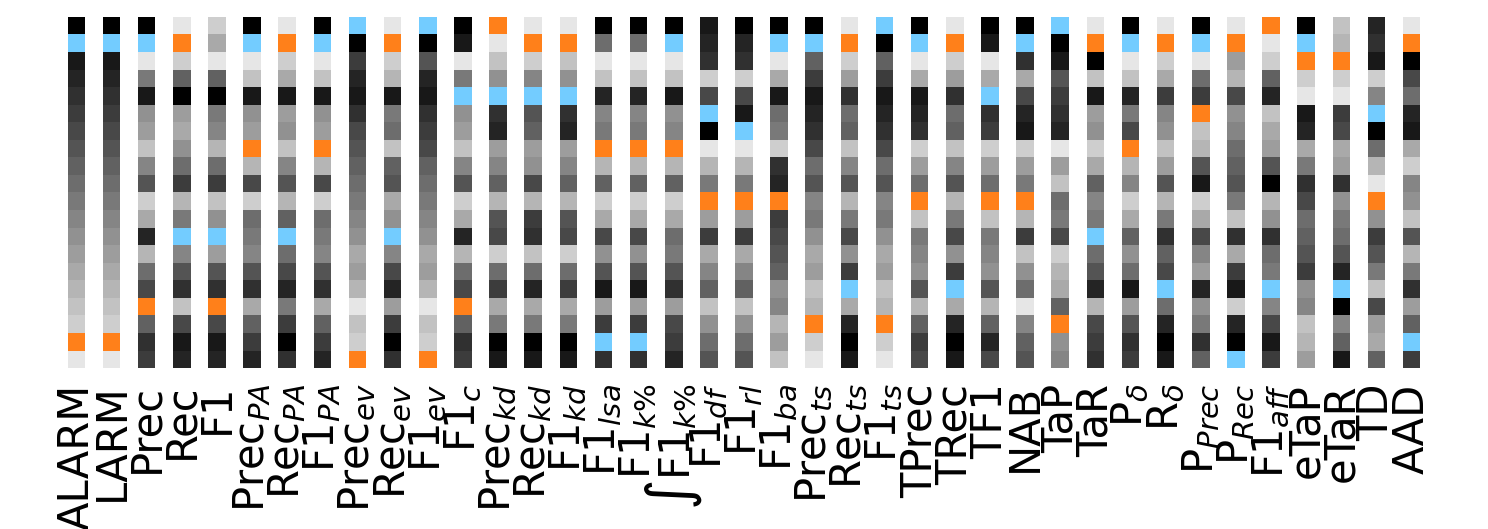}
    \caption*{\centering SMD4}
    \end{subfigure}
    \vspace{\verticalbetweenfigures}
    \caption{Comparison of the rankings of different evaluation metrics on eight SMD datasets show that (A)LARM provide significantly different rankings compared to other metrics on multiple datasets.}
    \label{fig:full_evaluation}
\end{figure}

To illustrate the high variance in rankings produced by different metrics, we take the predictions of the methods implemented and trained on SMD datasets in \cite{wagner2023timesead} and evaluate their performance with every metric considered in \cref{tab:metrics-properties} and the (A)LARM scores proposed in \cref{theorem:main} and \cref{theorem:advanced}.
We assign each method a unique color and order them according to their performance for each metric.
The results on more of the SMD datasets are shown in \cref{fig:full_evaluation}.
This visualization illustrates the significant differences between the rankings with respect to different metrics. 
Notably, no existing metric provides consistent rankings with the (A)LARM scores across different datasets. 
\subsection{Qualitative Evaluation of (A)LARM}

To illustrate the behavior of the metrics proposed in \cref{theorem:main} and \cref{theorem:advanced}, we compare the rankings produced by (A)LARM with ranking produced by the point-wise F1 score.
\begin{figure}[h!]
    \centering
    \includegraphics[width=0.49\linewidth]{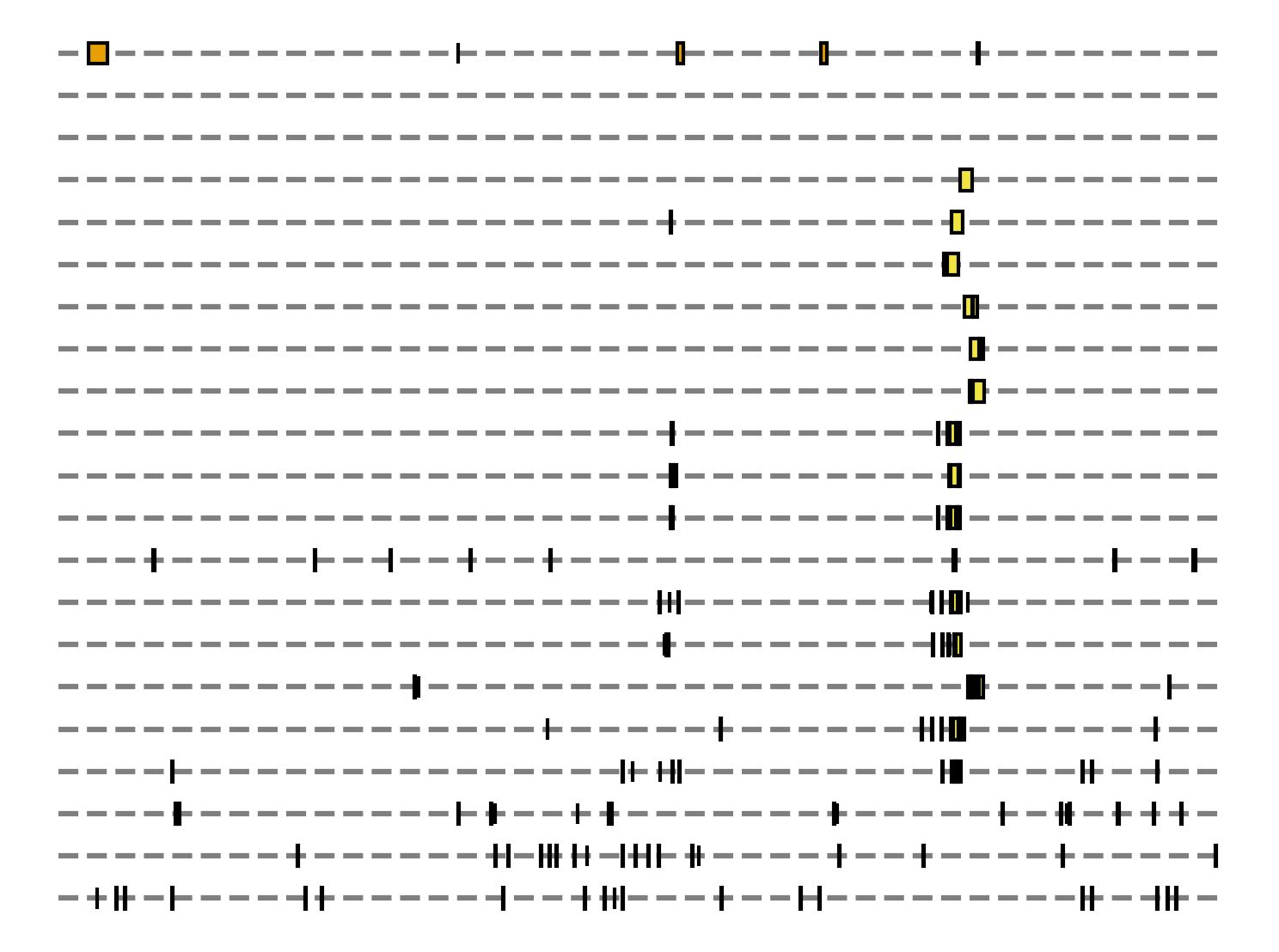}
    \includegraphics[width=0.49\linewidth]{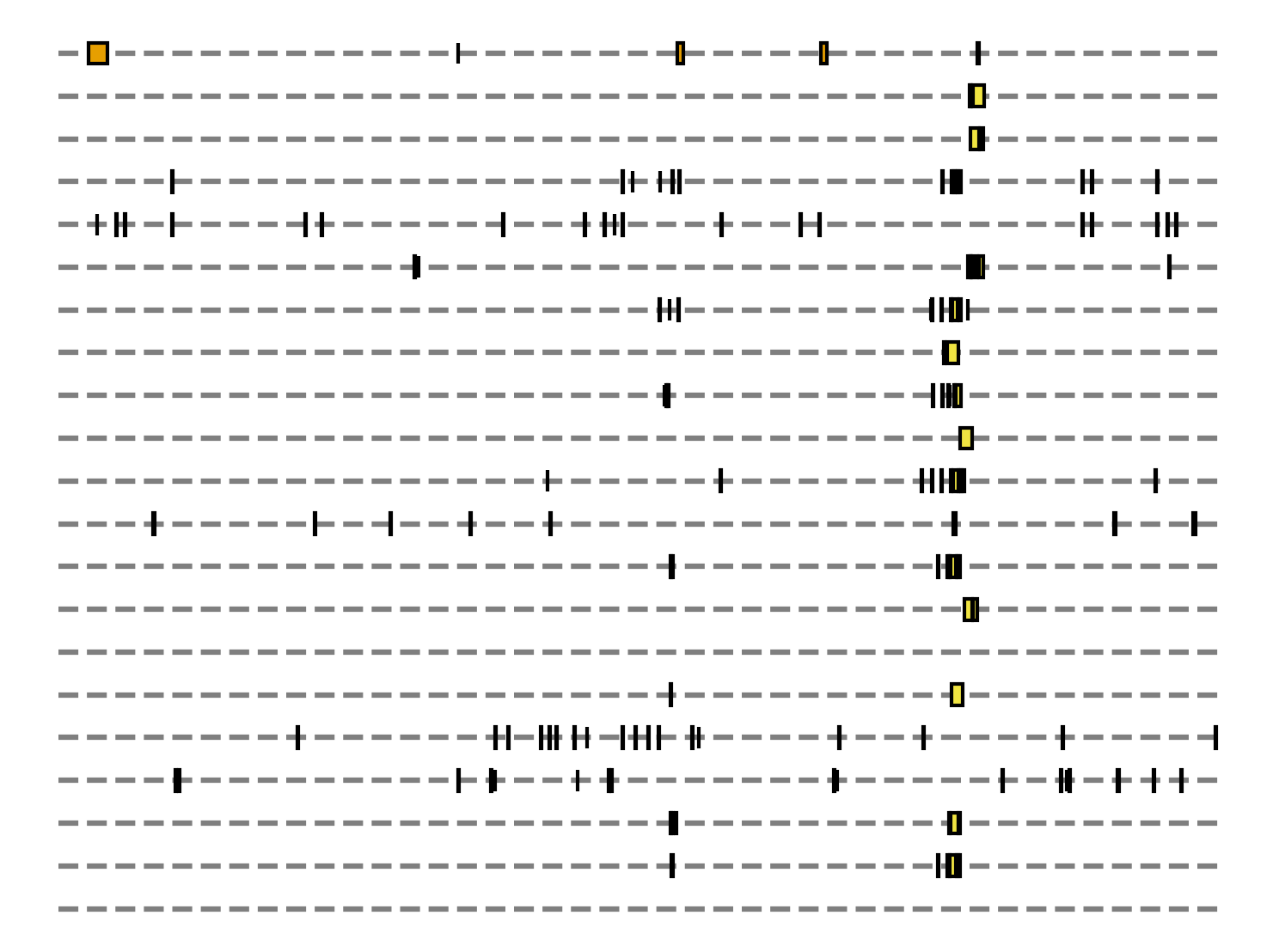}
    \caption{Comparison of LARM (left) and point-wise F1 score (right) on SMD8. The top row represents the ground truth. The methods are presented in descending order, with the top method ranked highest by each metric.}
    \label{fig:SMD8_ours_vs_pf1}
\end{figure}
In \cref{fig:SMD8_ours_vs_pf1}, we can observe the significant impact of false alarms on LARM.
Note that there appear to be multiple methods that make no predictions at all and that these predictions seem to appear in different positions in the rankings.
Due to the size of the dataset, we have to plot at a reduced resolution where small predictions are not visible.
Thus, we visualize a smaller subset of this particular dataset in \cref{fig:SMD8_cut_ours_vs_pf1}.
\begin{figure}[h!]
    \centering
    \includegraphics[width=0.49\linewidth]{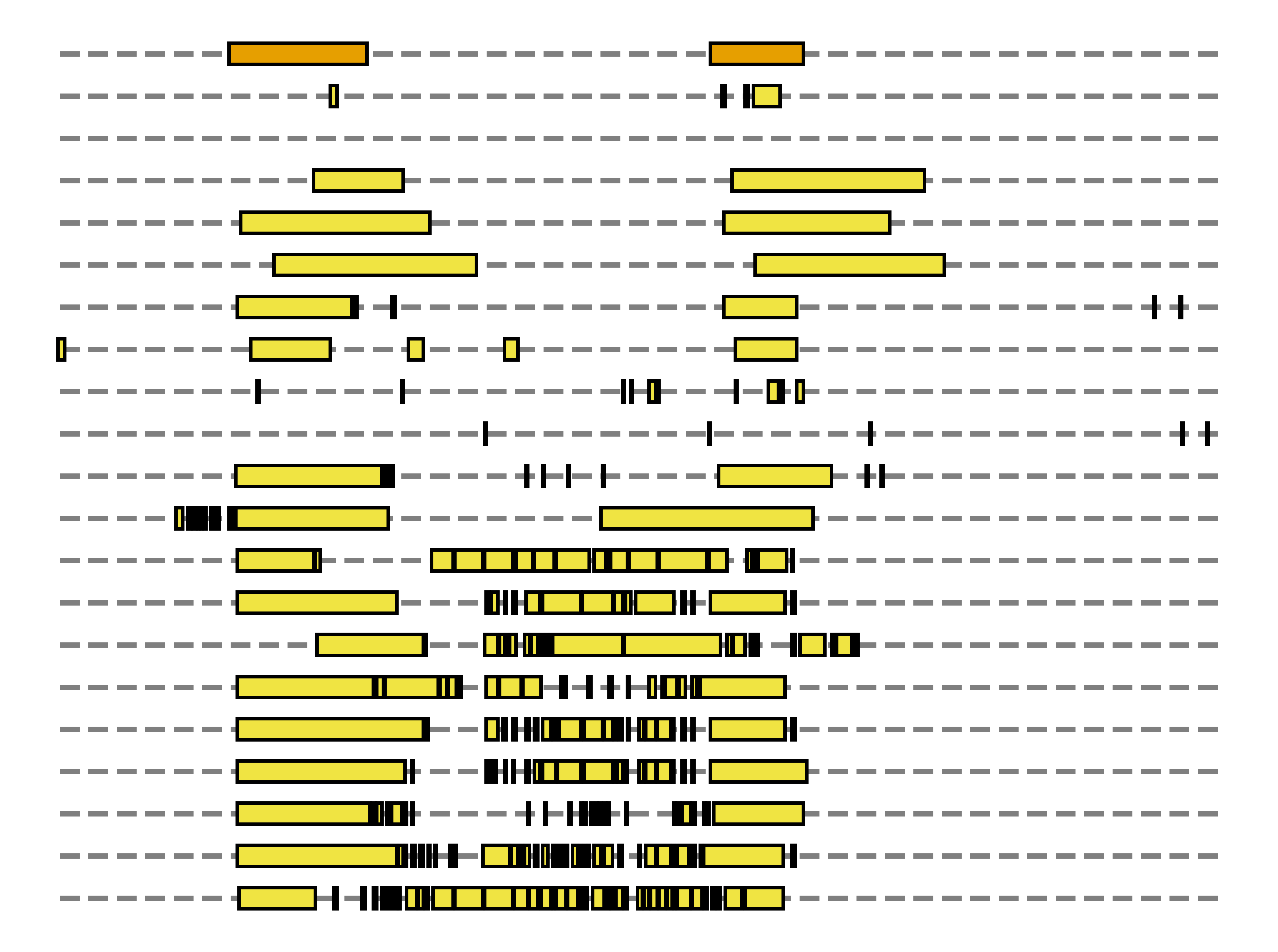}
    \includegraphics[width=0.49\linewidth]{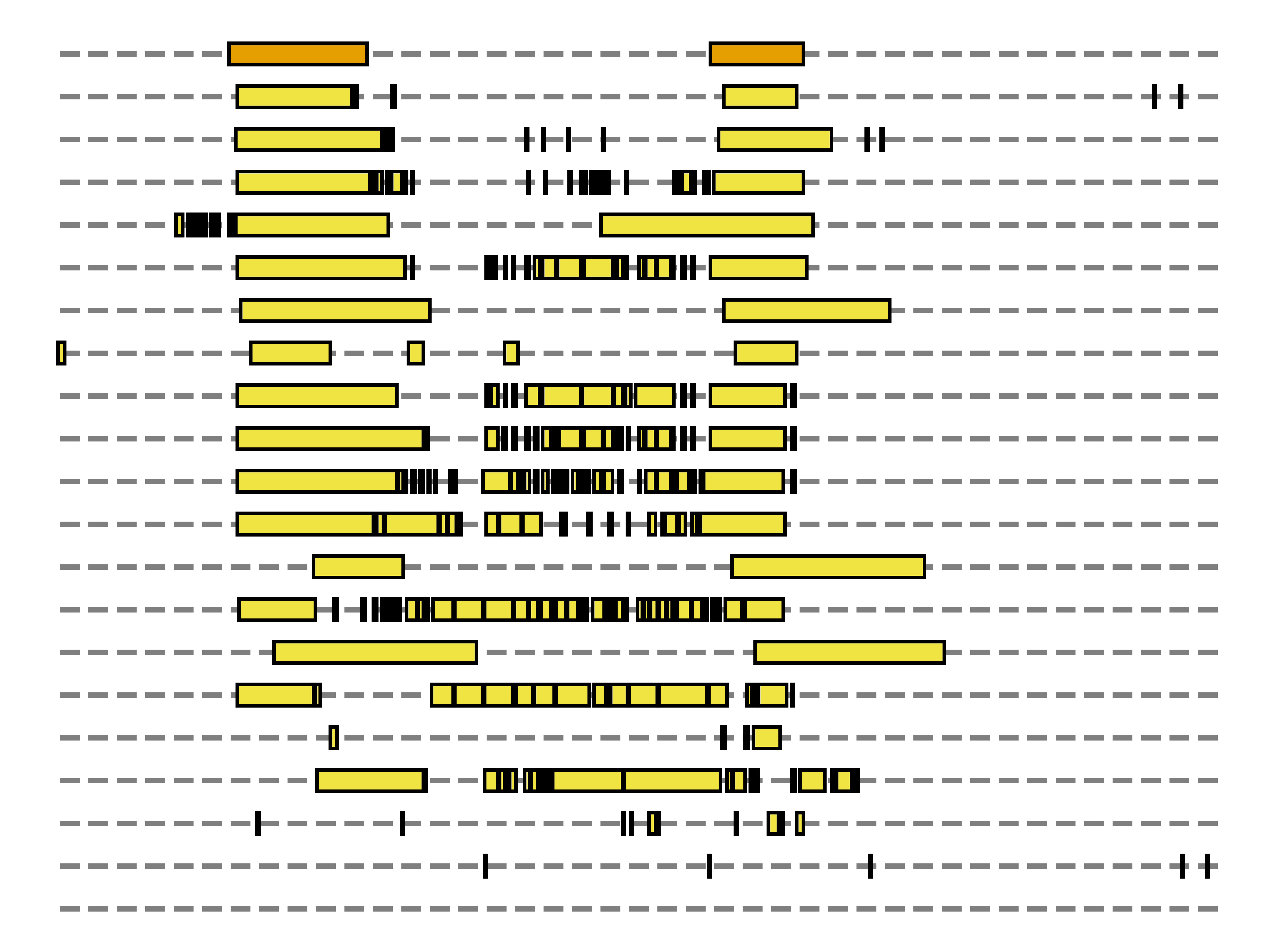}
    \caption{Comparison of LARM (left) and point-wise F1 score (right) on a subset of SMD8. The top row represents the ground truth. The methods are presented in descending order, with the top method ranked highest by each metric.}
    \label{fig:SMD8_cut_ours_vs_pf1}
\end{figure}
To illustrate the differences between rankings, we restrict the metric computation to this region only, which leaves one method with no predictions.
We can see that this method is ranked second with LARM and last with the point-wise F1 score.
This is a significant difference.
Since false positives have an overwhelming negative impact in LARM, it naturally follows that the score decreases with the number of false positives and, in particular, with the number of false alarms. 
We can see that the three methods ranked just below the method with no predictions
\begin{figure}[h!]
    \centering
    \includegraphics[width=0.5\linewidth]{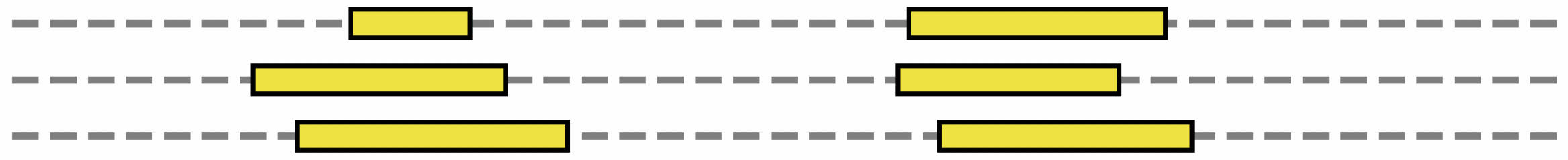}
    \caption{Three predictions of three different methods on a subset of SMD8 that all predict both anomalies at different times and all exceed the ground-truth labels. Due to these subtle differences on two different anomalies, these methods are difficult to rank.}
    \label{fig:simlar_smd_7}
\end{figure}
have almost the same score, each predicting both anomalies at different times and predicting anomalies for a longer time than indicated in the ground truth.
Since these methods exhibit very similar behavior with slight differences in the prediction delay and the number of false positives, they are naturally hard to distinguish.
Notably, these methods appear in different positions in the ranking with the F1-score.
Since we assume the labels are correct in \cref{ass:correct_labels} and that predictions are to be taken at face value in \cref{ass:predictions}, a response system or operator might need to investigate the predictions that exceed the labels an additional time. 
For example, we might know based on the response used to resolve the anomaly when it should end and based on the predictions would need to initiate additional investigations.
If these three methods would still represent desired behavior, it could imply that \cref{ass:correct_labels} is violated in this dataset.
In this case, we could either introduce more lenient labels, or we could adjust the metric to account for such cases.
For example, we could weaken \cref{prop:false_positive_alarms} and \cref{prop:trust} to allow for predictions that exceed the labels by modifying the sum in the second term, as for example in ALARM (\cref{theorem:advanced}).
This highlights the open design space for metrics based on (A)LARM and the need for precise specification of the desired properties. 
Based on our very strict requirements for reliability in LARM, the anomaly detection system might be disabled after the first false alarm, never giving it the chance to detect the second anomaly when a false positive occurs in between.
Thus, the ranking produced by the metric in \cref{theorem:main} aligns with our assumptions and requirements discussed in \cref{sec:assumptions} and \cref{sec:properties}.
This explains why the following method
\begin{figure}[h!]
    \centering
    \includegraphics[width=0.5\linewidth]{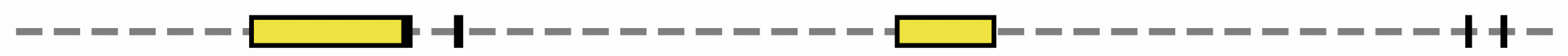}
\end{figure}
is clearly ranked lower with LARM, while it is ranked highest with the F1-score, since the F1-score is clearly not as sensitive to false alarms and values point-wise recall much higher.

Comparing the ranking of ALARM with LARM in the same setting shows that the three methods highlighted above are still ranked closely together. 
However, due to the tolerance for false alarms, ALARM ranks the method with no predictions lower.
\begin{figure}[h!]
    \centering
    \includegraphics[width=0.49\textwidth]{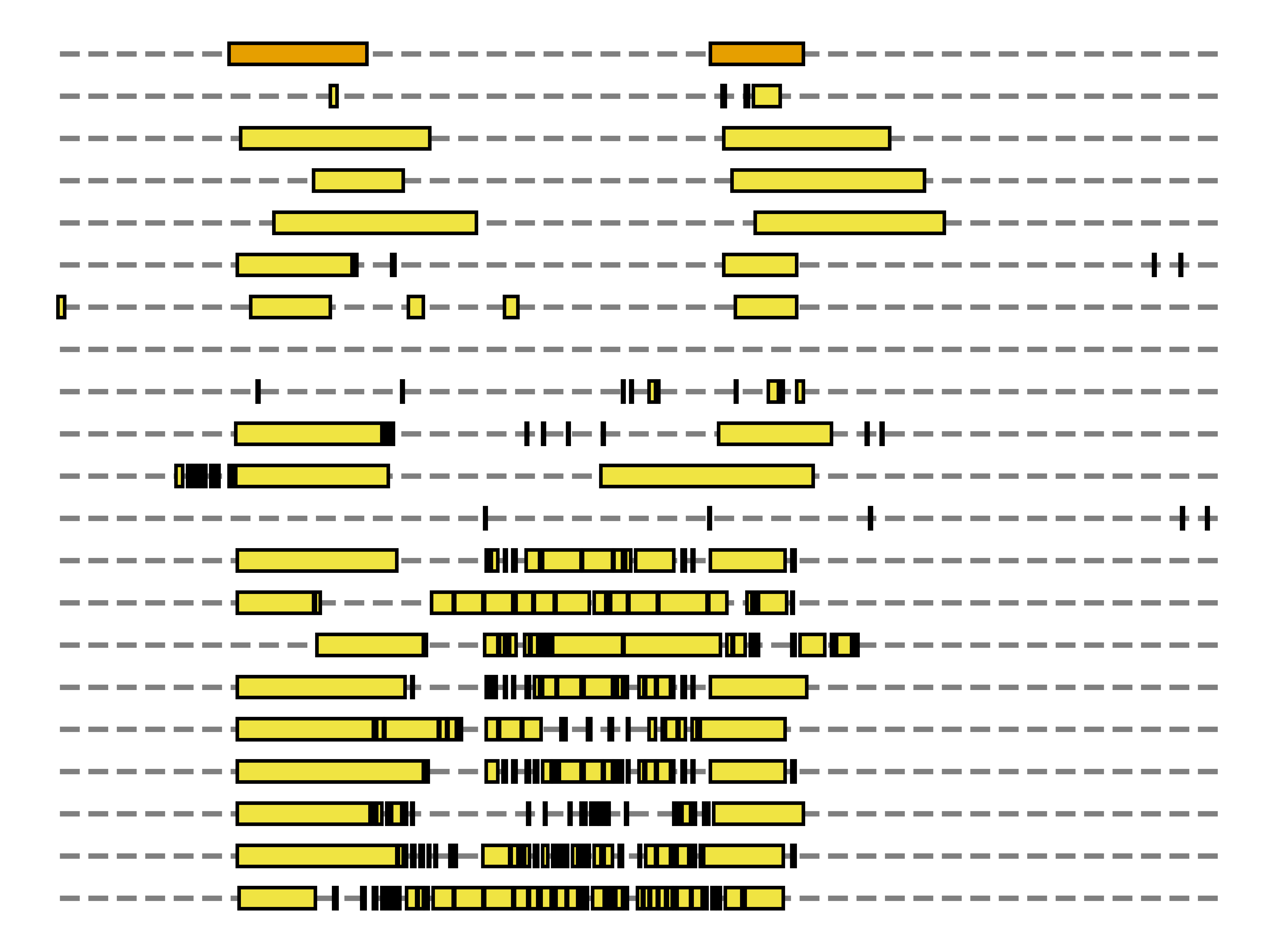}
    \includegraphics[width=0.49\linewidth]{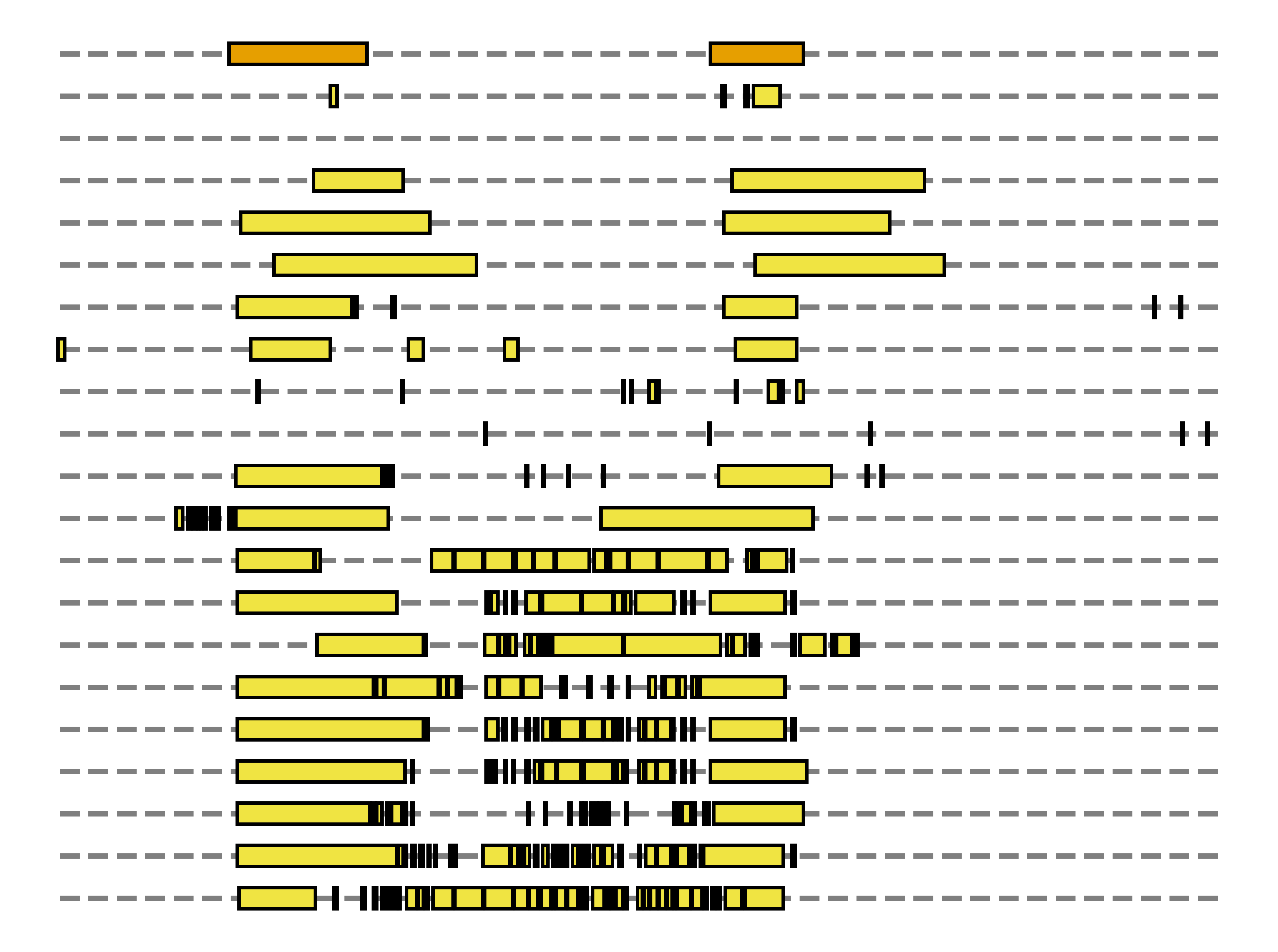}
    \caption{Comparison of ALARM (left) and LARM (right) on a subset of SMD8. The top row represents the ground truth. The methods are presented in descending order, with the top method ranked highest by each metric.}
    \label{fig:SMD8_cut_ours_vs_pf1}
\end{figure}
Noticeably, the methods in the bottom half of the rankings of both ALARM and LARM produce an increasing number of false alarms approaching almost random predictions under the assumption that the labels are correct (\cref{ass:correct_labels}).

\clearpage

\section{Proofs}\label{appendix:proofs}
\begin{theorem}\label{proof:point-wise_precision}
    \textbf{Point-wise Precision}
    defined as 
    \[
    Precision(g,p) = \frac{|p^{-1}(1) \cap g^{-1}(1)|}{|p^{-1}(1)|}
    \]
    satisfies property %\cref{prop:false_positives} and 
    \cref{prop:permutations}.
\end{theorem}
\begin{proof}
    \item
    Let $\colon [n] \rightarrow \{0,1\}^n$ be the ground truth and $p,\hat{p} \colon [n] \rightarrow \{0,1\}^n$ predictions.
    
    \paragraph{\cref{prop:detection}} 
    Consider the following example:
    \begin{align*}
        g &= 000 110 011 000 \\
        p &= 000 110 000 000 \\
        \hat{p}&= 000 110 001 000 
    \end{align*}
    Then $Precision(g,p) = 1 = Precision(g,\hat{p})$.
    
    \paragraph{\cref{prop:alarms}} 
    Consider the following example:
    \begin{align*}
        g &= 000 111 111 000\\
        p &= 000 111 000 000\\
        \hat{p}&= 000 111 011 000
    \end{align*}
    Then it holds $Precision(g,p) = 1 = Precision(g,\hat{p})$.
    
    \paragraph{\cref{prop:false_positives}}
    Consider the following example:
    \begin{align*}
        g &= 000 000 000 001\\
        p &= 000 000 010 000\\
        \hat{p}&= 000 000 110 000
    \end{align*}
    Then it holds $Precision(g,p) = 0 = Precision(g,\hat{p})$.
    
    \paragraph{\cref{prop:false_positive_alarms}}
    Consider the following example:
    \begin{align*}
        g &= 000 000 111 000 \\
        p &= 010 010 010 000 \\
        \hat{p}&= 011 100 010 000 
    \end{align*}
    Then $Precision(g,p) = \frac{1}{3} > \frac{1}{4} = Precision(g,\hat{p})$.
    
    \paragraph{\cref{prop:permutations}}
    Let $W \in A(|g-1|)$, $\hat{p}_{[n] \setminus W} = p_{[n] \setminus W}$, $|\hat{p}_W^{-1}(1)| = |p_W^{-1}(1)|$, and $|A(\hat{p}_W)| = |A(p_W)|$.
    Then $|\hat{p}^{-1}(1) \cap g^{-1}(1)| = |p^{-1}(1) \cap g^{-1}(1)|$ and $|\hat{p}^{-1}(1)| = |p^{-1}(1)|$.
    Thus, $Precision(g,p) = Precision(g,\hat{p})$
    
    \paragraph{\cref{prop:trust}}
    Consider the following example:
    \begin{align*}
        g &= 000 111 111 000 \\
        p &= 000 010 000 010 \\
        \hat{p}&= 010 111 111 010 
    \end{align*}
    Then $Precision(g,p) = \frac{1}{2} < \frac{3}{4} = Precision(g,\hat{p})$.
    
    \paragraph{\cref{prop:prediction_size}}
    Consider the following example:
    \begin{align*}
        g &= 000 111 111 000 \\
        p &= 000 010 000 000 \\
        \hat{p}&= 000 111 111 000 
    \end{align*}
    Then $Precision(g,p) = 1 = Precision(g,\hat{p})$.

    \paragraph{\cref{prop:temporal_order}}
    Consider the following example:
    \begin{align*}
        g &= 000 111 111 000 \\
        p &= 000 010 000 000 \\
        \hat{p}&= 000 000 010 000 
    \end{align*}
    Then $Precision(g,p) = 1 = Precision(g,\hat{p})$.
    
    \paragraph{\cref{prop:early_bias}}
    Consider the following example:
    \begin{align*}
        g &= 000 111 111 000 \\
        p &= 000 010 010 000 \\
        \hat{p}&= 000 110 000 000 
    \end{align*}
    Then $Precision(g,p) = 1 = Precision(g,\hat{p})$.
    
\end{proof}

\begin{theorem}\label{proof:point-wise_recall}
    \textbf{Point-wise Recall} defined as 
    \[
        Recall(g,p) = \frac{|p^{-1}(1) \cap g^{-1}(1)|}{|g^{-1}(1)|}
    \]
    satisfies properties \cref{prop:detection}, \cref{prop:permutations}, and \cref{prop:prediction_size}.
\end{theorem}
\begin{proof}
    \item
    Let $\colon [n] \rightarrow \{0,1\}^n$ be the ground truth and $p,\hat{p} \colon [n] \rightarrow \{0,1\}^n$ predictions.
    
    \paragraph{\cref{prop:detection}} 
    Let $W \in A(g)$, $\hat{p}_{[n] \setminus W} = p_{[n] \setminus W}$, $p_W = 0$, and $|\hat{p}_W^{-1}(1)| > 0$.
    Then $|\hat{p}^{-1}(1) \cap g^{-1}(1)| > |p^{-1}(1) \cap g^{-1}(1)|$.
    Thus, $Recall(g, p) < Recall(g, \hat{p})$.
    
    \paragraph{\cref{prop:alarms}} 
    Consider the following example:
    \begin{align*}
        g &= 000 111 111 000\\
        p &= 000 111 000 000\\
        \hat{p}&= 000 111 011 000
    \end{align*}
    Then it holds $Recall(g,p) = \frac{1}{2} < \frac{5}{6}= Recall(g,\hat{p})$.
    
    \paragraph{\cref{prop:false_positives}} 
    Consider the following example:
    \begin{align*}
        g &= 000 000 111 000\\
        p &= 000 100 010 000\\
        \hat{p}&= 001 100 010 000
    \end{align*}
    Then it holds $Recall(g,p) = \frac{1}{3} = Recall(g,\hat{p})$.
    
    \paragraph{\cref{prop:false_positive_alarms}}
    Consider the following example:
    \begin{align*}
        g &= 000 000 111 000\\
        p &= 000 010 111 000\\
        \hat{p}&= 000 000 111 000
    \end{align*}
    Then it holds $Recall(g,p) = 1 = Recall(g,\hat{p})$.
    
    \paragraph{\cref{prop:permutations}}
    Let $W \in A(|g-1|)$, $\hat{p}_{[n] \setminus W} = p_{[n] \setminus W}$, $|\hat{p}_W^{-1}(1)| = |p_W^{-1}(1)|$, and $|A(\hat{p}_W)| = |A(p_W)|$.
    Then $|\hat{p}^{-1}(1) \cap g^{-1}(1)| = |p^{-1}(1) \cap g^{-1}(1)|$.
    Thus, $Recall(g,p) = Recall(g,\hat{p})$.
    
    \paragraph{\cref{prop:trust}}
    Consider the following example:
    \begin{align*}
        g &= 000 000 111 000\\
        p &= 000 000 111 000\\
        \hat{p}&= 000 010 111 000
    \end{align*}
    Then it holds $Recall(g,p) = 1 = Recall(g,\hat{p})$.
    
    \paragraph{\cref{prop:prediction_size}}
    Let $W \in A(g)$, $\hat{p}_{[n] \setminus \{i^*\}} = p_{[n] \setminus \{i^*\}}$ for some $i^* \in W$, $\hat{p}(i^*) \neq p(i^*) = 0$, and $|A(\hat{p}_W)| \leq |A(p_W)|$.
    Then $|\hat{p}^{-1}(1) \cap g^{-1}(1)| > |p^{-1}(1) \cap g^{-1}(1)|$.
    Thus, $Recall(g,p) < Recall(g,\hat{p})$.
    
    \paragraph{\cref{prop:temporal_order}}
    Consider the following example:
    \begin{align*}
        g &= 000 111 111 000\\
        p &= 000 111 000 000\\
        \hat{p}&= 000 000 111 000
    \end{align*}
    Then it holds $Recall(g,p) = \frac{1}{2} = Recall(g,\hat{p})$.
    
    \paragraph{\cref{prop:early_bias}}
    Consider the following example:
    \begin{align*}
        g &= 000 111 111 000\\
        p &= 000 000 111 000\\
        \hat{p}&= 000 001 110 000
    \end{align*}
    Then it holds $Recall(g,p) = \frac{1}{2} = Recall(g,\hat{p})$.
\end{proof}
 
\begin{theorem}\label{proof:point-wise_f1-score}
    \textbf{Point-wise F1-score} defined as 
    \[
        F1(g,p) = \frac{2|p^{-1}(1) \cap g^{-1}(1)|}{2|p^{-1}(1) \cap g^{-1}(1)| + |p^{-1}(1) \cap g^{-1}(0)| + |p^{-1}(0) \cap g^{-1}(1)|} = \frac{2|p^{-1}(1) \cap g^{-1}(1)|}{|p^{-1}(1)| + |g^{-1}(1)|}
    \]
    satisfies properties \cref{prop:detection}, %\cref{prop:false_positives}, 
    \cref{prop:permutations}, and \cref{prop:prediction_size}.
\end{theorem}
\begin{proof}
    \item
    Let $\colon [n] \rightarrow \{0,1\}^n$ be the ground truth and $p,\hat{p} \colon [n] \rightarrow \{0,1\}^n$ predictions.
    \paragraph{\cref{prop:detection}} 
    Let $W \in A(g)$, $\hat{p}_{[n] \setminus W} = p_{[n] \setminus W}$, $p_W = 0$, and $|\hat{p}_W^{-1}(1)| > 0$.
    Then $|\hat{p}^{-1}(1) \cap g^{-1}(1)| > |p^{-1}(1) \cap g^{-1}(1)|$.
    Let $\epsilon = |\hat{p}^{-1}(1) \cap g^{-1}(1)| - |p^{-1}(1) \cap g^{-1}(1)|$.
    Then it holds 
    \[
    F1(g,\hat{p}) 
    = \frac{2|\hat{p}^{-1}(1) \cap g^{-1}(1)|}{|\hat{p}^{-1}(1)| + |g^{-1}(1)|}
    = \frac{2|p^{-1}(1) \cap g^{-1}(1)| + 2\epsilon}{|p^{-1}(1)| + \epsilon + |g^{-1}(1)|}
    > \frac{2|p^{-1}(1) \cap g^{-1}(1)|}{|p^{-1}(1)| + |g^{-1}(1)|}
    = F1(g,p)
    \]
    Thus, $F1(g,p) < F1(g,\hat{p})$.
    
    \paragraph{\cref{prop:alarms}} 
    Consider the following example:
    \begin{align*}
        g &= 000 111 111 000\\
        p &= 000 111 000 000\\
        \hat{p}&= 000 111 011 000
    \end{align*}
    Then it holds $F1(g,p) = \frac{2}{3} < \frac{10}{11} = F1(g,\hat{p})$.
    
    \paragraph{\cref{prop:false_positives}}
    Consider the following example:
    \begin{align*}
        g &= 000 000 000 001\\
        p &= 000 000 010 000\\
        \hat{p}&= 000 010 010 000
    \end{align*}
    Then it holds $F1(g,p) = 0 = F1(g,\hat{p})$.
    
    \paragraph{\cref{prop:false_positive_alarms}}
    Consider the following example:
    \begin{align*}
        g &= 000 000 111 000\\
        p &= 010 010 010 000\\
        \hat{p}&= 001 100 010 000
    \end{align*}
    Then it holds $F1(g,p) = \frac{1}{3} = F1(g,\hat{p})$.
    
    \paragraph{\cref{prop:permutations}}
    Let $W \in A(|g-1|)$, $\hat{p}_{[n] \setminus W} = p_{[n] \setminus W}$, $|\hat{p}_W^{-1}(1)| = |p_W^{-1}(1)|$, and $|A(\hat{p}_W)| = |A(p_W)|$.
    Then $|\hat{p}^{-1}(1) \cap g^{-1}(1)| = |p^{-1}(1) \cap g^{-1}(1)|$ and $|\hat{p}^{-1}(1)| = |p^{-1}(1)|$.
    Thus, $F1(g,p) = F1(g, \hat{p})$.
    
    \paragraph{\cref{prop:trust}}
    Consider the following example:
    \begin{align*}
        g &= 000 111 111 000 \\
        p &= 000 000 010 010 \\
        \hat{p}&= 010 001 111 010 
    \end{align*}
    Then $F1(g,p) = \frac{1}{4} < \frac{2}{3} = F1(g,\hat{p})$.
    
    \paragraph{\cref{prop:prediction_size}}
    Let $W \in A(g)$, $\hat{p}_{[n] \setminus \{i^*\}} = p_{[n] \setminus \{i^*\}}$ for some $i^* \in W$, $\hat{p}(i^*) \neq p(i^*) = 0$, and $|A(\hat{p}_W)| \leq |A(p_W)|$.
    Then $|\hat{p}^{-1}(1) \cap g^{-1}(1)| = |p^{-1}(1) \cap g^{-1}(1)| + 1$ and $|\hat{p}^{-1}(1)| = |p^{-1}(1)| + 1$.
    Thus, $F1(g,p) < F1(g, \hat{p})$.
    
    \paragraph{\cref{prop:temporal_order}}
    Consider the following example:
    \begin{align*}
        g &= 000 111 111 000 \\
        p &= 000 110 000 000 \\
        \hat{p}&= 000 000 011 000 
    \end{align*}
    Then $F1(g,p) = \frac{1}{2} = F1(g,\hat{p})$.
    
    \paragraph{\cref{prop:early_bias}}
    Consider the following example:
    \begin{align*}
        g &= 000 111 111 000 \\
        p &= 000 010 010 000 \\
        \hat{p}&= 000 110 000 000 
    \end{align*}
    Then $F1(g,p) = \frac{1}{2} = F1(g,\hat{p})$.
\end{proof}
 
\begin{theorem}\label{proof:pa-precision}
    \textbf{Point-adjusted Precision} \[
        Precision_{PA}(g,p) =  \frac{\sum\limits_{\stackrel{W \in A(g)\colon}{W \cap p^{-1}(1) \neq \emptyset}} |W|}{|p^{-1}(1) \cap g^{-1}(0)| + \sum\limits_{\stackrel{W \in A(g)\colon}{W \cap p^{-1}(1) \neq \emptyset}} |W|}
    \]
    has properties %\cref{prop:false_positives}, 
    \cref{prop:permutations} and \cref{prop:trust}.
\end{theorem}
\begin{proof}
    \item
    Let $\colon [n] \rightarrow \{0,1\}^n$ be the ground truth and $p,\hat{p} \colon [n] \rightarrow \{0,1\}^n$ predictions.
    
    \paragraph{\cref{prop:detection}} 
    Consider the following example:
    \begin{align*}
        g &= 000 110 011 000 \\
        p &= 000 110 000 000 \\
        \hat{p}&= 000 110 011 000 
    \end{align*}
    Then $Precision_{PA}(g,p) = 1 = Precision_{PA}(g,\hat{p})$.
    
    \paragraph{\cref{prop:alarms}} 
    Consider the following example:
    \begin{align*}
        g &= 000 111 111 000\\
        p &= 000 111 000 000\\
        \hat{p}&= 000 111 011 000
    \end{align*}
    Then it holds $Precision_{PA}(g,p) = 1 = Precision_{PA}(g,\hat{p})$.
    
    \paragraph{\cref{prop:false_positives}}
    Consider the following example:
    \begin{align*}
        g &= 000 000 000 001\\
        p &= 000 000 010 000\\
        \hat{p}&= 000 010 010 000
    \end{align*}
    Then it holds $Precision_{PA}(g,p) = 0 = Precision_{PA}(g,\hat{p})$.
    
    \paragraph{\cref{prop:false_positive_alarms}}
    Consider the following example:
    \begin{align*}
        g &= 000 000 111 000 \\
        p &= 010 010 010 000 \\
        \hat{p}&= 011 100 010 000 
    \end{align*}
    Then $Precision_{PA}(g,p) = \frac{3}{5} > \frac{1}{2} = Precision_{PA}(g,\hat{p})$.
    
    \paragraph{\cref{prop:permutations}}
    Let $W \in A(|g-1|)$, $\hat{p}_{[n] \setminus W} = p_{[n] \setminus W}$, $|\hat{p}_W^{-1}(1)| = |p_W^{-1}(1)|$, and $|A(\hat{p}_W)| = |A(p_W)|$.
    Then $|\hat{p}^{-1}(1)| = |p^{-1}(1)|$ and $\{W \in A(g) \colon W \cap \hat{p}^{-1}(1) \neq \emptyset \} = \{W \in A(g) \colon W \cap p^{-1}(1) \neq \emptyset \}$.
    Thus, $Precision_{PA}(g,p) = Precision_{PA}(g,\hat{p})$.
    
    \paragraph{\cref{prop:trust}}
    Let $W \in A(g)$, $\hat{W} \in A(|g-1|)$, $|A(\hat{p}_{W})| = |A(p_{W})|$, $\hat{p}_{[n] \setminus (W \cup \hat{W})} = p_{[n] \setminus (W \cup \hat{W})}$, $p_{\hat{W}} = 0$, and $\hat{p}_{\hat{W}} = \mathds{1}_{\{i^*\}}$ for some $i^* \in \hat{W}$.
    Then $\{W \in A(g) \colon W \cap \hat{p}^{-1}(1) \neq \emptyset \} = \{W \in A(g) \colon W \cap p^{-1}(1) \neq \emptyset \}$ and 
    $|\hat{p}^{-1}(1) \cap g^{-1}(0)| = |p^{-1}(1) \cap g^{-1}(0)| + 1$.
    Thus, $Precision_{PA}(g,p) > Precision_{PA}(g,\hat{p})$.
    
    \paragraph{\cref{prop:prediction_size}}
    Consider the following example:
    \begin{align*}
        g &= 000 111 111 000 \\
        p &= 000 010 000 000 \\
        \hat{p}&= 000 110 000 000 
    \end{align*}
    Then $Precision_{PA}(g,p) = 1 = Precision_{PA}(g,\hat{p})$.
    
    \paragraph{\cref{prop:temporal_order}}
    Consider the following example:
    \begin{align*}
        g &= 000 111 111 000 \\
        p &= 000 010 000 000 \\
        \hat{p}&= 000 000 001 000 
    \end{align*}
    Then $Precision_{PA}(g,p) = 1 = Precision_{PA}(g,\hat{p})$.
    
    \paragraph{\cref{prop:early_bias}}
    Consider the following example:
    \begin{align*}
        g &= 000 111 111 000 \\
        p &= 000 010 001 000 \\
        \hat{p}&= 000 110 000 000 
    \end{align*}
    Then $Precision_{PA}(g,p) = 1 = Precision_{PA}(g,\hat{p})$.
\end{proof}
 
\begin{theorem}\label{proof:pa-recall}
    \textbf{Point-adjusted Recall} \[
        Recall_{PA}(g,p) = \sum_{\stackrel{W \in A(g)\colon}{W \cap p^{-1}(1) \neq \emptyset}} \frac{|W|}{|g^{-1}(1)|}
    \]
    has \cref{prop:detection} and \cref{prop:permutations}.
\end{theorem}
\begin{proof}
    \item
    Let $\colon [n] \rightarrow \{0,1\}^n$ be the ground truth and $p,\hat{p} \colon [n] \rightarrow \{0,1\}^n$ predictions.
    
    \paragraph{\cref{prop:detection}} 
    Let $W \in A(g)$ an interval, $\hat{p}_{[n] \setminus W} = p_{[n] \setminus W}$, $p_W = 0$, and $|\hat{p}_W^{-1}(1)| > 0$.
    Then $Recall_{PA}(g,\hat{p}) = Recall_{PA}(g,p) + \frac{|W|}{|g^{-1}(1)|}$.
    Thus, $Recall_{PA}(g,p) < Recall_{PA}(g,\hat{p})$.
    
    \paragraph{\cref{prop:alarms}} 
    Consider the following example:
    \begin{align*}
        g &= 000 111 111 000\\
        p &= 000 111 000 000\\
        \hat{p}&= 000 111 011 000
    \end{align*}
    Then it holds $Recall_{PA}(g,p) = 1 = Recall_{PA}(g,\hat{p})$.
    
    \paragraph{\cref{prop:false_positives}} 
    Consider the following example:
    \begin{align*}
        g &= 000 000 111 000\\
        p &= 000 100 010 000\\
        \hat{p}&= 011 100 010 000
    \end{align*}
    Then it holds $Recall_{PA}(g,p) = 1 = Recall_{PA}(g,\hat{p})$.
    
    \paragraph{\cref{prop:false_positive_alarms}}
    Consider the following example:
    \begin{align*}
        g &= 000 000 111 000\\
        p &= 000 010 010 000\\
        \hat{p}&= 000 000 010 000
    \end{align*}
    Then it holds $Recall_{PA}(g,p) = 1 = Recall_{PA}(g,\hat{p})$.
    
    \paragraph{\cref{prop:permutations}}
    Let $W \subset W' \in A(|g-1|)$, $\hat{p}_{[n] \setminus W} = p_{[n] \setminus W}$, $|\hat{p}_W^{-1}(1)| = |p_W^{-1}(1)|$, and $|A(\hat{p}_W)| = |A(p_W)|$.
    Then $\{W \in A(g) \colon W \cap \hat{p}^{-1}(1) \neq \emptyset\} = \{W \in A(g) \colon W \cap p^{-1}(1) \neq \emptyset\}$.
    Thus, $Recall_{PA}(g,p) = Recall_{PA}(g,\hat{p})$.
    
    \paragraph{\cref{prop:trust}}
    Consider the following example:
    \begin{align*}
        g &= 000 000 111 000\\
        p &= 000 000 010 000\\
        \hat{p}&= 000 010 110 000
    \end{align*}
    Then it holds $Recall_{PA}(g,p) = 1 = Recall_{PA}(g,\hat{p})$.
    
    \paragraph{\cref{prop:prediction_size}}
    Consider the following example:
    \begin{align*}
        g &= 000 111 111 000\\
        p &= 000 010 000 000\\
        \hat{p}&= 000 011 000 000
    \end{align*}
    Then it holds $Recall_{PA}(g,p) = 1 = Recall_{PA}(g,\hat{p})$.
    
    \paragraph{\cref{prop:temporal_order}}
    Consider the following example:
    \begin{align*}
        g &= 000 111 111 000\\
        p &= 000 111 000 000\\
        \hat{p}&= 000 000 111 000
    \end{align*}
    Then it holds $Recall_{PA}(g,p) = 1 = Recall_{PA}(g,\hat{p})$.
    
    \paragraph{\cref{prop:early_bias}}
    Consider the following example:
    \begin{align*}
        g &= 000 111 111 000\\
        p &= 000 000 111 000\\
        \hat{p}&= 000 001 110 000
    \end{align*}
    Then it holds $Recall_{PA}(g,p) = 1 = Recall_{PA}(g,\hat{p})$.
\end{proof}
 
\begin{theorem}\label{proof:pa-f1-score}
    \textbf{Point-adjusted F1-score} 
    \begin{align*}
        F1_{PA}(p,g) &= \frac{2\sum\limits_{\stackrel{W \in A(g)\colon}{W \cap p^{-1}(1) \neq \emptyset}} |W|}{2\sum\limits_{\stackrel{W \in A(g)\colon}{W \cap p^{-1}(1) \neq \emptyset}} |W| + |p^{-1}(1) \cap g^{-1}(0)| + \sum\limits_{\stackrel{W \in A(g)\colon}{W \cap p^{-1}(1) = \emptyset}}|W|}\\
        &= \frac{2\sum\limits_{\stackrel{W \in A(g)\colon}{W \cap p^{-1}(1) \neq \emptyset}} |W|}{\sum\limits_{\stackrel{W \in A(g)\colon}{W \cap p^{-1}(1) \neq \emptyset}} |W| + |p^{-1}(1) \cap g^{-1}(0)| + |g^{-1}(1)|}
    \end{align*}
    has properties \cref{prop:detection}, %\cref{prop:false_positives}, 
    \cref{prop:permutations}, and \cref{prop:trust}.
\end{theorem}
\begin{proof}
    \item
    Let $\colon [n] \rightarrow \{0,1\}^n$ be the ground truth and $p,\hat{p} \colon [n] \rightarrow \{0,1\}^n$ predictions.
    
    \paragraph{\cref{prop:detection}} 
    Let $W \in A(g)$ an interval, $\hat{p}_{[n] \setminus W} = p_{[n] \setminus W}$, $p_W = 0$, and $|\hat{p}_W^{-1}(1)| > 0$.
    Then $\{W \in A(g) \colon W \cap \hat{p}^{-1}(1) \neq \emptyset\} = \{W \in A(g) \colon W \cap p^{-1}(1) \neq \emptyset\} \cup \{W\}$ with $|W| > 0$.
    Thus, $F1_{PA}(g, p) < F1_{PA}(g, \hat{p})$.
    
    \paragraph{\cref{prop:alarms}} 
    Consider the following example:
    \begin{align*}
        g &= 000 111 111 000\\
        p &= 000 111 000 000\\
        \hat{p}&= 000 111 011 000
    \end{align*}
    Then it holds $F1_{PA}(g,p) = 1 = F1_{PA}(g,\hat{p})$.
    
    \paragraph{\cref{prop:false_positives}}
    Consider the following example:
    \begin{align*}
        g &= 000 000 000 001\\
        p &= 000 000 010 000\\
        \hat{p}&= 000 010 010 000
    \end{align*}
    Then it holds $F1_{PA}(g,p) = 0 = F1_{PA}(g,\hat{p})$.
    
    \paragraph{\cref{prop:false_positive_alarms}}
    Consider the following example:
    \begin{align*}
        g &= 000 000 111 000\\
        p &= 010 010 010 000\\
        \hat{p}&= 001 100 010 000
    \end{align*}
    Then it holds $F1_{PA}(g,p) = \frac{3}{4} = F1_{PA}(g,\hat{p})$.
    
    \paragraph{\cref{prop:permutations}}
    Let $W \subset W' \in A(|g-1|)$, $\hat{p}_{[n] \setminus W} = p_{[n] \setminus W}$, $|\hat{p}_W^{-1}(1)| = |p_W^{-1}(1)|$, and $|A(\hat{p}_W)| = |A(p_W)|$.
    Then $\{W \in A(g) \colon W \cap \hat{p}^{-1}(1) \neq \emptyset\} = \{W \in A(g) \colon W \cap p^{-1}(1) \neq \emptyset\}$ and $|\hat{p}^{-1}(1) \cap g^{-1}(0)| = |p^{-1}(1) \cap g^{-1}(0)|$.
    Thus, $F1_{PA}(g,p) = F1_{PA}(g,\hat{p})$.
    
    \paragraph{\cref{prop:trust}}
    Let $W \in A(g)$, $\hat{W} \in A(|g-1|)$, $|A(\hat{p}_{W})| = |A(p_{W})|$, $\hat{p}_{[n] \setminus (W \cup \hat{W})} = p_{[n] \setminus (W \cup \hat{W})}$, $p_{\hat{W}} = 0$, and $\hat{p}_{\hat{W}} = \mathds{1}_{\{i^*\}}$ for some $i^* \in \hat{W}$. Then $\{W \in A(g) \colon W \cap \hat{p}^{-1}(1) \neq \emptyset\} = \{W \in A(g) \colon W \cap p^{-1}(1) \neq \emptyset\}$ and $|\hat{p}^{-1}(1) \cap g^{-1}(0)| = |p^{-1}(1) \cap g^{-1}(0)|+1$. Thus, $F1_{PA}(g,p) > F1_{PA}(g,\hat{p})$.
    
    \paragraph{\cref{prop:prediction_size}}
    Consider the following example:
    \begin{align*}
        g &= 000 111 111 000\\
        p &= 000 010 000 000\\
        \hat{p}&= 000 011 000 000
    \end{align*}
    Then it holds $F1_{PA}(g,p) = 1 = F1_{PA}(g,\hat{p})$.
    
    \paragraph{\cref{prop:temporal_order}}
    Consider the following example:
    \begin{align*}
        g &= 000 111 111 000\\
        p &= 000 110 000 000\\
        \hat{p}&= 000 011 000 000
    \end{align*}
    Then it holds $F1_{PA}(g,p) = 1 = F1_{PA}(g,\hat{p})$.
    
    \paragraph{\cref{prop:early_bias}}
    Consider the following example:
    \begin{align*}
        g &= 000 111 111 000\\
        p &= 000 010 010 000\\
        \hat{p}&= 000 110 000 000
    \end{align*}
    Then it holds $F1_{PA}(g,p) = 1 = F1_{PA}(g,\hat{p})$.
\end{proof}
 
\begin{theorem}\label{proof:pa-precision-event}
    \textbf{Event-wise Precision} \[
        Precision_{event}(g,p) = \frac{|\{W \in A(g) \colon W \cap p^{-1}(1) \neq \emptyset \}|}{|\{W \in A(g) \colon W \cap p^{-1}(1) \neq \emptyset \}| + |\{W \in A(p) \colon W \cap g^{-1}(1) = \emptyset\}|}
    \]
    has none of the properties.
\end{theorem}
\begin{proof}
    \item
    Let $\colon [n] \rightarrow \{0,1\}^n$ be the ground truth and $p,\hat{p} \colon [n] \rightarrow \{0,1\}^n$ predictions.
    
    \paragraph{\cref{prop:detection}} 
    Consider the following example:
    \begin{align*}
        g &= 000 110 011 000 \\
        p &= 000 110 000 000 \\
        \hat{p}&= 000 110 011 000 
    \end{align*}
    Then $Precision_{event}(g,p) = 1 = Precision_{event}(g,\hat{p})$.
    
    \paragraph{\cref{prop:alarms}} 
    Consider the following example:
    \begin{align*}
        g &= 000 111 111 000\\
        p &= 000 111 000 000\\
        \hat{p}&= 000 111 011 000
    \end{align*}
    Then it holds $Precision_{event}(g,p) = 1 = Precision_{event}(g,\hat{p})$.
    
    \paragraph{\cref{prop:false_positives}} 
    Consider the following example:
    \begin{align*}
        g &= 000 111 111 000\\
        p &= 001 111 000 000\\
        \hat{p}&= 011 111 000 000
    \end{align*}
    Then it holds $Precision_{event}(g,p) = 1 = Precision_{event}(g,\hat{p})$.
    
    \paragraph{\cref{prop:false_positive_alarms}}
    Consider the following example:
    \begin{align*}
        g &= 000 111 111 000\\
        p &= 000 111 000 000\\
        \hat{p}&= 001 111 000 000
    \end{align*}
    Then it holds $Precision_{event}(g,p) = 1 = Precision_{event}(g,\hat{p})$.

    \paragraph{\cref{prop:permutations}}
    Consider the following example:
    \begin{align*}
        g &= 000 111 111 000\\
        p &= 110 111 000 000\\
        \hat{p}&= 011 111 000 000
    \end{align*}
    Then it holds $Precision_{event}(g,p) = \frac{1}{2} < 1 = Precision_{event}(g,\hat{p})$.
    
    \paragraph{\cref{prop:trust}}
    Consider the following example:
    \begin{align*}
        g &= 000 111 111 000\\
        p &= 000 111 000 000\\
        \hat{p}&= 001 111 000 000
    \end{align*}
    Then it holds $Precision_{event}(g,p) = 1 = Precision_{event}(g,\hat{p})$.
    
    \paragraph{\cref{prop:prediction_size}}
    Consider the following example:
    \begin{align*}
        g &= 000 111 111 000\\
        p &= 000 101 100 000\\
        \hat{p}&= 000 111 100 000
    \end{align*}
    Then it holds $Precision_{event}(g,p) = 1 = Precision_{event}(g,\hat{p})$.
    
    \paragraph{\cref{prop:temporal_order}}
    Consider the following example:
    \begin{align*}
        g &= 000 111 111 000\\
        p &= 000 111 000 000\\
        \hat{p}&= 000 000 111 000
    \end{align*}
    Then it holds $Precision_{event}(g,p) = 1 = Precision_{event}(g,\hat{p})$. 
    
    \paragraph{\cref{prop:early_bias}}
    Consider the following example:
    \begin{align*}
        g &= 000 111 111 000\\
        p &= 000 001 001 000\\
        \hat{p}&= 000 011 000 000
    \end{align*}
    Then it holds $Precision_{event}(g,p) = 1 = Precision_{event}(g,\hat{p})$. 
    
\end{proof}
 
\begin{theorem}\label{proof:pa-recall-event}
    \textbf{Event-wise Recall} \[
        Recall_{event}(g,p) = \frac{|\{W \in A(g) \colon W \cap p^{-1}(1) \neq \emptyset \}|}{|A(g)|}
    \]
    has properties \cref{prop:detection} and \cref{prop:permutations}.
\end{theorem}
\begin{proof}
    \item
    Let $\colon [n] \rightarrow \{0,1\}^n$ be the ground truth and $p,\hat{p} \colon [n] \rightarrow \{0,1\}^n$ predictions.
    
    \paragraph{\cref{prop:detection}} 
    Let $W \in A(g)$, $\hat{p}_{[n] \setminus W} = p_{[n] \setminus W}$, $p_W = 0$, and $|\hat{p}_W^{-1}(1)| > 0$. Then $\{W' \in A(g) \colon W' \cap \hat{p}^{-1}(1) \neq \emptyset \} = \{W' \in A(g) \colon W' \cap p^{-1}(1) \neq \emptyset \} \cup \{W\}$. Thus $Recall_{event}(g, p)<Recall_{event}(g, \hat{p})$.
    
    \paragraph{\cref{prop:alarms}} 
    Consider the following example:
    \begin{align*}
        g &= 000 111 111 000\\
        p &= 000 111 000 000\\
        \hat{p}&= 000 111 011 000
    \end{align*}
    Then it holds $Recall_{event}(g,p) = 1 = Recall_{event}(g,\hat{p})$.
    
    \paragraph{\cref{prop:false_positives}} 
     Consider the following example:
    \begin{align*}
        g &= 000 111 111 000\\
        p &= 001 111 000 000\\
        \hat{p}&= 011 111 000 000
    \end{align*}
    Then it holds $Recall_{event}(g,p) = 1 = Recall_{event}(g,\hat{p})$.
    
    \paragraph{\cref{prop:false_positive_alarms}}
    Consider the following example:
    \begin{align*}
        g &= 000 111 111 000\\
        p &= 001 111 000 000\\
        \hat{p}&= 000 111 000 000
    \end{align*}
    Then it holds $Recall_{event}(g,p) = 1 = Recall_{event}(g,\hat{p})$.
    
    \paragraph{\cref{prop:permutations}}
    Let $W \in A(|g-1|)$, $\hat{p}_{[n] \setminus W} = p_{[n] \setminus W}$, $|\hat{p}_W^{-1}(1)| = |p_W^{-1}(1)|$, and $|A(\hat{p}_W)| = |A(p_W)|$. Then $\{W' \in A(g) \colon W' \cap \hat{p}^{-1}(1) \neq \emptyset \} = \{W' \in A(g) \colon W' \cap p^{-1}(1) \neq \emptyset \}$. Thus $Recall_{event}(g, p) = Recall_{event}(g, \hat{p})$.
    
    \paragraph{\cref{prop:trust}}
    Consider the following example:
    \begin{align*}
        g &= 000 111 111 000\\
        p &= 000 111 000 000\\
        \hat{p}&= 001 111 000 000
    \end{align*}
    Then it holds $Recall_{event}(g,p) = 1 = Recall_{event}(g,\hat{p})$.
    
    \paragraph{\cref{prop:prediction_size}}
    Consider the following example:
    \begin{align*}
        g &= 000 111 111 000\\
        p &= 000 111 000 000\\
        \hat{p}&= 000 111 100 000
    \end{align*}
    Then it holds $Recall_{event}(g,p) = 1 = Recall_{event}(g,\hat{p})$.
    
    \paragraph{\cref{prop:temporal_order}}
    Consider the following example:
    \begin{align*}
        g &= 000 111 111 000\\
        p &= 000 111 000 000\\
        \hat{p}&= 000 000 111 000
    \end{align*}
    Then it holds $Recall_{event}(g,p) = 1 = Recall_{event}(g,\hat{p})$.
    
    \paragraph{\cref{prop:early_bias}}
    Consider the following example:
    \begin{align*}
        g &= 000 111 111 000\\
        p &= 000 010 100 000\\
        \hat{p}&= 000 110 000 000
    \end{align*}
    Then it holds $Recall_{event}(g,p) = 1 = Recall_{event}(g,\hat{p})$.
\end{proof}
 
\begin{theorem}\label{proof:pa-f1-event}
    \textbf{Event-wise F1-score} 
    \begin{align*}
        &F1_{event}(g,p) = \\
        &\frac{2|\{W \in A(g) \colon W \cap p^{-1}(1) \neq \emptyset \}|}{2 |\{W \in A(g) \colon W \cap p^{-1}(1) \neq \emptyset \}| + |\{W \in A(p) \colon W \cap g^{-1}(1) = \emptyset\}| + |\{W \in A(g) \colon W \cap p^{-1}(1) = \emptyset \}|}
    \end{align*}
    has property \cref{prop:detection}.
\end{theorem}
\begin{proof}
    \item
    Let $\colon [n] \rightarrow \{0,1\}^n$ be the ground truth and $p,\hat{p} \colon [n] \rightarrow \{0,1\}^n$ predictions.
    
    \paragraph{\cref{prop:detection}} 
    Let $W \in A(g)$ an interval, $\hat{p}_{[n] \setminus W} = p_{[n] \setminus W}$, $p_W = 0$, and $|\hat{p}_W^{-1}(1)| > 0$. Then $|\{W \in A(g) \colon W \cap \hat{p}^{-1}(1) \neq \emptyset \}|=|\{W \in A(g) \colon W \cap p^{-1}(1) \neq \emptyset \}|+1$, $|\{W \in A(\hat{p}) \colon W \cap g^{-1}(1) = \emptyset\}|\leq |\{W \in A(p) \colon W \cap g^{-1}(1) = \emptyset\}|$, and $|\{W \in A(g) \colon W \cap p^{-1}(1) = \emptyset \}| = |\{W \in A(g) \colon W \cap \hat{p}^{-1}(1) = \emptyset \}|+1$. Thus, $F1_{event}(g,\hat{p}) > F1_{event}(g,p)$.
    
    \paragraph{\cref{prop:alarms}} 
    Consider the following example:
    \begin{align*}
        g &= 000 111 111 000\\
        p &= 000 111 000 000\\
        \hat{p}&= 000 111 011 000
    \end{align*}
    Then it holds $F1_{event}(g,p) = 1 = F1_{event}(g,\hat{p})$.
    
    \paragraph{\cref{prop:false_positives}} 
    Consider the following example:
    \begin{align*}
        g &= 000 111 111 000\\
        p &= 001 111 000 000\\
        \hat{p}&= 011 111 000 000
    \end{align*}
    Then it holds $F1_{event}(g,p) = 1 = F1_{event}(g,\hat{p})$.
    
    \paragraph{\cref{prop:false_positive_alarms}}
    Consider the following example:
    \begin{align*}
        g &= 000 111 111 000\\
        p &= 001 111 000 000\\
        \hat{p}&= 000 111 000 000
    \end{align*}
    Then it holds $F1_{event}(g,p) = 1 = F1_{event}(g,\hat{p})$.
    
    \paragraph{\cref{prop:permutations}}
    Consider the following example:
    \begin{align*}
        g &= 000 111 111 000\\
        p &= 001 111 000 000\\
        \hat{p}&= 010 111 000 000
    \end{align*}
    Then it holds $F1_{event}(g,p) = 1 > \frac{2}{3} = F1_{event}(g,\hat{p})$.
    
    \paragraph{\cref{prop:trust}}
    Consider the following example:
    \begin{align*}
        g &= 000 111 111 000\\
        p &= 000 111 000 000\\
        \hat{p}&= 001 111 000 000
    \end{align*}
    Then it holds $F1_{event}(g,p) = 1 = F1_{event}(g,\hat{p})$.
    
    \paragraph{\cref{prop:prediction_size}}
    Consider the following example:
    \begin{align*}
        g &= 000 111 111 000\\
        p &= 000 111 000 000\\
        \hat{p}&= 000 111 100 000
    \end{align*}
    Then it holds $F1_{event}(g,p) = 1 = F1_{event}(g,\hat{p})$.
    
    \paragraph{\cref{prop:temporal_order}}
    Consider the following example:
    \begin{align*}
        g &= 000 111 111 000\\
        p &= 000 111 000 000\\
        \hat{p}&= 000 000 111 000
    \end{align*}
    Then it holds $F1_{event}(g,p) = 1 = F1_{event}(g,\hat{p})$.
    
    \paragraph{\cref{prop:early_bias}}
    Consider the following example:
    \begin{align*}
        g &= 000 111 111 000\\
        p &= 000 010 100 000\\
        \hat{p}&= 000 110 000 000
    \end{align*}
    Then it holds $F1_{event}(g,p) = 1 = F1_{event}(g,\hat{p})$.
\end{proof}
 
\begin{theorem}\label{proof:point-wise_composite_f1-score}
    \textbf{Composite F1-score} \[
        F_{c1} = 2\frac{Precision(g,p) \cdot Recall_{event}(g,p)}{Precision(g,p) + Recall_{event}(g,p)}
    \]
    where 
    \[
        Precision(g,p) = \frac{|p^{-1}(1) \cap g^{-1}(1)|}{|p^{-1}(1)|} \qquad \text{and}\qquad Recall_{event}(g,p) = \frac{|\{W \in A(g) \colon W \cap p^{-1}(1) \neq \emptyset\}|}{|A(g)|}
    \]
    has properties \cref{prop:detection}, %\cref{prop:false_positives}, 
    \cref{prop:permutations}, and \cref{prop:trust}.
\end{theorem}
\begin{proof}
    \item
    Let $\colon [n] \rightarrow \{0,1\}^n$ be the ground truth and $p,\hat{p} \colon [n] \rightarrow \{0,1\}^n$ predictions.
    \paragraph{\cref{prop:detection}} 
    Let $W \in A(g)$, $\hat{p}_{[n] \setminus W} = p_{[n] \setminus W}$, $p_W = 0$, and $|\hat{p}_W^{-1}(1)| > 0$.
    Then $|\{W \in A(g) \colon W \cap \hat{p}^{-1}(1) \neq \emptyset\}| = |\{W \in A(g) \colon W \cap p^{-1}(1) \neq \emptyset\}| + 1$.
    Thus, $Recall_{event}(g, p) < Recall_{event}(g,\hat{p})$.
    We also know that $Precision(g,p) < Precision(g, \hat{p})$ (\cref{proof:point-wise_precision}).
    Thus, $F_{c1}(g, p) < F_{c1}(g,\hat{p})$.
    
    \paragraph{\cref{prop:alarms}}
    Consider the following example:
    \begin{align*}
        g &= 000 111 111 000\\
        p &= 000 111 000 000\\
        \hat{p}&= 000 111 011 000
    \end{align*}
    Then it holds $Precision(g,p) = 1 = Precision(g,\hat{p})$, $Recall_{event}(g,p) = 1 = Recall_{event}(g, \hat{p})$, and $F_{c1}(g,p) = 1 = F_{c1}(g, \hat{p})$.
    
    \paragraph{\cref{prop:false_positives}}
    Consider the following example:
    \begin{align*}
        g &= 000 000 000 001\\
        p &= 000 000 010 000\\
        \hat{p}&= 000 010 010 000
    \end{align*}
    Then it holds $F_{c1}(g,p) = 0 = F_{c1}(g,\hat{p})$.
    
    \paragraph{\cref{prop:false_positive_alarms}}
    Consider the following example:
    \begin{align*}
        g &= 000 000 111 000\\
        p &= 010 010 010 000\\
        \hat{p}&= 001 100 010 000
    \end{align*}
    Then it holds $Precision(g,p) = 1 = Precision(g,\hat{p})$, $Recall_{event}(g,p) = 1 = Recall_{event}(g, \hat{p})$, and $F_{c1}(g,p) = 1 = F_{c1}(g, \hat{p})$.
    
    \paragraph{\cref{prop:permutations}}
    Let $W \in A(|g-1|)$, $\hat{p}_{[n] \setminus W} = p_{[n] \setminus W}$, $|\hat{p}_W^{-1}(1)| = |p_W^{-1}(1)|$, and $|A(\hat{p}_W)| = |A(p_W)|$.
    Then $Precision(g, p) = Precision(g, \hat{p})$ (\cref{proof:point-wise_precision}), $|\{W \in A(g) \colon W \cap \hat{p}^{-1}(1) \neq \emptyset\}| = |\{W \in A(g) \colon W \cap p^{-1}(1) \neq \emptyset\}|$, and $Recall_{event}(g,p) = Recall_{event}(g, \hat{p})$.
    Thus, $F_{c1}(g,p) = F_{c1}(g, \hat{p})$.
    
    \paragraph{\cref{prop:trust}}
    Let $W \in A(g)$, $\hat{W} \in A(|g-1|)$, $|A(\hat{p}_{W})| = |A(p_{W})|$, $\hat{p}_{[n] \setminus (W \cup \hat{W})} = p_{[n] \setminus (W \cup \hat{W})}$, $p_{\hat{W}} = 0$, and $\hat{p}_{\hat{W}} = \mathds{1}_{\{i^*\}}$ for some $i^* \in \hat{W}$. Then $|\{W \in A(g) \colon W \cap \hat{p}^{-1}(1) \neq \emptyset\}| = |\{W \in A(g) \colon W \cap p^{-1}(1) \neq \emptyset\}|$ and $Recall_{event}(g,p) = Recall_{event}(g, \hat{p})$. $|p^{-1}(1) \cap g^{-1}(1)| = |\hat{p}^{-1}(1) \cap g^{-1}(1)|$ and $|\hat{p}^{-1}(1)|=|p^{-1}(1)|+1$ thus $Precision(g, p) > Precision(g, \hat{p})$ and finally $F_{c1}(g,p) > F_{c1}(g, \hat{p})$.
    
    \paragraph{\cref{prop:prediction_size}}
    Consider the following example:
    \begin{align*}
        g &= 000 000 111 000\\
        p &= 000 000 011 000\\
        \hat{p}&= 000 000 111 000
    \end{align*}
    Then it holds $Precision(g,p) = 1 = Precision(g,\hat{p})$, $Recall_{event}(g,p) = 1 = Recall_{event}(g, \hat{p})$, and $F_{c1}(g,p) = 1 = F_{c1}(g, \hat{p})$.
    
    \paragraph{\cref{prop:temporal_order}}
    Consider the following example:
    \begin{align*}
        g &= 000 111 111 000\\
        p &= 000 111 000 000\\
        \hat{p}&= 000 000 111 000
    \end{align*}
    Then it holds $Precision(g,p) = 1 = Precision(g,\hat{p})$, $Recall_{event}(g,p) = 1 = Recall_{event}(g, \hat{p})$, and $F_{c1}(g,p) = 1 = F_{c1}(g, \hat{p})$.
    
    \paragraph{\cref{prop:early_bias}}
    Consider the following example:
    \begin{align*}
        g &= 000 111 111 000\\
        p &= 000 000 111 000\\
        \hat{p}&= 000 001 110 000
    \end{align*}
    Then it holds $Precision(g,p) = 1 = Precision(g,\hat{p})$, $Recall_{event}(g,p) = 1 = Recall_{event}(g, \hat{p})$, and $F_{c1}(g,p) = 1 = F_{c1}(g, \hat{p})$.
\end{proof}

\begin{theorem}\label{proof:pa-precision-k-delay}
    \textbf{K-Delay Precision} \[
        Prescision_{k-delay}(g,p) = \frac{\sum\limits_{\stackrel{W \in A(g)\colon}{\min\limits_{i \in W \cap p^{-1}(1)} i \leq \min\limits_{i \in W} i + k}} |W|}{|p^{-1}(1) \cap g^{-1}(0)| + \sum\limits_{\stackrel{W \in A(g)\colon}{\min\limits_{i \in W \cap p^{-1}(1)} i \leq \min\limits_{i \in W} i + k}} |W|}
    \]
    where $k \geq 0$
    has property %\cref{prop:false_positives} and 
    \cref{prop:permutations}.
\end{theorem}
\begin{proof}
    \item
    Let $\colon [n] \rightarrow \{0,1\}^n$ be the ground truth and $p,\hat{p} \colon [n] \rightarrow \{0,1\}^n$ predictions.
    
    \paragraph{\cref{prop:detection}} 
    Consider the following example: $g = \mathds{1}_{[a, a + k + 1]} + \mathds{1}_{[b,c]}$ with $0 < a < a + k +1 < b < c < n$, $p = \mathds{1}_{[b,c]}$, and $\hat{p} = \mathds{1}_{\{a + k + 1\}} + \mathds{1}_{[b,c]}$.
    Then it holds $\min\limits_{i \in [a,a+k+1] \cap p^{-1}(1)}i = a + k +1 > a + k = \min\limits_{i \in [a,a+k+1]} i + k$.
    Thus, $Prescision_{k-delay}(g,p) = 1 = Prescision_{k-delay}(g,\hat{p})$.
    
    For example, let $k=1$:
    \begin{align*}
        g &= 000 111 011 000\\
        p &= 000 000 011 000\\
        \hat{p}&= 000 001 011 000
    \end{align*}
    
    \paragraph{\cref{prop:alarms}} 
    Consider the following example:
    \begin{align*}
        g &= 000 111 111 000\\
        p &= 000 111 000 000\\
        \hat{p}&= 000 111 011 000
    \end{align*}
    Then it holds $Prescision_{k-delay}(g,p) = 1 = Prescision_{k-delay}(g,\hat{p})$.
    
    \paragraph{\cref{prop:false_positives}}
    Consider the following example:
    \begin{align*}
        g &= 000 000 000 001\\
        p &= 000 000 010 000\\
        \hat{p}&= 000 010 010 000
    \end{align*}
    Then it holds $Prescision_{k-delay}(g,p) = 0 = Prescision_{k-delay}(g,\hat{p})$.
    
    \paragraph{\cref{prop:false_positive_alarms}}
    Consider the following example:
    \begin{align*}
        g &= 000 000 111 000\\
        p &= 010 100 111 000\\
        \hat{p}&= 011 000 111 000
    \end{align*}
    Then it holds $Prescision_{k-delay}(g,p) = \frac{3}{5} = Prescision_{k-delay}(g,\hat{p})$.
    
    \paragraph{\cref{prop:permutations}}
    Let $W \in A(|g-1|)$, $\hat{p}_{[n] \setminus W} = p_{[n] \setminus W}$, $|\hat{p}_W^{-1}(1)| = |p_W^{-1}(1)|$, and $|A(\hat{p}_W)| = |A(p_W)|$. Then \[\sum\limits_{\stackrel{W \in A(g)\colon}{\min\limits_{i \in W \cap p^{-1}(1)} i \leq \min\limits_{i \in W} i + k}} |W| = \sum\limits_{\stackrel{W \in A(g)\colon}{\min\limits_{i \in W \cap \hat{p}^{-1}(1)} i \leq \min\limits_{i \in W} i + k}} |W|\] and $|\hat{p}^{-1}(1) \cap g^{-1}(0)| = |p^{-1}(1) \cap g^{-1}(0)|$. Thus $Prescision_{k-delay}(g,p) = Prescision_{k-delay}(g,\hat{p})$.
    
    \paragraph{\cref{prop:trust}}
    Consider the following example:
    $g=\mathds{1}_{[a,a + k + 1]}$ with $a > 1$, $p=\mathds{1}_{\{a + k + 1\}}$, and $\hat{p} = \mathds{1}_{\{a\}} + \mathds{1}_{\{0\}}$.
    Then it holds $Prescision_{k-delay}(g,p) = 0 < \frac{k+2}{k + 3} = Prescision_{k-delay}(g,\hat{p})$.\\
    For example, let $k=0$:
    \begin{align*}
        g &= 000 110 000 000\\
        p &= 000 010 000 000\\
        \hat{p}&= 100 100 000 000
    \end{align*}
    Then it holds $Prescision_{k-delay}(g,p) = 0 < \frac{2}{3} = Prescision_{k-delay}(g,\hat{p})$.
    
    \paragraph{\cref{prop:prediction_size}}
    Consider the following example:
    \begin{align*}
        g &= 000 111 111 000\\
        p &= 000 110 000 000\\
        \hat{p}&= 000 111 000 000
    \end{align*}
    Then it holds $Prescision_{k-delay}(g,p) = 1 = Prescision_{k-delay}(g,\hat{p})$.
    
    \paragraph{\cref{prop:temporal_order}}
    Consider the following example for $k > 0$:
    \begin{align*}
        g &= 000 111 111 000\\
        p &= 000 111 000 000\\
        \hat{p}&= 000 011 100 000
    \end{align*}
    Then it holds $Prescision_{k-delay}(g,p) = 1 = Prescision_{k-delay}(g,\hat{p})$.\\
    Consider the following example for $k = 0$:
    \begin{align*}
        g &= 000 111 111 000\\
        p &= 000 011 000 000\\
        \hat{p}&= 000 001 100 000
    \end{align*}
    Then it holds $Prescision_{k-delay}(g,p) = 0 = Prescision_{k-delay}(g,\hat{p})$.
    
    \paragraph{\cref{prop:early_bias}}
    Consider the following example:
    \begin{align*}
        g &= 000 111 111 000\\
        p &= 000 100 100 000\\
        \hat{p}&= 000 110 000 000
    \end{align*}
    Then it holds $Prescision_{k-delay}(g,p) = 1 = Prescision_{k-delay}(g,\hat{p})$.
\end{proof}

\begin{theorem}\label{proof:pa-recall-k-delay}
    \textbf{K-Delay Recall} \[
        Recall_{k-delay}(g,p) = \frac{\sum\limits_{\stackrel{W \in A(g)\colon}{\min\limits_{i \in W \cap p^{-1}(1)} i \leq \min\limits_{i \in W} i + k}} |W|}{\sum\limits_{W \in A(g)} |W|}
    \]
    has property \cref{prop:permutations}.
\end{theorem}
\begin{proof}
    \item
    Let $\colon [n] \rightarrow \{0,1\}^n$ be the ground truth and $p,\hat{p} \colon [n] \rightarrow \{0,1\}^n$ predictions.
    
    \paragraph{\cref{prop:detection}} 
    Consider the following example: $g = \mathds{1}_{[a, a + k + 1]} + \mathds{1}_{[b,c]}$ with $0 < a < a + k +1 < b < c < n$, $p = \mathds{1}_{[b,c]}$, and $\hat{p} = \mathds{1}_{\{a + k + 1\}} + \mathds{1}_{[b,c]}$.
    Then it holds $\min\limits_{i \in [a,a+k+1] \cap \hat{p}^{-1}(1)}i = a + k +1 > a + k = \min\limits_{i \in [a,a+k+1]} i + k$.
    Thus, $Recall_{k-delay}(g,p) = \frac{c-b+1}{c-b+k+3} = Recall_{k-delay}(g,\hat{p})$.
    
    For example, let $k=1$:
    \begin{align*}
        g &= 000 111 011 000\\
        p &= 000 000 011 000\\
        \hat{p}&= 000 001 011 000
    \end{align*}
    
    \paragraph{\cref{prop:alarms}}
    Consider the following example:
    \begin{align*}
        g &= 000 111 111 000\\
        p &= 000 111 000 000\\
        \hat{p}&= 000 111 011 000
    \end{align*}
    Then it holds $Recall_{k-delay}(g,p) = 1 = Recall_{k-delay}(g,\hat{p})$.
    
    \paragraph{\cref{prop:false_positives}} 
    Consider the following example:
    \begin{align*}
        g &= 000 111 111 000\\
        p &= 010 111 000 000\\
        \hat{p}&= 110 111 000 000
    \end{align*}
    Then it holds $Recall_{k-delay}(g,p) = 1 = Recall_{k-delay}(g,\hat{p})$.
    
    \paragraph{\cref{prop:false_positive_alarms}}
    Consider the following example:
    \begin{align*}
        g &= 000 111 111 000\\
        p &= 000 111 000 000\\
        \hat{p}&= 010 111 000 000
    \end{align*}
    Then it holds $Recall_{k-delay}(g,p) = 1 = Recall_{k-delay}(g,\hat{p})$.
    
    \paragraph{\cref{prop:permutations}}
    Let $W \in A(|g-1|)$, $\hat{p}_{[n] \setminus W} = p_{[n] \setminus W}$, $|\hat{p}_W^{-1}(1)| = |p_W^{-1}(1)|$, and $|A(\hat{p}_W)| = |A(p_W)|$. Then \[\sum\limits_{\stackrel{W \in A(g)\colon}{\min\limits_{i \in W \cap \hat{p}^{-1}(1)} i \leq \min\limits_{i \in W} i + k}} |W| = \sum\limits_{\stackrel{W \in A(g)\colon}{\min\limits_{i \in W \cap p^{-1}(1)} i \leq \min\limits_{i \in W} i + k}} |W|\]. Thus, $Recall_{k-delay}(g,p) = Recall_{k-delay}(g,\hat{p})$.
    
    \paragraph{\cref{prop:trust}}
    Consider the following example:
    \begin{align*}
        g &= 000 111 111 000\\
        p &= 000 110 100 000\\
        \hat{p}&= 010 110 010 000
    \end{align*}
    Then it holds $Recall_{k-delay}(g,p) = 1 = Recall_{k-delay}(g,\hat{p})$.
    
    \paragraph{\cref{prop:prediction_size}}
    Consider the following example:
    \begin{align*}
        g &= 000 111 111 000\\
        p &= 000 100 000 000\\
        \hat{p}&= 000 110 000 000
    \end{align*}
    Then it holds $Recall_{k-delay}(g,p) = 1 = Recall_{k-delay}(g,\hat{p})$.
    
    \paragraph{\cref{prop:temporal_order}}
    Consider the following example for $k > 0$:
    \begin{align*}
        g &= 000 111 111 000\\
        p &= 000 111 000 000\\
        \hat{p}&= 000 011 100 000
    \end{align*}
    Then it holds $Recall_{k-delay}(g,p) = 1 = Recall_{k-delay}(g,\hat{p})$.\\
    Consider the following example for $k = 0$:
    \begin{align*}
        g &= 000 111 111 000\\
        p &= 000 011 000 000\\
        \hat{p}&= 000 001 100 000
    \end{align*}
    Then it holds $Recall_{k-delay}(g,p) = 0 = Recall_{k-delay}(g,\hat{p})$.
    
    \paragraph{\cref{prop:early_bias}}
    Consider the following example:
    \begin{align*}
        g &= 000 111 111 000\\
        p &= 000 100 100 000\\
        \hat{p}&= 000 110 000 000
    \end{align*}
    Then it holds $Recall_{k-delay}(g,p) = 1 = Recall_{k-delay}(g,\hat{p})$.
\end{proof}

\begin{theorem}\label{proof:pa-f1-k-delay}
    \textbf{K-Delay F1-Score} \[
        F1_{k-delay}(g,p) = \frac{2\sum\limits_{\stackrel{W \in A(g)\colon}{\min\limits_{i \in W \cap p^{-1}(1)} i \leq \min\limits_{i \in W} i + k}} |W|}{|p^{-1}(1) \cap g^{-1}(0)| + 2\sum\limits_{\stackrel{W \in A(g)\colon}{\min\limits_{i \in W \cap p^{-1}(1)} i \leq \min\limits_{i \in W} i + k}} |W| 
        + \sum\limits_{\stackrel{W \in A(g)\colon}{\min\limits_{i \in W \cap p^{-1}(1)} i > \min\limits_{i \in W} i + k}} |W| 
        + \sum\limits_{\stackrel{W \in A(g)\colon}{W \cap p^{-1}(1) = \emptyset}} |W|}
    \]
    has properties %\cref{prop:false_positives} and 
    \cref{prop:permutations}.
\end{theorem}
\begin{proof}
    \item
    Let $\colon [n] \rightarrow \{0,1\}^n$ be the ground truth and $p,\hat{p} \colon [n] \rightarrow \{0,1\}^n$ predictions.
    
    \paragraph{\cref{prop:detection}} 
    Consider the following example: $g = \mathds{1}_{[a, a + k + 1]} + \mathds{1}_{[b,c]}$ with $0 < a < a + k +1 < b < c < n$, $p = \mathds{1}_{[b,c]}$, and $\hat{p} = \mathds{1}_{\{a + k + 1\}} + \mathds{1}_{[b,c]}$.
    Then it holds $\min\limits_{i \in [a,a+k+1] \cap p^{-1}(1)}i = a + k +1 > a + k = \min\limits_{i \in [a,a+k+1]} i + k$.
    Thus, $F1_{k-delay}(g,p) = \frac{2*(c-b+1)}{2*(c-b+1)+k+2} = F1_{k-delay}(g,\hat{p})$.
    
    For example, let $k=1$:
    \begin{align*}
        g &= 000 111 011 000\\
        p &= 000 000 011 000\\
        \hat{p}&= 000 001 011 000
    \end{align*}
    
    \paragraph{\cref{prop:alarms}}
    Consider the following example:
    \begin{align*}
        g &= 000 111 111 000\\
        p &= 000 111 000 000\\
        \hat{p}&= 000 111 011 000
    \end{align*}
    Then it holds $F1_{k-delay}(g,p) = 1 = F1_{k-delay}(g,\hat{p})$.
    
    \paragraph{\cref{prop:false_positives}}
    Consider the following example:
    \begin{align*}
        g &= 000 000 000 001\\
        p &= 000 000 010 000\\
        \hat{p}&= 000 010 010 000
    \end{align*}
    Then it holds $F1_{k-delay}(g,p) = 0 = F1_{k-delay}(g,\hat{p})$.

    \paragraph{\cref{prop:false_positive_alarms}}
    Consider the following example:
    \begin{align*}
        g &= 000 000 111 000\\
        p &= 010 100 111 000\\
        \hat{p}&= 011 000 111 000
    \end{align*}
    Then it holds $F1_{k-delay}(g,p) = \frac{3}{4} = F1_{k-delay}(g,\hat{p})$.
    
    \paragraph{\cref{prop:permutations}}
    Let $W \in A(|g-1|)$, $\hat{p}_{[n] \setminus W} = p_{[n] \setminus W}$, $|\hat{p}_W^{-1}(1)| = |p_W^{-1}(1)|$, and $|A(\hat{p}_W)| = |A(p_W)|$. Then \[\sum\limits_{\stackrel{W \in A(g)\colon}{\min\limits_{i \in W \cap p^{-1}(1)} i \leq \min\limits_{i \in W} i + k}} |W| = \sum\limits_{\stackrel{W \in A(g)\colon}{\min\limits_{i \in W \cap \hat{p}^{-1}(1)} i \leq \min\limits_{i \in W} i + k}} |W|\] and $|\hat{p}^{-1}(1) \cap g^{-1}(0)| = |p^{-1}(1) \cap g^{-1}(0)|$. Thus $F1_{k-delay}(g,p) = F1_{k-delay}(g,\hat{p})$.
    
    \paragraph{\cref{prop:trust}}
    Consider the following example:
    $g=\mathds{1}_{[a,a + k + 1]}$ with $a > 1$, $p=\mathds{1}_{\{a + k + 1\}}$, and $\hat{p} = \mathds{1}_{\{a\}} + \mathds{1}_{\{0\}}$.
    Then it holds $F1_{k-delay}(g,p) = 0 < \frac{2(k+2)}{2(k + 2)+1} = F1_{k-delay}(g,\hat{p})$.\\
    For example, let $k=0$:
    \begin{align*}
        g &= 000 110 000 000\\
        p &= 000 010 000 000\\
        \hat{p}&= 100 100 000 000
    \end{align*}
    Then it holds $F1_{k-delay}(g,p) = 0 < \frac{4}{5} = F1_{k-delay}(g,\hat{p})$.
    
    \paragraph{\cref{prop:prediction_size}}
    Consider the following example:
    \begin{align*}
        g &= 000 111 111 000\\
        p &= 000 110 000 000\\
        \hat{p}&= 000 111 000 000
    \end{align*}
    Then it holds $F1_{k-delay}(g,p) = 1 = F1_{k-delay}(g,\hat{p})$.
    
    \paragraph{\cref{prop:temporal_order}}
    Consider the following example for $k > 0$:
    \begin{align*}
        g &= 000 111 111 000\\
        p &= 000 111 000 000\\
        \hat{p}&= 000 011 100 000
    \end{align*}
    Then it holds $F1_{k-delay}(g,p) = 1 = F1_{k-delay}(g,\hat{p})$.\\
    Consider the following example for $k = 0$:
    \begin{align*}
        g &= 000 111 111 000\\
        p &= 000 011 000 000\\
        \hat{p}&= 000 001 100 000
    \end{align*}
    Then it holds $F1_{k-delay}(g,p) = 0 = F1_{k-delay}(g,\hat{p})$.
    
    \paragraph{\cref{prop:early_bias}}
    Consider the following example:
    \begin{align*}
        g &= 000 111 111 000\\
        p &= 000 100 100 000\\
        \hat{p}&= 000 110 000 000
    \end{align*}
    Then it holds $F1_{k-delay}(g,p) = 1 = F1_{k-delay}(g,\hat{p})$.
\end{proof}

\begin{theorem}\label{proof:pa-f1-lsa}
    \textbf{Latency- and Sparsity-Aware F1-score} \[
        F1_{lsa}(g,p) = \frac{2|p_{\delta_{PA}}^{-1}(1) \cap g_\delta^{-1}(1)|}{2|p_{\delta_{PA}}^{-1}(1) \cap g_\delta^{-1}(1)| + |p_{\delta_{PA}}^{-1}(1) \cap g_\delta^{-1}(0)| + |p_{\delta_{PA}}^{-1}(0) \cap g_\delta^{-1}(1)|} 
    \]
    where: \\
    \begin{align*}
    b &\in \mathbb N\\
    g_\delta(i) &= \max_{j \in [b\cdot i, b\cdot (i+1)-1]} g(j)\\
    g' &= \sum\limits_{i \in [\lceil \frac{n}{b} \rceil]} \mathds{1}_{[i*b, (i+1) * b-1] \cap [n]}\mathds{1}_{g(i * b) = 1}\\
     i^*(g', i)&= \begin{cases}\left\lfloor\frac{\min\limits_{k\in W: W\in A(g'), bi\in W} k}{b}\right\rfloor &\text{,if $\exists W\in A(g'): bi\in W$}\\bi &\text{,else}\end{cases}\\
    p_{\delta_{PA}}(i) &= \max_{j \in [bi^*(g', i), b(i+1)-1]} p(j)
    \end{align*}
    has property \cref{prop:temporal_order} if $b=1$.
    
\end{theorem}
\begin{proof}
    \item
    Let $\colon [n] \rightarrow \{0,1\}^n$ be the ground truth and $p,\hat{p} \colon [n] \rightarrow \{0,1\}^n$ predictions.
    
    \paragraph{\cref{prop:detection}}
    Consider the following example:
    $g=\mathds{1}_{\{b-1\}}$, $p=\mathds{1}_{\{0\}}$, and $\hat{p} = \mathds{1}_{\{0\}}+ \mathds{1}_{\{b-1\}}.$
    Then it holds $F1_{lsa}(g,p) = 1 = F1_{lsa}(g,\hat{p})$.

    For $b=3$:
    \begin{align*}
        g &= 001 000 000 000\\
        p &= 100 000 000 000\\
        \hat{p}&= 101 000 000 000
    \end{align*}
    Then it holds $F1_{lsa}(g,p) = 1 = F1_{lsa}(g,\hat{p})$.
    
    \paragraph{\cref{prop:alarms}}
    Consider the following example:
    $g=\mathds{1}_{[a,c]}$ where $0<a<c-1$, $p=\mathds{1}_{\{a\}}$, and $\hat{p} = \mathds{1}_{\{a\}}+ \mathds{1}_{\{a+2\}}.$
    Then it holds $F1_{lsa}(g,p) = 1 = F1_{lsa}(g,\hat{p})$.
    
    For $b=3$:
    \begin{align*}
        g &= 000 111 111 000\\
        p &= 000 100 000 000\\
        \hat{p}&= 000 101 000 000
    \end{align*}
    Then it holds $F1_{lsa}(g,p) = 1 = F1_{lsa}(g,\hat{p})$.
    
    \paragraph{\cref{prop:false_positives}}
    Consider the following example:
    $g=\mathds{1}_{\{0\}}$, $p=\mathds{1}_{\{b\}}$, and $\hat{p} = \mathds{1}_{\{b\}}+ \mathds{1}_{\{2b\}}.$
    Then it holds $F1_{lsa}(g,p) = 0 = F1_{lsa}(g,\hat{p})$.
    
    For $b=3$:
    \begin{align*}
        g &= 100 000 000 000\\
        p &= 000 100 000 000\\
        \hat{p}&= 000 100 100 000
    \end{align*}
    Then it holds $F1_{lsa}(g,p) = 0 = F1_{lsa}(g,\hat{p})$.
    
    \paragraph{\cref{prop:false_positive_alarms}}
    Consider the following example:
    $g=\mathds{1}_{\{0\}}$, $p=\mathds{1}_{\{0\}}+ \mathds{1}_{\{b\}}$, and $\hat{p} = \mathds{1}_{\{0\}}+ \mathds{1}_{\{b\}}+ \mathds{1}_{\{2b-1\}}.$
    Then it holds $F1_{lsa}(g,p) = \frac{2}{3} = F1_{lsa}(g,\hat{p})$.
    
    For $b=3$:
    \begin{align*}
        g &= 100 000 000 000\\
        p &= 100 100 000 000\\
        \hat{p}&= 100 101 000 000
    \end{align*}
    Then it holds $F1_{lsa}(g,p) = \frac{2}{3} = F1_{lsa}(g,\hat{p})$.
    
    \paragraph{\cref{prop:permutations}}
    Consider the following example:
    $g=\mathds{1}_{\{0\}}$, $p=\mathds{1}_{\{0\}}+ \mathds{1}_{\{b-1\}}$, and $\hat{p} = \mathds{1}_{\{0\}}+ \mathds{1}_{\{b\}}.$
    Then it holds $F1_{lsa}(g,p) = 1 > \frac{2}{3} = F1_{lsa}(g,\hat{p})$.
    
    For $b=3$:
    \begin{align*}
        g &= 100 000 000 000\\
        p &= 101 000 000 000\\
        \hat{p}&= 100 100 000 000
    \end{align*}
    Then it holds $F1_{lsa}(g,p) = 1 > \frac{2}{3} = F1_{lsa}(g,\hat{p})$.
    
    \paragraph{\cref{prop:trust}}
    Consider the following example:
    $g=\mathds{1}_{[0,2b-1]}$, $p=\mathds{1}_{\{b\}}$, and $\hat{p} = \mathds{1}_{\{0\}}+ \mathds{1}_{\{2b\}}.$
    Then it holds $F1_{lsa}(g,p) = \frac{2}{3} < \frac{4}{5} = F1_{lsa}(g,\hat{p})$.
    
    For $b=3$:
    \begin{align*}
        g &= 111 111 000 000\\
        p &= 000 100 000 000\\
        \hat{p}&= 100 000 100 000
    \end{align*}
    Then it holds $F1_{lsa}(g,p) = \frac{2}{3}< \frac{4}{5} = F1_{lsa}(g,\hat{p})$.
    
    \paragraph{\cref{prop:prediction_size}}
    Consider the following example:
    \begin{align*}
        g &= 110 000 000 000\\
        p &= 100 000 000 000\\
        \hat{p}&= 110 000 000 000
    \end{align*}
    Then it holds $F1_{lsa}(g,p) = 1 = F1_{lsa}(g,\hat{p})$.
    
    \paragraph{\cref{prop:temporal_order}}
    \textbf{Case $b=1$:}
    Let $W \in A(g)$, $\hat{p}_{[n] \setminus W} = p_{[n] \setminus W}$, $|A(\hat{p}_W)| = |A(p_W)|$, $|\hat{p}^{-1}(1)| = |p^{-1}(1)|$, and $\min\limits_{i \in W\colon p(i) = 1} i < \min\limits_{i \in W\colon \hat{p}(i) = 1} i$. Then due to the right-ward point adjusting of $p_{\delta_{PA}}, |\hat{p}_{\delta_{PA}}^{-1}(1) \cap g_\delta^{-1}(1)|>|p_{\delta_{PA}}^{-1}(1) \cap g_\delta^{-1}(1)|$ while $|\hat{p}_{\delta_{PA}}^{-1}(0) \cap g_\delta^{-1}(1)| < |p_{\delta_{PA}}^{-1}(0) \cap g_\delta^{-1}(1)|.$ Thus $F1_{lsa}(g,p) < F1_{lsa}(g,\hat{p})$.

    \textbf{Case $b\geq 2$:}
    Consider the following example:
    \begin{align*}
        g &= 110 000 000 000\\
        p &= 100 000 000 000\\
        \hat{p}&= 010 000 000 000
    \end{align*}
    Then it holds $F1_{lsa}(g,p) = 1 = F1_{lsa}(g,\hat{p})$.
    
    \paragraph{\cref{prop:early_bias}}
    Consider the following example:
    $g=\mathds{1}_{[0,2b-1]}$, $p=\mathds{1}_{\{0\}}+\mathds{1}_{\{2b-1\}}$, and $\hat{p} = \mathds{1}_{\{0\}}+ \mathds{1}_{\{b\}}.$
    Then it holds $F1_{lsa}(g,p) = 1 = F1_{lsa}(g,\hat{p})$.
    
    For $b=3$:
    \begin{align*}
        g &= 111 111 000 000\\
        p &= 100 001 000 000\\
        \hat{p}&= 100 100 000 000
    \end{align*}
    Then it holds $F1_{lsa}(g,p) = 1 = F1_{lsa}(g,\hat{p})$.
\end{proof}

\begin{theorem}\label{proof:pa-f1-k-percent}
    \textbf{Point-adjusted F1-Score at $k$\%} 
    \begin{align*}
        &F1_{PA\% k}(g,p) = \\
        &\frac{2\left(\sum\limits_{\stackrel{W \in A(g)\colon}{\frac{|p^{-1}(1) \cap W|}{|W|} > k}} |W| + \sum\limits_{\stackrel{W \in A(g)\colon}{\frac{|p^{-1}(1) \cap W|}{|W|} \leq k}} |p^{-1}(1) \cap W|\right)}{2\left(\sum\limits_{\stackrel{W \in A(g)\colon}{\frac{|p^{-1}(1) \cap W|}{|W|} > k}} |W| + \sum\limits_{\stackrel{W \in A(g)\colon}{\frac{|p^{-1}(1) \cap W|}{|W|} \leq k}} |p^{-1}(1) \cap W|\right) + |p^{-1}(1) \cap g^{-1}(0)| + \sum\limits_{\stackrel{W \in A(g)\colon}{\frac{|p^{-1}(1) \cap W|}{|W|} \leq k}} |p^{-1}(0) \cap W|}
    \end{align*}
    satisfies properties \cref{prop:detection} %\cref{prop:false_positives}, 
    and \cref{prop:permutations}.
    Furthermore, it satisfies \cref{prop:trust} for $k=0$.
\end{theorem}
\begin{proof}
    \item
    Let $\colon [n] \rightarrow \{0,1\}^n$ be the ground truth and $p,\hat{p} \colon [n] \rightarrow \{0,1\}^n$ predictions. We define:
    \[\#correct(p, k) := \sum\limits_{\stackrel{W \in A(g)\colon}{\frac{|p^{-1}(1) \cap W|}{|W|} > k}} |W| + \sum\limits_{\stackrel{W \in A(g)\colon}{\frac{|p^{-1}(1) \cap W|}{|W|} \leq k}} |p^{-1}(1) \cap W|\]
    Then \[F1_{PA\% k}(g,p) = \frac{2\#correct(p, k)}{2\#correct(p, k) + |p^{-1}(1) \cap g^{-1}(0)| + \sum\limits_{\stackrel{W \in A(g)\colon}{\frac{|p^{-1}(1) \cap W|}{|W|} \leq k}} |p^{-1}(0) \cap W|}\]
    
    \paragraph{\cref{prop:detection}} 
    Let $W \in A(g)$, $\hat{p}_{[n] \setminus W} = p_{[n] \setminus W}$, $p_W = 0$, and $|\hat{p}_W^{-1}(1)| > 0$. Then $\#correct(\hat{p}, k) > \#correct(p, k)$ while $|\hat{p}^{-1}(1) \cap g^{-1}(0)| = |p^{-1}(1) \cap g^{-1}(0)|$ and $\sum\limits_{\stackrel{W \in A(g)\colon}{\frac{|\hat{p}^{-1}(1) \cap W|}{|W|} \leq k}} |\hat{p}^{-1}(0) \cap W| < \sum\limits_{\stackrel{W \in A(g)\colon}{\frac{|p^{-1}(1) \cap W|}{|W|} \leq k}} |p^{-1}(0) \cap W|$. Thus $F1_{PA\% k}(g,\hat{p})> F1_{PA\% k}(g,p)$.
    
    \paragraph{\cref{prop:alarms}} 
     Consider the following example:
    \begin{align*}
        g &= 000 111 111 000\\
        p &= 000 111 000 000\\
        \hat{p}&= 000 111 011 000
    \end{align*}
    Then it holds
    \begin{itemize}
    \item For $k<0.5:$ $F1_{PA\% k}(g,p) = 1 = F1_{PA\% k}(g,\hat{p})$.
    \item For $0.5\leq k <\frac{5}{6}:$ $F1_{PA\% k}(g,p) = \frac{2}{3} < 1 = F1_{PA\% k}(g,\hat{p})$.
    \item For$\frac{5}{6}\leq k:$ $F1_{PA\% k}(g,p) = \frac{2}{3} < \frac{10}{11} = F1_{PA\% k}(g,\hat{p})$.
    \end{itemize}
    Regardless of the choice of $k$ this property does not hold.
    
    \paragraph{\cref{prop:false_positives}}
    Consider the following example:
    \begin{align*}
        g &= 000 000 000 001\\
        p &= 000 000 010 000\\
        \hat{p}&= 000 010 010 000
    \end{align*}
    Then it holds $F1_{PA\% k}(g,p) = 0 = F1_{PA\% k}(g,\hat{p})$.
    
    \paragraph{\cref{prop:false_positive_alarms}}
    Consider the following example:
    \begin{align*}
        g &= 000 000 111 000\\
        p &= 001 010 111 000\\
        \hat{p}&= 000 110 111 000
    \end{align*}
    Then it holds $F1_{PA\% k}(g,p) = \frac{3}{4} = F1_{PA\% k}(g,\hat{p})$.
    
    \paragraph{\cref{prop:permutations}}
    Let $W \in A(|g-1|)$, $\hat{p}_{[n] \setminus W} = p_{[n] \setminus W}$, $|\hat{p}_W^{-1}(1)| = |p_W^{-1}(1)|$, and $|A(\hat{p}_W)| = |A(p_W)|$. Then $\#correct(\hat{p}, k) = \#correct(p, k)$, $\sum\limits_{\stackrel{W \in A(g)\colon}{\frac{|\hat{p}^{-1}(1) \cap W|}{|W|} \leq k}} |\hat{p}^{-1}(0) \cap W| = \sum\limits_{\stackrel{W \in A(g)\colon}{\frac{|p^{-1}(1) \cap W|}{|W|} \leq k}} |p^{-1}(0) \cap W|$, and $|\hat{p}^{-1}(1) \cap g^{-1}(0)| = |p^{-1}(1) \cap g^{-1}(0)|$. Thus $F1_{PA\% k}(g,\hat{p})= F1_{PA\% k}(g,p)$.
    
    \paragraph{\cref{prop:trust}}
    \textbf{Case $k=0$}:
    In this case it holds $F1_{PA\% k}(g,p) = F1_{PA}(g,p)$ and thus by \cref{proof:pa-f1-score} this property holds.
    \\
    \textbf{Case $k > 0$}:
    Consider the following example: Let $2\leq k'\in\mathbb N$ such that $\frac{1}{k'}\leq k.$ Such a $k'$ exists because $k>0.$
    $g=\mathds{1}_{[a,a+k'-1]}$ with $a > 0$, $p=\mathds{1}_{\{a\}}$, and $\hat{p} = \mathds{1}_{\{0\}}+\mathds{1}_{[a,a+k'-1]}.$
    Then it holds $F1_{PA\% k}(g,p) = \frac{2}{k'+1} \leq \frac{k'}{k'+1} = F1_{PA\% k}(g,\hat{p})$.
    
    For $k \geq \frac{1}{3}$:
    \begin{align*}
        g &= 000 000 111 000\\
        p &= 000 000 100 000\\
        \hat{p}&= 000 010 111 000
    \end{align*}
    Then it holds $F1_{PA\% k}(g,p) = \frac{1}{2} < \frac{6}{7} = F1_{PA\% k}(g,\hat{p})$.
    
    \paragraph{\cref{prop:prediction_size}}
    \textbf{Case $k=1$}:
    In this case it holds $F1_{PA\% k}(g,p) = F1(g,p)$ and thus by \cref{proof:point-wise_f1-score} this property holds.

    \textbf{Case $k<1$}:
    Consider the following example: Let $k'\in\mathbb N$ such that $\frac{k'-1}{k'}> k.$ Such a $k'$ exists because $k<0.$
    $g=\mathds{1}_{[a,a+k'-1]}$ with $a > 0$, $p=\mathds{1}_{[a,a+k'-2]}$, and $\hat{p} = \mathds{1}_{[a,a+k'-1]}.$
    Then it holds $F1_{PA\% k}(g,p) = 1 = F1_{PA\% k}(g,\hat{p})$.

    For $k < \frac{5}{6}$:
    \begin{align*}
        g &= 000 111 111 000\\
        p &= 000 111 110 000\\
        \hat{p}&= 000 111 111 000
    \end{align*}
    Then it holds $F1_{PA\% k}(g,p) = 1 = F1_{PA\% k}(g,\hat{p})$.
    
    \paragraph{\cref{prop:temporal_order}}
    Consider the following example:
    \begin{align*}
        g &= 000 111 111 000\\
        p &= 000 111 000 000\\
        \hat{p}&= 000 000 111 000
    \end{align*}
    Then it holds
    \begin{itemize}
    \item For $k<\frac{1}{2}$ $F1_{PA\% k}(g,p) = 1 = F1_{PA\% k}(g,\hat{p})$.
    \item For $k\geq\frac{1}{2}$ $F1_{PA\% k}(g,p) = \frac{2}{3} = F1_{PA\% k}(g,\hat{p})$.
    \end{itemize}
    In either case the property does not hold.
    
    \paragraph{\cref{prop:early_bias}}
    Consider the following example:
    \begin{align*}
        g &= 000 111 111 000\\
        p &= 000 100 100 000\\
        \hat{p}&= 000 110 000 000
    \end{align*}
    Then it holds
    \begin{itemize}
    \item For $k<\frac{1}{3}$ $F1_{PA\% k}(g,p) = 1 = F1_{PA\% k}(g,\hat{p})$.
    \item For $k\geq\frac{1}{3}$ $F1_{PA\% k}(g,p) = \frac{1}{3} = F1_{PA\% k}(g,\hat{p})$.
    \end{itemize}
    In either case the property does not hold.
\end{proof}

\begin{theorem}\label{proof:pa-f1-k-percent-int}
    \textbf{Integrated point-adjusted F1-Score at $k$\%} 
    \begin{align*}
        &\int F1_{PA\% k}(g,p) \\
        &= \int_{0}^{1}\frac{2\left(\sum\limits_{\stackrel{W \in A(g)\colon}{\frac{|p^{-1}(1) \cap W|}{|W|} > k}} |W| + \sum\limits_{\stackrel{W \in A(g)\colon}{\frac{|p^{-1}(1) \cap W|}{|W|} \leq k}} |p^{-1}(1) + W|\right)}{2\left(\sum\limits_{\stackrel{W \in A(g)\colon}{\frac{|p^{-1}(1) \cap W|}{|W|} > k}} |W| + \sum\limits_{\stackrel{W \in A(g)\colon}{\frac{|p^{-1}(1) \cap W|}{|W|} \leq k}} |p^{-1}(1) + W|\right) + |p^{1}(1) \cap g^{-1}(0)| + \sum\limits_{\stackrel{W \in A(g)\colon}{\frac{|p^{-1}(1) \cap W|}{|W|} \leq k}} |p^{-1}(0) + W|} \;dk
    \end{align*}
    has properties \cref{prop:detection},
    \cref{prop:permutations}, and \cref{prop:prediction_size}.
\end{theorem}
\begin{proof}
    \item
    Let $\colon [n] \rightarrow \{0,1\}^n$ be the ground truth and $p,\hat{p} \colon [n] \rightarrow \{0,1\}^n$ predictions.
    
    \paragraph{\cref{prop:detection}} 
    Let $W \in A(g)$, $\hat{p}_{[n] \setminus W} = p_{[n] \setminus W}$, $p_W = 0$, and $|\hat{p}_W^{-1}(1)| > 0$.
    Consider \ref{proof:pa-f1-k-percent}, which implies that $F1_{PA\% k}(g,\hat{p}) > F1_{PA\% k}(g,p)$ for any $k$. Thus $\int F1_{PA\% k}(g,\hat{p}) > \int F1_{PA\% k}(g,p)$.
    
    \paragraph{\cref{prop:alarms}} 
    Consider the counter example in \ref{proof:pa-f1-k-percent}, which shows that $F1_{PA\% k}(g,\hat{p}) \geq F1_{PA\% k}(g,p)$ for any $k$. Thus $\int F1_{PA\% k}(g,\hat{p}) \geq \int F1_{PA\% k}(g,p)$ and this property does not hold.
    
    \paragraph{\cref{prop:false_positives}}
    Consider the following example:
    \begin{align*}
        g &= 000 000 000 001\\
        p &= 000 000 010 000\\
        \hat{p}&= 000 010 010 000
    \end{align*}
    Then it holds $\int F1_{PA\% k}(g,p) = 0 = \int F1_{PA\% k}(g,\hat{p})$.
    
    \paragraph{\cref{prop:false_positive_alarms}}
    Consider the counter example in \ref{proof:pa-f1-k-percent}, which shows that $F1_{PA\% k}(g,\hat{p}) = F1_{PA\% k}(g,p)$ for any $k$. Thus $\int F1_{PA\% k}(g,\hat{p}) = \int F1_{PA\% k}(g,p)$ and this property does not hold.
    
    \paragraph{\cref{prop:permutations}}
    Let $W \in A(|g-1|)$, $\hat{p}_{[n] \setminus W} = p_{[n] \setminus W}$, $|\hat{p}_W^{-1}(1)| = |p_W^{-1}(1)|$, and $|A(\hat{p}_W)| = |A(p_W)|$. Consider \ref{proof:pa-f1-k-percent}, which implies that $F1_{PA\% k}(g,\hat{p}) = F1_{PA\% k}(g,p)$ for any $k$. Thus $\int F1_{PA\% k}(g,\hat{p}) = \int F1_{PA\% k}(g,p)$.
    
    \paragraph{\cref{prop:trust}}
    Consider the following example:
    \begin{align*}
        g &= 000 000 111 000\\
        p &= 000 000 100 000\\
        \hat{p}&= 000 010 111 000
    \end{align*}
    Then for $k \geq \frac{1}{3}$ it holds $F1_{PA\% k}(g,p) = \frac{1}{2}$ and $ \frac{6}{7} = F1_{PA\% k}(g,\hat{p})$. For $k < \frac{1}{3}$ it holds that $F1_{PA\% k}(g,p) = 1$ and $ \frac{6}{7} = F1_{PA\% k}(g,\hat{p})$. From this we can follow that $\int F1_{PA\% k}(g,p) = \frac{1}{3}1+\frac{2}{3}\frac{1}{2} = \frac{2}{3}$ and $\int F1_{PA\% k}(g,\hat{p}) = \frac{6}{7}.$ Thus $\int F1_{PA\% k}(g,p) = \frac{2}{3} < \frac{6}{7} = \int F1_{PA\% k}(g,\hat{p}).$
    
    \paragraph{\cref{prop:prediction_size}}
    Let $W \in A(g)$, $\hat{p}_{[n] \setminus \{i^*\}} = p_{[n] \setminus \{i^*\}}$ for some $i^* \in W$, $\hat{p}(i^*) \neq p(i^*) = 0$, and $|A(\hat{p}_W)| \leq |A(p_W)|$. Let $w_1=min_{i\in W} i$, $w_2=max_{i\in W} i$, $c=|p^{-1}(1)\cap W|$, then $c+1=|\hat{p}^{-1}(1)\cap W|$.
    \begin{itemize}
    \item For $k<\frac{c}{|W|}$ $F1_{PA\% k}(g,p) = F1_{PA\% k}(g,\hat{p}).$
    \item For $\frac{c}{|W|}\leq k$ $F1_{PA\% k}(g,p) < F1_{PA\% k}(g,\hat{p}).$
    \end{itemize}
    From this we can follow that $\int F1_{PA\% k}(g,p) < \int F1_{PA\% k}(g,\hat{p}).$
    
    \paragraph{\cref{prop:temporal_order}}
    Consider the counter example in \ref{proof:pa-f1-k-percent}, which shows that $F1_{PA\% k}(g,\hat{p}) = F1_{PA\% k}(g,p)$ for any $k$. Thus $\int F1_{PA\% k}(g,\hat{p}) = \int F1_{PA\% k}(g,p)$ and this property does not hold.
    
    \paragraph{\cref{prop:early_bias}}
    Consider the counter example in \ref{proof:pa-f1-k-percent}, which shows that $F1_{PA\% k}(g,\hat{p}) = F1_{PA\% k}(g,p)$ for any $k$. Thus $\int F1_{PA\% k}(g,\hat{p}) = \int F1_{PA\% k}(g,p)$ and this property does not hold.
\end{proof}

\begin{theorem}\label{proof:pa-f1-df}
    \textbf{Point-Adjusted F1-score with Decay Function} 
    \[
    F1_{PAdf}(g,p) = \frac{2 \sum\limits_{\substack{W \in A(g): \\ W \cap p^{-1}(1) \neq \emptyset}} d^{(\min\limits_{i\in W\cap p^{-1}(1)}{i} - \min\limits_{i\in W} i)}|W|}{2\sum\limits_{\substack{W \in A(g): \\ W \cap p^{-1}(1) \neq \emptyset}} |W|+|p^{-1}(1)\cap g^{-1}(0)| + \sum\limits_{\substack{W \in A(g): \\ W \cap p^{-1}(1) = \emptyset}} |W|}
    \]
    where: \\
    $d \in (0, 1]$ = decay rate \\
    has properties \cref{prop:detection} and \cref{prop:permutations}. It also has property \cref{prop:temporal_order} for $d<1$ or \cref{prop:trust} if $d=1$.
\end{theorem}
\begin{proof}
    \item
    Let g$\colon [n] \rightarrow \{0,1\}^n$ be the ground truth and $p,\hat{p} \colon [n] \rightarrow \{0,1\}^n$ predictions.
    
    \paragraph{\cref{prop:detection}}
    \textbf{Case $d=1$} Then $F1_{PAdf}(g,p) = F1_{PA}(g,p)$ and this property is fulfilled following \cref{proof:pa-f1-score}.

    \textbf{Case $d<1$}
    Consider the following example:
    Let $a\in\mathbb N$ be large enough such that $\frac{1}{a+1} > d^{a+2}.$
    $g=\mathds{1}_{\{0\}}+\mathds{1}_{[2,a]}$, $p=\mathds{1}_{\{0\}}$, and $\hat{p} = \mathds{1}_{\{0\}}+ \mathds{1}_{\{a\}}.$
    Then it holds $F1_{PAdf}(g,p) = \frac{2}{a+1} > \frac{2+2d^{a-2}(a-1)}{2a} = F1_{PAdf}(g,\hat{p})$.

    For example for $d\leq 0.7$ (this example also holds for some more $0.7<d<0.8$, however, the exact value is irrational):
    \begin{align*}
        g &= 101 111 111 111\\
        p &= 100 000 000 000\\
        \hat{p}&= 100 000 000 001
    \end{align*}
    Then it holds $F1_{PAdf}(g,p) = \frac{2}{12} > \frac{2+20d^9}{22} = F1_{PAdf}(g,\hat{p})$.

    \paragraph{\cref{prop:alarms}} 
    Consider the following example:
    \begin{align*}
        g &= 000 111 111 000\\
        p &= 000 100 000 000\\
        \hat{p}&= 000 100 001 000
    \end{align*}
    Then it holds $F1_{PAdf}(g,p) = 1 = F1_{PAdf}(g,\hat{p})$.
    
    \paragraph{\cref{prop:false_positives}} 
    Consider the following example:
    \begin{align*}
        g &= 000 111 111 000\\
        p &= 000 000 000 000\\
        \hat{p}&= 010 000 000 000
    \end{align*}
    Then it holds $F1_{PAdf}(g,p) = 0 = F1_{PAdf}(g,\hat{p})$.
    \paragraph{\cref{prop:false_positive_alarms}}
    Consider the following example:
    \begin{align*}
        g &= 000 111 111 000\\
        p &= 001 000 000 000\\
        \hat{p}&= 101 000 000 000
    \end{align*}
    Then it holds $F1_{PAdf}(g,p) = 0 = F1_{PAdf}(g,\hat{p})$.
    
    \paragraph{\cref{prop:permutations}}
    Let $W \in A(|g-1|)$, $\hat{p}_{[n] \setminus W} = p_{[n] \setminus W}$, $|\hat{p}_W^{-1}(1)| = |p_W^{-1}(1)|$, and $|A(\hat{p}_W)| = |A(p_W)|$. Then every component is equal and thus $F1_{PAdf}(g,p) = F1_{PAdf}(g,\hat{p}).$
    
    \paragraph{\cref{prop:trust}}
    \textbf{Case $d=1$} Then $F1_{PAdf}(g,p) = F1_{PA}(g,p)$ and this property is fulfilled following \cref{proof:pa-f1-score}.

    \textbf{Case $d<1$}
    Consider the following example:
    Let $a\in\mathbb N$ be large enough such that $d^{a-2} < \frac{2a-2}{2a-1}.$
    $g=\mathds{1}_{\{0\}}+\mathds{1}_{[2,a]}$, $p=\mathds{1}_{\{a\}}$, and $\hat{p} = \mathds{1}_{\{0\}}+ \mathds{1}_{\{2\}}.$
    Then it holds $F1_{PAdf}(g,p) = d^{a-2} < \frac{2a-2}{2a-1} = F1_{PAdf}(g,\hat{p})$.

    For example for $d\leq 0.99$:
    \begin{align*}
        g &= 001 111 111 111\\
        p &= 000 000 000 001\\
        \hat{p}&= 101 000 000 000
    \end{align*}
    Then it holds $F1_{PAdf}(g,p) = d^9 < \frac{20}{21} = F1_{PAdf}(g,\hat{p})$.
    
    \paragraph{\cref{prop:prediction_size}}
    Consider the following example:
    \begin{align*}
        g &= 000 111 111 000\\
        p &= 000 100 000 000\\
        \hat{p}&= 000 110 000 000
    \end{align*}
    Then it holds $F1_{PAdf}(g,p) = 1 = F1_{PAdf}(g,\hat{p})$.
    
    \paragraph{\cref{prop:temporal_order}}
    \textbf{Case $d=1$} Then $F1_{PAdf}(g,p) = F1_{PA}(g,p)$ and this property is not fulfilled following \cref{proof:pa-f1-score}.

    \textbf{Case $d<1$}
    Let $W \in A(g)$, $\hat{p}_{[n] \setminus W} = p_{[n] \setminus W}$, $|A(\hat{p}_W)| = |A(p_W)|$, $|\hat{p}^{-1}(1)| = |p^{-1}(1)|$, and $\min\limits_{i \in W\colon p(i) = 1} i < \min\limits_{i \in W\colon \hat{p}(i) = 1} i$. Then $(\min\limits_{i\in W\cap \hat{p}^{-1}(1)}{i} - \min\limits_{i\in W} i) < (\min\limits_{i\in W\cap p^{-1}(1)}{i} - \min\limits_{i\in W} i),$ while all other components remain the same. Thus $F1_{PAdf}(g,p) < F1_{PAdf}(g,\hat{p}).$
    
    \paragraph{\cref{prop:early_bias}}
    \textbf{Case $d=1$} Then $F1_{PAdf}(g,p) = F1_{PA}(g,p)$ and this property is not fulfilled following \cref{proof:pa-f1-score}.

    \textbf{Case $d<1$}
    Consider the following example:
    \begin{align*}
        g &= 000 111 111 000\\
        p &= 000 100 001 000\\
        \hat{p}&= 000 110 000 000
    \end{align*}
    Then it holds $F1_{PAdf}(g,p) = 1 = F1_{PAdf}(g,\hat{p})$.
\end{proof}

\begin{theorem}\label{proof:pa-f1-reduced-length}
    \textbf{Reduced-Length F1-Score} \[
        F1_{reduced-length}(g,p) = \frac{2\sum\limits_{\stackrel{W \in A(g)\colon}{W \cap p^{-1}(1) \neq \emptyset}} \ln |W|}{2\sum\limits_{\stackrel{W \in A(g)\colon}{W \cap p^{-1}(1) \neq \emptyset}} \ln |W| + |p^{-1}(1) \cap g^{-1}(0)| + \sum\limits_{\stackrel{W \in A(g)\colon}{W \cap p^{-1}(1) = \emptyset}}\ln |W|}
    \]
    has properties \cref{prop:detection}, %\cref{prop:false_positives}, 
    \cref{prop:permutations}, and \cref{prop:trust}.
\end{theorem}
\begin{proof}
    \item
    Let $\colon [n] \rightarrow \{0,1\}^n$ be the ground truth and $p,\hat{p} \colon [n] \rightarrow \{0,1\}^n$ predictions.
    
    \paragraph{\cref{prop:detection}} 
    Let $W \in A(g)$ an interval, $\hat{p}_{[n] \setminus W} = p_{[n] \setminus W}$, $p_W = 0$, and $|\hat{p}_W^{-1}(1)| > 0$.
    Then $\{W \in A(g) \colon W \cap \hat{p}^{-1}(1) \neq \emptyset\} = \{W \in A(g) \colon W \cap p^{-1}(1) \neq \emptyset\} \cup \{W\}$ with $|W| > 0$.
    Thus, $F1_{reduced-length}(g, p) < F1_{reduced-length}(g, \hat{p})$.
    
    \paragraph{\cref{prop:alarms}} 
    Consider the following example:
    \begin{align*}
        g &= 000 111 111 000\\
        p &= 000 111 000 000\\
        \hat{p}&= 000 111 011 000
    \end{align*}
    Then it holds $F1_{reduced-length}(g,p) = 1 = F1_{reduced-length}(g,\hat{p})$.
    
    \paragraph{\cref{prop:false_positives}}
    Consider the following example:
    \begin{align*}
        g &= 000 000 000 001\\
        p &= 000 000 010 000\\
        \hat{p}&= 000 010 010 000
    \end{align*}
    Then it holds $F1_{reduced-length}(g,p) = 0 = F1_{reduced-length}(g,\hat{p})$.
    
    \paragraph{\cref{prop:false_positive_alarms}}
    Consider the following example:
    \begin{align*}
        g &= 000 000 111 000\\
        p &= 010 010 010 000\\
        \hat{p}&= 001 100 010 000
    \end{align*}
    Then it holds $F1_{reduced-length}(g,p) = \frac{\ln 3}{1+\ln 3} = F1_{reduced-length}(g,\hat{p})$.
    
    \paragraph{\cref{prop:permutations}}
    Let $W \subset W' \in A(|g-1|)$, $\hat{p}_{[n] \setminus W} = p_{[n] \setminus W}$, $|\hat{p}_W^{-1}(1)| = |p_W^{-1}(1)|$, and $|A(\hat{p}_W)| = |A(p_W)|$.
    Then $\{W \in A(g) \colon W \cap \hat{p}^{-1}(1) \neq \emptyset\} = \{W \in A(g) \colon W \cap p^{-1}(1) \neq \emptyset\}$ and $|\hat{p}^{-1}(1) \cap g^{-1}(0)| = |p^{-1}(1) \cap g^{-1}(0)|$.
    Thus, $F1_{reduced-length}(g,p) = F1_{reduced-length}(g,\hat{p})$.
    
    \paragraph{\cref{prop:trust}}
    Let $W \in A(g)$, $\hat{W} \in A(|g-1|)$, $|A(\hat{p}_{W})| = |A(p_{W})|$, $\hat{p}_{[n] \setminus (W \cup \hat{W})} = p_{[n] \setminus (W \cup \hat{W})}$, $p_{\hat{W}} = 0$, and $\hat{p}_{\hat{W}} = \mathds{1}_{\{i^*\}}$ for some $i^* \in \hat{W}$. Then $\{W \in A(g) \colon W \cap \hat{p}^{-1}(1) \neq \emptyset\} = \{W \in A(g) \colon W \cap p^{-1}(1) \neq \emptyset\}$ and $|\hat{p}^{-1}(1) \cap g^{-1}(0)| = |p^{-1}(1) \cap g^{-1}(0)|+1$. Thus, $F1_{reduced-length}(g,p) > F1_{reduced-length}(g,\hat{p})$.
    
    \paragraph{\cref{prop:prediction_size}}
    Consider the following example:
    \begin{align*}
        g &= 000 111 111 000\\
        p &= 000 010 000 000\\
        \hat{p}&= 000 011 000 000
    \end{align*}
    Then it holds $F1_{reduced-length}(g,p) = 1 = F1_{reduced-length}(g,\hat{p})$.
    
    \paragraph{\cref{prop:temporal_order}}
    Consider the following example:
    \begin{align*}
        g &= 000 111 111 000\\
        p &= 000 110 000 000\\
        \hat{p}&= 000 011 000 000
    \end{align*}
    Then it holds $F1_{reduced-length}(g,p) = 1 = F1_{reduced-length}(g,\hat{p})$.
    
    \paragraph{\cref{prop:early_bias}}
    Consider the following example:
    \begin{align*}
        g &= 000 111 111 000\\
        p &= 000 010 010 000\\
        \hat{p}&= 000 110 000 000
    \end{align*}
    Then it holds $F1_{reduced-length}(g,p) = 1 = F1_{reduced-length}(g,\hat{p})$.
\end{proof}

\begin{theorem}\label{proof:pa-f1-ba}
    \textbf{Balanced Point Adjusted F1-Score} \[
        F1_{BA}(g,p) = \frac{2TP(g,p)}{2TP(g,p) + FP(g,p) + FN(g,p)}
    \]
    where $B\in 2\mathbb N$ is an even integer and
    \begin{align*}
        TP(g,p) &= \sum\limits_{\stackrel{W \in A(g)\colon}{W \cap p^{-1}(1) \neq \emptyset}} |W| + \sum\limits_{\stackrel{W \in A(g)\colon}{W \cap p^{-1}(1) = \emptyset}} |\{i \in W \colon p^{-1}(1) \cap g^{-1}(0) \cap [i - \frac{B}{2}, i + \frac{B}{2} + 1]\}| \neq \emptyset\}|\\
        FP(g,p) &= |\{i \in p^{-1}(0) \cap g^{-1}(0) \colon p^{-1}(1) \cap g^{-1}(0) \cap [i - \frac{B}{2}, i + \frac{B}{2} + 1] \neq \emptyset\}| + |p^{-1}(1) \cap g^{-1}(0)|\\
        FN(g,p) &= \sum\limits_{\stackrel{W \in A(g)\colon}{W \cap p^{-1}(1) = \emptyset}} |\{i \in W \colon p^{-1}(1) \cap g^{-1}(0) \cap [i - \frac{B}{2}, i + \frac{B}{2} + 1] = \emptyset\}|
    \end{align*}
    has no properties.
\end{theorem}
\begin{proof}
    \item
    For simplicity we slightly modify the above definition to use just an integer $b\in \mathbb N$ and we consider the intersection with intervals above to be entirely balanced. Namely, the intervals of the form $[i - \frac{B}{2}, i + \frac{B}{2} + 1]$ are changed to $[i - B, i + B].$ This changes the definition as follows:
    \begin{align*}
        TP(g,p) &= \sum\limits_{\stackrel{W \in A(g)\colon}{W \cap p^{-1}(1) \neq \emptyset}} |W| + \sum\limits_{\stackrel{W \in A(g)\colon}{W \cap p^{-1}(1) = \emptyset}} |\{i \in W \colon p^{-1}(1) \cap g^{-1}(0) \cap [i - B, i + B]\}| \neq \emptyset\}|\\
        FP(g,p) &= |\{i \in p^{-1}(0) \cap g^{-1}(0) \colon p^{-1}(1) \cap g^{-1}(0) \cap [i - B, i + B] \neq \emptyset\}| + |p^{-1}(1) \cap g^{-1}(0)|\\
        FN(g,p) &= \sum\limits_{\stackrel{W \in A(g)\colon}{W \cap p^{-1}(1) = \emptyset}} |\{i \in W \colon p^{-1}(1) \cap g^{-1}(0) \cap [i - B, i + B] = \emptyset\}|,
    \end{align*}
    where $B\in\mathbb N$. All the following proof can be modified to prove the original definition's lack of properties by slightly modifying each of the counterexamples. We leave this slight modification to the reader.
    
    Let $\colon [n] \rightarrow \{0,1\}^n$ be the ground truth and $p,\hat{p} \colon [n] \rightarrow \{0,1\}^n$ predictions.
    
    \paragraph{\cref{prop:detection}}
    Consider the following example:
    $g=\mathds{1}_{\{a\}\cup[a+2B+1,a+4B+1]}$ with $a \geq B$, $p=\mathds{1}_{\{a+3B+1\}}$, and $\hat{p} = \mathds{1}_{\{a\}}+ \mathds{1}_{\{a+3B+1\}}.$
    Then it holds $F1_{BA}(g,p) = \frac{2(2B+1)}{2(2B+1)+1} > \frac{2(2B+2)}{2(2B+2)+2B} = F1_{BA}(g,\hat{p})$.

    For $B=1$:
    \begin{align*}
        g &= 000 001 000 111\\
        p &= 000 000 000 010\\
        \hat{p}&= 000 001 000 010
    \end{align*}
    Then it holds $F1_{BA}(g,p) = \frac{6}{7} > \frac{4}{5} = F1_{BA}(g,\hat{p})$.
    
    \paragraph{\cref{prop:alarms}}
    Consider the following example:
    $g=\mathds{1}_{[a,a+4B+1]}$ with $a > 0$, $p=\mathds{1}_{\{a\}}$, and $\hat{p} = \mathds{1}_{\{a\}}+ \mathds{1}_{\{a+2B+1\}}.$
    Then it holds $F1_{BA}(g,p) = 1 = F1_{BA}(g,\hat{p})$.
    
    For $B\leq 1$:
    \begin{align*}
        g &= 000 111 111 000\\
        p &= 000 010 000 000\\
        \hat{p}&= 000 010 010 000
    \end{align*}
    Then it holds $F1_{BA}(g,p) = 1 = F1_{BA}(g,\hat{p})$.
    
    \paragraph{\cref{prop:false_positives}}
    Consider the following example:
    $g=\mathds{1}_{\{a\}}$ with $a > B+1$, $p=\mathds{1}_{\{a-B-1\}}$, and $\hat{p} = \mathds{1}_{[a-B-1,a-B]}.$
    Then it holds $F1_{BA}(g,p) = 0 < \frac{2}{2B+3} = F1_{BA}(g,\hat{p})$.
    
    For $B = 1$:
    \begin{align*}
        g &= 000 000 100 000\\
        p &= 000 010 000 000\\
        \hat{p}&= 000 011 000 000
    \end{align*}
    Then it holds $F1_{BA}(g,p) = 0 < \frac{2}{5} = F1_{BA}(g,\hat{p})$.
    \paragraph{\cref{prop:false_positive_alarms}}
    Consider the following example:
    \begin{align*}
        g &= 000 000 000 000\\
        p &= 000 010 100 000\\
        \hat{p}&= 000 011 000 000
    \end{align*}
    Then it holds $F1_{BA}(g,p) = 0 = F1_{BA}(g,\hat{p})$.
    
    \paragraph{\cref{prop:permutations}}
    Consider the following example:
    $g=\mathds{1}_{\{a\}}$ with $a > B+1$, $p=\mathds{1}_{\{a-B-1\}}$, and $\hat{p} = \mathds{1}_{\{a-B\}}.$
    Then it holds $F1_{BA}(g,p) = 0 < \frac{2}{2B+2} = F1_{BA}(g,\hat{p})$.
    
    For $B = 1$:
    \begin{align*}
        g &= 000 000 100 000\\
        p &= 000 010 000 000\\
        \hat{p}&= 000 001 000 000
    \end{align*}
    Then it holds $F1_{BA}(g,p) = 0 < \frac{1}{2} = F1_{BA}(g,\hat{p})$.
    
    \paragraph{\cref{prop:trust}}
    Consider the following example:
    $g=\mathds{1}_{[a,a+B]}+\mathds{1}_{[a+B+2,a+2B+2]}$ with $a > B$, $p=\mathds{1}_{\{a, a+2B+2\}}$, and $\hat{p} = \mathds{1}_{[a+B, a+B+2]}.$
    Then it holds $F1_{BA}(g,p) = \frac{4}{5} < \frac{8B}{8B+1} = F1_{BA}(g,\hat{p})$.
    
    For $B = 1$:
    \begin{align*}
        g &= 001 101 100 000\\
        p &= 001 000 100 000\\
        \hat{p}&= 000 111 000 000
    \end{align*}
    Then it holds $F1_{BA}(g,p) = \frac{4}{5} < \frac{8}{9} = F1_{BA}(g,\hat{p})$.
    
    \paragraph{\cref{prop:prediction_size}}
    Consider the following example:
    $g=\mathds{1}_{[a,a+2B]}$ with $a > B$, $p=\mathds{1}_{\{a+B\}}$, and $\hat{p} = \mathds{1}_{[a+B, a+B+1]}.$
    Then it holds $F1_{BA}(g,p) = 1 > \frac{2(2B+1)}{2(2B+1)+1} = F1_{BA}(g,\hat{p})$.
    
    For $B = 1$:
    \begin{align*}
        g &= 000 111 000 000\\
        p &= 000 010 000 000\\
        \hat{p}&= 000 011 000 000
    \end{align*}
    Then it holds $F1_{BA}(g,p) = 1 > \frac{6}{7} = F1_{BA}(g,\hat{p})$.
    
    \paragraph{\cref{prop:temporal_order}}
    Consider the following example:
    $g=\mathds{1}_{[a,a+2B+1]}$ with $a > 0$, $p=\mathds{1}_{\{a+B+1\}}$, and $\hat{p} = \mathds{1}_{\{a+B\}}.$
    Then it holds $F1_{BA}(g,p) = 1 = F1_{BA}(g,\hat{p})$.
    
    For $B = 1$:
    \begin{align*}
        g &= 000 111 100 000\\
        p &= 000 001 000 000\\
        \hat{p}&= 000 010 000 000
    \end{align*}
    Then it holds $F1_{BA}(g,p) = 1 = F1_{BA}(g,\hat{p})$.
    
    \paragraph{\cref{prop:early_bias}}
    Consider the following example:
    $g=\mathds{1}_{[a,a+2B+2]}$ with $a > 0$, $p=\mathds{1}_{\{a+B, a+B+2\}}$, and $\hat{p} = \mathds{1}_{[a+B,a+B+1]}.$
    Then it holds $F1_{BA}(g,p) = 1 = F1_{BA}(g,\hat{p})$.
    
    For $B = 1$:
    \begin{align*}
        g &= 000 111 110 000\\
        p &= 000 010 100 000\\
        \hat{p}&= 000 011 000 000
    \end{align*}
    Then it holds $F1_{BA}(g,p) = 1 = F1_{BA}(g,\hat{p})$.
\end{proof}

\begin{theorem}\label{proof:ts-precision}
    \textbf{Range-Based Precision} defined as 
    \[
        Precision_{TS}(g, p) = \frac{1}{|A(p)|}\sum\limits_{W \in A(p)}\gamma'(W,p) \sum\limits_{W_g\in A(g)} \omega(W, W \cap W_g, \delta)
    \]
    where
    \begin{align*}
    \delta(i;W)
    &=\;
    |W|- \bigl(i - \min(W)\bigr) + 1 \geq 1,\\
    %\quad\text{(``front'' bias)} ,
    0\leq\omega\bigl(W,S,\delta\bigr)
    &=
    \frac{
      \sum_{\,i\in W}\;\delta(i;W)\,\mathds{1}_{\{i\in S\}}
    }{
      \sum_{\,i\in W}\;\delta(i;W)
    }\leq 1,\\
    k(W, g) &= |\{W_g \in A(g) \colon W \cap W_g \neq \emptyset\}|,\\
    \gamma'(W, g) &= \frac{1}{k(W, g)}\mathds{1}_{k(W, g) > 1}(W) + \mathds{1}_{k(W, g) \leq 1}(W).
    \end{align*}

\end{theorem}
\begin{proof}
    \item
    Notice that for $\delta$ usually $i\in W=[\ell,u]$ and thus $\delta(i;W)=u-(i-1)+1$. 
    Let $g\colon [n] \rightarrow \{0,1\}^n$ be the ground truth and $p,\hat{p} \colon [n] \rightarrow \{0,1\}^n$ predictions.

        \paragraph{\cref{prop:detection}} 
        Consider the following example:
        \begin{align*}
            g &= 000 111 000 111\\
            p &= 000 111 000 000\\
            \hat{p}&= 000 111 000 001
        \end{align*}
        Then it holds $\mathrm{Precision}_{TS}(g,p) = 1 = \mathrm{Precision}_{TS}(g,\hat{p})$.

        \paragraph{\cref{prop:alarms}} 
        Consider the following example:
        \begin{align*}
            g &= 000 111 111 000\\
            p &= 000 100 000 000\\
            q &= 000 110 011 000
        \end{align*}
        
        Then it holds
        $
        \mathrm{Precision}_{TS}(g,p) = 1
        =
        \mathrm{Precision}_{TS}(g,q).
        $

        \paragraph{\cref{prop:false_positives}} 
        Consider the following example:
        \[
        \begin{aligned}
        g &= 000 000 000 000,\\
        p &= 000 001 100 000,\\
        q &= 000 011 100 000.
        \end{aligned}
        \]
 
        $
        \mathrm{Precision}_{TS}(g,p) = 0
        =
        \mathrm{Precision}_{TS}(g,q).
        $
        \paragraph{\cref{prop:false_positive_alarms}}

        Consider the following example:
        \[
        \begin{aligned}
        g &= 000 000 000 001,\\
        p &= 000 100 000 000,\\
        q &= 000 100 100 000.
        \end{aligned}
        \]
        Then it follows
        $
        \mathrm{Precision}_{TS}(g,p) = 0
        =
        \mathrm{Precision}_{TS}(g,q).
        $
        
        \paragraph{\cref{prop:permutations}}
        Consider the following example:
        \[
        \begin{aligned}
        g &= 000 011 110 000,\\
        p &= 110 011 000 000,\\
        q &= 001 111 000 000.
        \end{aligned}
        \]
        Then it holds        
        $
        \mathrm{Precision}_{TS}(g,p) = 0.5
        > 0.3 =
        \mathrm{Precision}_{TS}(g,q)
        $
        
        \paragraph{\cref{prop:trust}}
        Let $W \in A(g)$, $\hat{W} \in A(|g-1|)$, $|A(\hat{p}_{W})| = |A(p_{W})|$, $\hat{p}_{[n] \setminus (W \cup \hat{W})} = p_{[n] \setminus (W \cup \hat{W})}$, $p_{\hat{W}} = 0$, and $\hat{p}_{\hat{W}} = \mathds{1}_{\{i^*\}}$ for some $i^* \in \hat{W}$.
        Consider the following example:
        \[
        \begin{aligned}
        g &= 011 111 111 110,\\
        p &= 000 000 001 011,\\
        q &= 111 111 111 011.
        \end{aligned}
        \]
        Then it follows
        $
        \mathrm{Precision}_{TS}(g,p) = 0.6667 < 
        0.7222 = \mathrm{Precision}_{TS}(g,q).
        $
        
        \paragraph{\cref{prop:prediction_size}}
        Consider the following example:
        \[
        \begin{aligned}
        g &= 000 111 000 000,\\
        p &= 000 110 000 000,\\
        q &= 000 100 000 000.
        \end{aligned}
        \]
        Then it holds
        $
        \mathrm{Precision}_{TS}(g,p) = 1
        =
        \mathrm{Precision}_{TS}(g,q).
        $
        
        \paragraph{\cref{prop:temporal_order}}
        Consider the following example:
        \[
        \begin{aligned}
        g &= 000 111 100 000,\\
        p &= 000 110 000 000,\\
        q &= 000 011 000 000.
        \end{aligned}
        \]
        Then it holds
        $
        \mathrm{Precision}_{TS}(g,p) = 1.0
        =
        \mathrm{Precision}_{TS}(g,q).
        $
        
        \paragraph{\cref{prop:early_bias}}
        Consider the following example:
        \[
        \begin{aligned}
        g &= 000 111 100 000,\\
        p &= 000 100 000 000,\\
        q &= 000 000 100 000.
        \end{aligned}
        \]
        Then it holds
        $
        \mathrm{Precision}_{TS}(g,p) = 1.0
        =
        \mathrm{Precision}_{TS}(g,q).
        $
\end{proof}

\begin{theorem}\label{proof:ts-recall}
    \textbf{Range-Based Recall} defined as 
    \[
        Recall_{TS}(g, p) = \frac{1}{|A(g)|}\sum\limits_{W \in A(g)}(\alpha\mathds{1}_{\exists W_p\in A(p):W\cap W_p \neq \emptyset}+(1-\alpha)\gamma'(W, p) \sum\limits_{W_p\in A(p)} \omega(W, W \cap W_p, \delta ))
    \]
    where $0\leq \alpha \leq 1$,
    \begin{align*}
    \delta(i;W)
    &=\;
    |W|- \bigl(i - \min(W)\bigr) + 1 \geq 1,\\
    %\quad\text{(``front'' bias)} ,
    0\leq\omega\bigl(W,S,\delta\bigr)
    &=
    \frac{
      \sum_{\,i\in W}\;\delta(i;W)\,\mathds{1}_{\{i\in S\}}
    }{
      \sum_{\,i\in W}\;\delta(i;W)
    }\leq 1,\\
    k(W, p) &= |\{W_p \in A(p) \colon W \cap W_p \neq \emptyset\}|,\\
    \gamma'(W, p) &= \frac{1}{k(W, p)}\mathds{1}_{k(W, p) > 1}(W) + \mathds{1}_{k(W, p) \leq 1}(W).
    \end{align*}
\end{theorem}
\begin{proof}
    \item
    Let $g\colon [n] \rightarrow \{0,1\}^n$ be the ground truth and $p,\hat{p} \colon [n] \rightarrow \{0,1\}^n$ predictions.
    \paragraph{\cref{prop:detection}} 
   
    Let $W \in A(g)$, $\hat{p}_{[n]\setminus W} = p_{[n]\setminus W}$, $p_W=0$, and $\lvert\hat{p}_W^{-1}(1)\rvert > 0$. Then there is $i_0\in W$ with $\hat{p}(i_0)=1$. Then 
    \begin{align*}
    0&=\alpha\mathds{1}_{\exists W_p\in A(p):W\cap W_p \neq \emptyset}+(1-\alpha)\gamma'(W, p) \sum\limits_{W_p\in A(p)} \omega(W, W \cap W_p, \delta )\\
    0&<\alpha\mathds{1}_{\exists W_{\hat{p}}\in A(\hat{p}):W\cap W_{\hat{p}} \neq \emptyset}+(1-\alpha)\gamma'(W, \hat{p}) \sum\limits_{W_{\hat{p}}\in A(\hat{p})} \omega(W, W \cap W_{\hat{p}}, \delta )
    \end{align*}
    While all other summands are equal and thus $\mathrm{Recall}_{TS}(g,\hat{p})>\mathrm{Recall}_{TS}(g,p).$

    \paragraph{\cref{prop:alarms}} 
    Consider the following example:
    \begin{align*}
        g &= 111111111111\\
        p &= 100000000000\\
        \hat{p}&= 110011111111
    \end{align*}
    Then it holds $Recall_{TS}(g, p) = 0.1667  < 0.8846  = Recall_{TS}(g, \hat{p})$.

    \paragraph{\cref{prop:false_positives}} 
    Consider the following example:
    \begin{align*}
        g &= 000 001 000 000\\
        p &= 000 010 000 000\\
        \hat{p}&= 000 110 000 000
    \end{align*}
    Then it holds $Recall_{TS}(g, p) = 0 = Recall_{TS}(g, \hat{p})$.

    \paragraph{\cref{prop:false_positive_alarms}}
    Consider the following example:
    \begin{align*}
        g &= 000 010 000 000\\
        p &= 000 001 000 000\\
        \hat{p}&= 000 101 000 000
    \end{align*}
    Then it holds $Recall_{TS}(g, p) = 0 = Recall_{TS}(g, \hat{p})$.

    \paragraph{\cref{prop:permutations}}
    Let $W \in A(|g-1|)$, $\hat{p}_{[n]\setminus W} = p_{[n]\setminus W}$, 
    $\lvert\hat{p}_W^{-1}(1)\rvert = \lvert p_W^{-1}(1)\rvert,$ 
    and $\lvert A(\hat{p}_W)\rvert = \lvert A(p_W)\rvert.$
    Then, since false positives do not affect the metric at all, $Recall_{TS}(g, p) = Recall_{TS}(g, \hat{p})$.
    
    \paragraph{\cref{prop:trust}}
     Consider the following example:
     \begin{align*}
         g &=  000010000000\\
         p &= 000001000000\\
         \hat{p}&= 000101000000
     \end{align*}
     Then it holds $Recall_{TS}(g, p) = 0 = Recall_{TS}(g, q)$.
    
    \paragraph{\cref{prop:prediction_size}}
    Let $W \in A(g)$, $i^*\in W$, $\hat{p}_{[n]\setminus\{i^*\}}=p_{[n]\setminus\{i^*\}},$ 
    $p(i^*)=0,\;\hat{p}(i^*)=1,$ 
    and $|A(\hat{p}_W)|\le|A(p_W)|.$
    Then,
    \begin{align*}
    &(\alpha\mathds{1}_{\exists W_p\in A(p):W\cap W_p \neq \emptyset}+(1-\alpha)\gamma'(W, p) \sum\limits_{W_p\in A(p)} \omega(W, W \cap W_p, \delta ))\\
    \leq&(\alpha\mathds{1}_{\exists W_{\hat{p}}\in A(\hat{p}):W\cap W_{\hat{p}} \neq \emptyset}+(1-\alpha)\gamma'(W, \hat{p}) \sum\limits_{W_p\in A(p)} \omega(W, W \cap W_p, \delta ))\\
    <&(\alpha\mathds{1}_{\exists W_{\hat{p}}\in A(\hat{p}):W\cap W_{\hat{p}} \neq \emptyset}+(1-\alpha)\gamma'(W, \hat{p}) \sum\limits_{W_{\hat{p}}\in A(\hat{p})} \omega(W, W \cap W_{\hat{p}}, \delta ))\\
    \end{align*}
    while all other summands remain the same. Thus $Recall_{TS}(g, p) < Recall_{TS}(g, \hat{p})$.
    
    \paragraph{\cref{prop:temporal_order}}
    Consider the following example:
    \begin{align*}
        g &= 111111111111\\
        p &= 100000000001 \\
        \hat{p}&= 010100000000 
    \end{align*}
    Then it holds $Recall_{TS}(g, p) =  0.1923  < 0.2821 = Recall_{TS}(g, \hat{p})$.

    \paragraph{\cref{prop:early_bias}}
    Let $W \in A(g)$, $\hat{p}_{[n] \setminus W} = p_{[n] \setminus W}$,
    Let $p, \hat{p}$ be two predictions with $|A(\hat{p}_W)| \leq |A(p_W)|$, let $i^{*}\in W$ with $p(i^*) = 1$, let $i^{**} < i^* \in W$ with $p(i^{**}) = 0$, and $p_W'(i) = \mathds{1}_{\{i^{**}\}}(i) + p(i) \mathds{1}_{W\setminus \{i^*, i^{**}\}}(i)$. Then,
    \begin{align*}
    &\alpha\mathds{1}_{\exists W_p\in A(p):W\cap W_p \neq \emptyset}+(1-\alpha)\gamma'(W, p) \sum\limits_{W_p\in A(p)} \omega(W, W \cap W_p, \delta )\\
    \leq&\alpha\mathds{1}_{\exists W_{\hat{p}}\in A(\hat{p}):W\cap W_{\hat{p}} \neq \emptyset}+(1-\alpha)\gamma'(W, \hat{p}) \sum\limits_{W_p\in A(p)} \omega(W, W \cap W_p, \delta )\\
    \end{align*}
    
    Now $\sum\limits_{W_p\in A(p)} \omega(W, W \cap W_p, \delta ) = \frac{\sum_{i\in I} \delta(i; W)}{d}$ For some set $I$ and $d=\sum_{i\in W}\delta(i; W).$ However, $\sum\limits_{W_{\hat{p}}\in A(\hat{p})} \omega(W, W \cap W_{\hat{p}}, \delta ) = \frac{\sum_{i\in I\setminus i^* \cup i^{**}} \delta(i; W)}{d}$ with $i^{**}< i^*.$ Thus,
    \begin{align*}
    &\alpha\mathds{1}_{\exists W_{\hat{p}}\in A(\hat{p}):W\cap W_{\hat{p}} \neq \emptyset}+(1-\alpha)\gamma'(W, \hat{p}) \sum\limits_{W_p\in A(p)} \omega(W, W \cap W_p, \delta )\\ 
    <& \alpha\mathds{1}_{\exists W_{\hat{p}}\in A(\hat{p}):W\cap W_{\hat{p}} \neq \emptyset}+(1-\alpha)\gamma'(W, \hat{p}) \sum\limits_{W_{\hat{p}}\in A(\hat{p})} \omega(W, W \cap W_{\hat{p}}, \delta )
    \end{align*}
    
    Notice that all other summands remain the same. Thus, finally $Recall_{TS}(g, p) < Recall_{TS}(g, \hat{p})$.
\end{proof}

\begin{theorem}\label{proof:ts-f1}
    \textbf{Range-Based F1-Score} defined as \[
        F1_{TS}(g,p) = 2\frac{Precision_{TS}(g,p) Recall_{TS}(g,p)}{Precision_{TS}(g,p) + Recall_{TS}(g,p)}
    \]
    where
    \[
        Precision_{TS}(g, p) = \frac{1}{|A(p)|}\sum\limits_{W \in A(p)}\gamma_p'(W, g) \sum\limits_{W_g\in A(g)} \omega_p(W, W \cap W_g, \delta_p)
    \]
    where $\delta_p \geq 1$, $0 \leq \omega_p \leq 1$, and \[
        \gamma_p'(W, p) = \gamma_p(W, g)\mathds{1}_{|\{W_p \in A(p) \colon W \cap W_p \neq \emptyset\}| \leq 1}(W) + \mathds{1}_{|\{W_p \in A(p) \colon W \cap W_p \neq \emptyset\}| > 1}(W)
    \]
    where $0 \leq \gamma_p \leq 1$
    and
    \[
        Recall_{TS}(g, p) = \frac{1}{|A(g)|}\sum\limits_{W \in A(g)}(\alpha \mathds{1}(\sum\limits_{W_p \in A(p)} |W \cap W_p| > 0) 
        + (1 - \alpha) \gamma_r'(W, p) \sum\limits_{W_p\in A(p)} \omega_r(W, W \cap W_p, \delta_r )
    \]
    where $0\leq \alpha \leq 1$, $\delta_r \geq 1$, $0 \leq \omega_r \leq 1$, and \[
        \gamma_r'(W, p) = \gamma_r(W, p)\mathds{1}_{|\{W_p \in A(p) \colon W \cap W_p \neq \emptyset\}| \leq 1}(W) + \mathds{1}_{|\{W_p \in A(p) \colon W \cap W_p \neq \emptyset\}| > 1}(W)
    \]
    where $0 \leq \gamma_r \leq 1$
\end{theorem}
\begin{proof}
    \item
    Let $g\colon [n] \rightarrow \{0,1\}^n$ be the ground truth and $p,\hat{p} \colon [n] \rightarrow \{0,1\}^n$ predictions.
    \paragraph{\cref{prop:detection}}
    Let $W \in A(g)$, $\hat{p}_{[n] \setminus W} = p_{[n] \setminus W}$, $p_W = 0$, and $|\hat{p}_W^{-1}(1)| > 0$.
    \begin{align*}
    \mathrm{Precision}_{TS}\bigl(g,\hat{p}\bigr)
    &\geq 
    \mathrm{Precision}_{TS}\bigl(g,p\bigr),
    \\
    \mathrm{Recall}_{TS}\bigl(g,\hat{p}\bigr)
    &> 
    \mathrm{Recall}_{TS}\bigl(g,p\bigr),
    \\
    F1_{TS}\bigl(g,p\bigr)
    &= 
    2\,\frac{
      \mathrm{Precision}_{TS}\bigl(g,p\bigr)
      \times
      \mathrm{Recall}_{TS}\bigl(g,p\bigr)
    }{
      \mathrm{Precision}_{TS}\bigl(g,p\bigr)
      +
      \mathrm{Recall}_{TS}\bigl(g,p\bigr)
    },
    \\
    \text{from the above it follows }\;
    F1_{TS}\bigl(g,\hat{p}\bigr)
    &> 
    F1_{TS}\bigl(g,p\bigr).
    \end{align*}
    
    \paragraph{\cref{prop:alarms}} 
    Consider the following example:
    \begin{align*}
        g &= 111111111111\\
        p &= 100000000000\\
        \hat{p}&= 110011111111
    \end{align*}
    Then it holds $F1_{TS}(g,p) = 0.2667 < 0.5488 = F1_{TS}(g,\hat{p})$.
    
    \paragraph{\cref{prop:false_positives}} 
    Consider the following example:
    \begin{align*}
        g &= 000 001 000 000\\
        p &= 000 010 000 000\\
        \hat{p}&= 000 110 000 000
    \end{align*}
    Then it holds $F1_{TS}(g,p) = 0.0000 = F1_{TS}(g,\hat{p})$.

    \paragraph{\cref{prop:false_positive_alarms}}
     Consider the following example:
    \begin{align*}
        g &= 000 010 000 000\\
        p &= 000 001 000 000\\
        \hat{p}&= 000 101 000 000
    \end{align*}
    Then it holds $F1_{TS}(g,p) = 0.0000 = F1_{TS}(g,\hat{p})$.
    
    \paragraph{\cref{prop:permutations}}
    Consider the following example:
    \begin{align*}
        g &= 000 001 000 000\\
        p &= 000 101 000 000\\
        \hat{p}&= 000 011 000 000
    \end{align*}

     Then it holds $F1_{TS}(g,p) = 0.6667 > 0.5000 = F1_{TS}(g,\hat{p})$.
    
    \paragraph{\cref{prop:trust}}
    Consider the following example:
    \begin{align*}
        g &= 000 010 000 000\\
        p &= 000 001 000 000\\
        \hat{p}&= 000 101 000 000
    \end{align*}
    Then it holds $F1_{TS}(g,p) = 0.000 = 0.0000 = F1_{TS}(g,\hat{p})$.

    \paragraph{\cref{prop:prediction_size}}
    Let $W \in A(g)$, $\hat{p}_{[n] \setminus \{i^*\}} = p_{[n] \setminus \{i^*\}}$ for some $i^* \in W$, $\hat{p}(i^*) \neq p(i^*) = 0$, and $|A(\hat{p}_W)| \leq |A(p_W)|$.
    \begin{align*}
    \mathrm{Precision}_{TS}\bigl(g,\hat{p}\bigr)
    &= 
    \mathrm{Precision}_{TS}\bigl(g,p\bigr),
    \\
    \mathrm{Recall}_{TS}\bigl(g,\hat{p}\bigr)
    &> 
    \mathrm{Recall}_{TS}\bigl(g,p\bigr),
    \\
    F1_{TS}\bigl(g,p\bigr)
    &= 
    2\,\frac{
      \mathrm{Precision}_{TS}\bigl(g,p\bigr)
      \times
      \mathrm{Recall}_{TS}\bigl(g,p\bigr)
    }{
      \mathrm{Precision}_{TS}\bigl(g,p\bigr)
      +
      \mathrm{Recall}_{TS}\bigl(g,p\bigr)
    },
    \\
    \text{from the above it follows }\;
    F1_{TS}\bigl(g,\hat{p}\bigr)
    &> 
    F1_{TS}\bigl(g,p\bigr).
    \end{align*}

    \paragraph{\cref{prop:temporal_order}}
    Consider the following example:
    \begin{align*}
        g &= 111111111111\\
        p &= 100000000001\\
        \hat{p}&= 010100000000
    \end{align*}
    Then it holds $F1_{TS}(g,p) = 0.1538 < 0.2273 = F1_{TS}(g,\hat{p})$.

    \paragraph{\cref{prop:early_bias}}
    Let $W \in A(g)$, $\hat{p}_{[n] \setminus W} = p_{[n] \setminus W}$,
    Consider the following example:
    \begin{align*}
        g &= 000000000011\\
        p &= 111111111110\\
        \hat{p}&= 111111111101
    \end{align*}
    Then it holds $F1_{TS}(g,p) =  0.0296 < 0.4000 = F1_{TS}(g,\hat{p})$.   
   
\end{proof}

\begin{theorem}\label{proof:timesead-precision}
    \textbf{Time-Series Precision} defined as 
    \[
        TPrec(g,p) = \frac{1}{\sum\limits_{W_p \in A(p)}|W_p|} \sum\limits_{W \in A(p)} |W|\gamma(|\{W_g \in A(g) \colon W \cap W_g \neq \emptyset\}|)
        \sum\limits_{W_g\in A(g)} \frac{\sum\limits_{i\in W_g \cap W}\delta(i - \min W, |W|)}{\sum\limits_{i\in W}\delta(i - \min W, |W|)}
    \]
    where $\delta \geq 1$, and $\gamma$ monotonically decreasing with $\gamma (1) = 1$ and 
    \[
        \gamma(1,W) = 1, \hspace{0.5cm}\gamma(n, W) = \max\limits_{0 < m < n} \frac{\sum\limits_{i\in W}\delta(i - \min W, |W|) - n + m}{\sum\limits_{i\in W}\delta(i - \min W, |W|)}\gamma(m, W)
    \]
    \[
    \Longrightarrow
    \delta(\text{pos}, |W|) 
    \;=\; 
    |W|\;-\;\text{pos}\;+\;1
    \quad(\text{front bias for each }i\in W)
    \]
\end{theorem}
\begin{proposition}[Front-Bias $\implies$ Strictly Positive Coverage]
\[
i\in W=[m,m+|W|-1]
\;\Longrightarrow\;
0 \;\le\; i-m \;\le\; |W|-1
\;\Longrightarrow\;
\delta(i;W)\;=\;|W|-\bigl(i-m\bigr)+1\;>\;0.
\]

\[
(W\cap W_p)\;\neq\;\varnothing
\;\Longrightarrow\;
\sum_{i\in(W\cap W_p)}\delta(i;W)\;>\;0,
\quad
\sum_{j\in W}\delta(j;W)\;>\;0
\;\Longrightarrow\;
\frac{\displaystyle\sum_{i\in(W\cap W_p)}\delta(i;W)}
     {\displaystyle\sum_{j\in W}\delta(j;W)}
\;>\;0.
\tag*{\qedsymbol}
\]
\end{proposition}

\begin{proof}
    \item
    Let $g\colon [n] \rightarrow \{0,1\}^n$ be the ground truth and $p,\hat{p} \colon [n] \rightarrow \{0,1\}^n$ predictions.
     \paragraph{\cref{prop:detection}} 
        Consider the following example:
        \begin{align*}
            g &= 000 111 000 111\\
            p &= 000 111 000 000\\
            \hat{p}&= 000 111 000 001
        \end{align*}
        Then it holds $TPrec(g,p) = 1 = TPrec(g,\hat{p})$.

        \paragraph{\cref{prop:alarms}} 
        Consider the following example:
        \begin{align*}
            g &= 000 111 111 000\\
            p &= 000 100 000 000\\
            q &= 000 110 011 000
        \end{align*}
        
        Then it holds
        $
        TPrec(g,p) = 1
        =
        TPrec(g,q).
        $

        \paragraph{\cref{prop:false_positives}} 
        Consider the following example:
        \[
        \begin{aligned}
        g &= 000 000 000 000,\\
        p &= 000 001 100 000,\\
        q &= 000 011 100 000.
        \end{aligned}
        \]
 
        $
        TPrec(g,p) = 0
        =
        TPrec(g,q).
        $
        \paragraph{\cref{prop:false_positive_alarms}}

        Consider the following example:
        \[
        \begin{aligned}
        g &= 000 000 000 001,\\
        p &= 000 100 000 000,\\
        q &= 000 100 100 000.
        \end{aligned}
        \]
        Then it follows
        $
        TPrec(g,p) = 0
        =
        TPrec(g,q).
        $

        \paragraph{\cref{prop:permutations}}
        Consider the following example:
        \[
        \begin{aligned}
        g &= 000 011 110 000,\\
        p &= 110 011 000 000,\\
        q &= 001 111 000 000.
        \end{aligned}
        \]
        Then it holds        
        $
        TPrec(g,p) = 0.5
        > 0.3 =
        TPrec(g,q)
        $
        
        \paragraph{\cref{prop:trust}}
        Consider the following example:
        \[
        \begin{aligned}
        g &= 011 111 111 110,\\
        p &= 000 000 001 011,\\
        q &= 111 111 111 011.
        \end{aligned}
        \]
        Then it follows
        $
        TPrec(g,p) = 0.6667 < 
        0.7222 = TPrec(g,q).
        $

        \paragraph{\cref{prop:prediction_size}}
        Consider the following example:
        \[
        \begin{aligned}
        g &= 000 111 000 000,\\
        p &= 000 110 000 000,\\
        q &= 000 100 000 000.
        \end{aligned}
        \]
        Then it holds
        $
        TPrec(g,p) = 1.0
        =
        TPrec(g,q).
        $
        
        \paragraph{\cref{prop:temporal_order}}
        Consider the following example:
        \[
        \begin{aligned}
        g &= 000 111 100 000,\\
        p &= 000 110 000 000,\\
        q &= 000 011 000 000.
        \end{aligned}
        \]
        Then it holds
        $
        TPrec(g,p) = 1.0
        =
        TPrec(g,q).
        $
        
        \paragraph{\cref{prop:early_bias}}
        Consider the following example:
        \[
        \begin{aligned}
        g &= 000 111 100 000,\\
        p &= 000 100 000 000,\\
        q &= 000 000 100 000.
        \end{aligned}
        \]
        Then it holds
        $
        TPrec(g,p) = 1.0
        =
        TPrec(g,q).
        $
\end{proof}

\begin{theorem}\label{proof:timesead-recall}
    \textbf{Time-Series Recall} defined as \[
        TRec(g, p) = \frac{1}{|A(g)|}\sum\limits_{W \in A(g)}\gamma(|\{W_p \in A(p) \colon W \cap W_p \neq \emptyset\}|)
        \sum\limits_{W_p\in A(p)} \frac{\sum\limits_{i\in W_p \cap W}\delta(i - \min W, |W|)}{\sum\limits_{i\in W}\delta(i - \min W, |W|)}
    \]
    where $\delta \geq 1$, and $\gamma$ monotonically non-increasing with $\gamma (k) = 1/k$ and 
    \[
        \gamma(1,W) = 1, \hspace{0.5cm}\gamma(n, W) = \max\limits_{0 < m < n} \frac{\sum\limits_{i\in W}\delta(i - \min W, |W|) - n + m}{\sum\limits_{i\in W}\delta(i - \min W, |W|)}\gamma(m, W)
    \]

\begin{proposition}[Front-Bias $\implies$ Strictly Positive Coverage]
\[
i\in W=[m,m+|W|-1]
\;\Longrightarrow\;
0 \;\le\; i-m \;\le\; |W|-1
\;\Longrightarrow\;
\delta(i;W)\;=\;|W|-\bigl(i-m\bigr)+1\;>\;0.
\]

\[
(W\cap W_p)\;\neq\;\varnothing
\;\Longrightarrow\;
\sum_{i\in(W\cap W_p)}\delta(i;W)\;>\;0,
\quad
\sum_{j\in W}\delta(j;W)\;>\;0
\;\Longrightarrow\;
\frac{\displaystyle\sum_{i\in(W\cap W_p)}\delta(i;W)}
     {\displaystyle\sum_{j\in W}\delta(j;W)}
\;>\;0.
\tag*{\qedsymbol}
\]
\end{proposition}
\end{theorem}

\begin{proof}
    \item
    Let $g\colon [n] \rightarrow \{0,1\}^n$ be the ground truth and $p,\hat{p} \colon [n] \rightarrow \{0,1\}^n$ predictions.
    \paragraph{\cref{prop:detection}} 
    Let $W \in A(g)$, $\hat{p}_{[n] \setminus W} = p_{[n] \setminus W}$, $p_W = 0$, and $|\hat{p}_W^{-1}(1)| > 0$.
    Then $TRec(g, p)$ and $TRec(g, \hat{p})$ differ in one summand
    \begin{align*}
     0&=\gamma(|\{W_p \in A(p) \colon W \cap W_p \neq \emptyset\}|)
        \sum\limits_{W_p\in A(p)} \frac{\sum\limits_{i\in W_p \cap W}\delta(i - \min W, |W|)}{\sum\limits_{i\in W}\delta(i - \min W, |W|)}\\
     0&<\gamma(|\{W_{\hat{p}} \in A(\hat{p}) \colon W \cap W_{\hat{p}} \neq \emptyset\}|)
        \sum\limits_{W_{\hat{p}}\in A(\hat{p})} \frac{\sum\limits_{i\in W_{\hat{p}} \cap W}\delta(i - \min W, |W|)}{\sum\limits_{i\in W}\delta(i - \min W, |W|)}   
    \end{align*}
    Thus $TRec(g, p) < TRec(g, \hat{p}).$
    
    \paragraph{\cref{prop:alarms}} 
    Consider the following example:
    \begin{align*}
        g &= 111111111111\\
        p &= 100000000000\\
        \hat{p}&= 110011111111
    \end{align*}
    Then it holds $TRec_(g, p) = 0.1667 <  0.8846 = TRec(g, \hat{p})$.
    
    \paragraph{\cref{prop:false_positives}} 
    Consider the following example:
    \begin{align*}
        g &= 000 001 000 000\\
        p &= 000 010 000 000\\
        \hat{p}&= 000 110 000 000
    \end{align*}
    Then it holds $TRec(g, p) = 0 = TRec(g, \hat{p})$.

    \paragraph{\cref{prop:false_positive_alarms}}
    Consider the following example:
    \begin{align*}
        g &= 000 010 000 000\\
        p &= 000 001 000 000\\
        \hat{p}&= 000 101 000 000
    \end{align*}
    Then it holds $TRec(g, p) = 0 = TRec(g, \hat{p})$.

    \paragraph{\cref{prop:permutations}}
    Let $W \in A(|g-1|)$, $\hat{p}_{[n] \setminus W} = p_{[n] \setminus W}$, $|\hat{p}_W^{-1}(1)| = |p_W^{-1}(1)|$, and $|A(\hat{p}_W)| = |A(p_W)|$.
    Let $W \in A(|g-1|)$, $\hat{p}_{[n]\setminus W} = p_{[n]\setminus W}$, 
    $\lvert\hat{p}_W^{-1}(1)\rvert = \lvert p_W^{-1}(1)\rvert,$ 
    and $\lvert A(\hat{p}_W)\rvert = \lvert A(p_W)\rvert.$
    Then, since false positives do not affect the metric at all, $TRec(g, p) = TRec(g, \hat{p})$.
    
     \paragraph{\cref{prop:trust}}
     Consider the following example:
     \begin{align*}
         g &=  000010000000\\
         p &= 000001000000\\
         \hat{p}&= 000101000000
     \end{align*}
     Then it holds $TRec(g, p) = 0 = TRec(g, q)$.
    
    \paragraph{\cref{prop:prediction_size}}
    Let $W \in A(g)$, $\hat{p}_{[n] \setminus \{i^*\}} = p_{[n] \setminus \{i^*\}}$ for some $i^* \in W$, $\hat{p}(i^*) \neq p(i^*) = 0$, and $|A(\hat{p}_W)| \leq |A(p_W)|$. Then
    \begin{align*}
    &\gamma(|\{W_p \in A(p) \colon W \cap W_p \neq \emptyset\}|)
        \sum\limits_{W_p\in A(p)} \frac{\sum\limits_{i\in W_p \cap W}\delta(i - \min W, |W|)}{\sum\limits_{i\in W}\delta(i - \min W, |W|)}\\
    \leq&\gamma(|\{W_{\hat{p}} \in A(\hat{p}) \colon W \cap W_{\hat{p}} \neq \emptyset\}|)
        \sum\limits_{W_p\in A(p)} \frac{\sum\limits_{i\in W_p \cap W}\delta(i - \min W, |W|)}{\sum\limits_{i\in W}\delta(i - \min W, |W|)}\\
    <&\gamma(|\{W_{\hat{p}} \in A(\hat{p}) \colon W \cap W_{\hat{p}} \neq \emptyset\}|)
        \sum\limits_{W_{\hat{p}}\in A(\hat{p})} \frac{\sum\limits_{i\in W_{\hat{p}} \cap W}\delta(i - \min W, |W|)}{\sum\limits_{i\in W}\delta(i - \min W, |W|)}
    \end{align*}
    Thus $TRec(g, p) < TRec(g, \hat{p}).$
    
    \paragraph{\cref{prop:temporal_order}}
    Consider the following example:
    \begin{align*}
        g &= 111111111111\\
        p &= 100000000001 \\
        \hat{p}&= 010100000000 
    \end{align*}
    Then it holds $TRec(g, p) = 0.1923 < 0.2821 = TRec(g, \hat{p})$.
    
    \paragraph{\cref{prop:early_bias}}
    Let $W \in A(g)$, $\hat{p}_{[n] \setminus W} = p_{[n] \setminus W}$,
    Let $p, \hat{p}$ be two predictions with $|A(\hat{p}_W)| \leq |A(p_W)|$, let $i^{*}\in W$ with $p(i^*) = 1$, let $i^{**} < i^* \in W$ with $p(i^{**}) = 0$, and $p_W'(i) = \mathds{1}_{\{i^{**}\}}(i) + p(i) \mathds{1}_{W\setminus \{i^*, i^{**}\}}(i)$. Then
    \begin{align*}
    &\gamma(|\{W_p \in A(p) \colon W \cap W_p \neq \emptyset\}|)
        \sum\limits_{W_p\in A(p)} \frac{\sum\limits_{i\in W_p \cap W}\delta(i - \min W, |W|)}{\sum\limits_{i\in W}\delta(i - \min W, |W|)}\\
    \leq&\gamma(|\{W_{\hat{p}} \in A(\hat{p}) \colon W \cap W_{\hat{p}} \neq \emptyset\}|)
        \sum\limits_{W_p\in A(p)} \frac{\sum\limits_{i\in W_p \cap W}\delta(i - \min W, |W|)}{\sum\limits_{i\in W}\delta(i - \min W, |W|)}
    \end{align*}
    Now
    \begin{align*}
    &\sum\limits_{W_p\in A(p)} \frac{\sum\limits_{i\in W_p \cap W}\delta(i - \min W, |W|)}{\sum\limits_{i\in W}\delta(i - \min W, |W|)}
    =\frac{\sum_{i\in I}\delta(i-minW, |W|)}{d} \\
    <&\frac{\sum_{i\in I\setminus i^* \cup i^{**}}\delta(i-minW, |W|)}{d}  = \sum\limits_{W_{\hat{p}}\in A(\hat{p})} \frac{\sum\limits_{i\in W_{\hat{p}} \cap W}\delta(i - \min W, |W|)}{\sum\limits_{i\in W}\delta(i - \min W, |W|)}
    \end{align*}
    for $d=\sum\limits_{i\in W}\delta(i - \min W, |W|)$ and some $I$. Thus 
    \begin{align*}
    &\gamma(|\{W_{\hat{p}} \in A(\hat{p}) \colon W \cap W_{\hat{p}} \neq \emptyset\}|)
        \sum\limits_{W_p\in A(p)} \frac{\sum\limits_{i\in W_p \cap W}\delta(i - \min W, |W|)}{\sum\limits_{i\in W}\delta(i - \min W, |W|)}\\
    <&\gamma(|\{W_{\hat{p}} \in A(\hat{p}) \colon W \cap W_{\hat{p}} \neq \emptyset\}|)
        \sum\limits_{W_{\hat{p}}\in A(\hat{p})} \frac{\sum\limits_{i\in W_{\hat{p}} \cap W}\delta(i - \min W, |W|)}{\sum\limits_{i\in W}\delta(i - \min W, |W|)}
    \end{align*}
    Thus finally $TRec(g, p) < TRec(g, \hat{p}).$
\end{proof}

\begin{theorem}\label{proof:timesead-f1}
    \textbf{Time-Series F1-Score} defined as 
    \[
        TF1(g,p) = 2\frac{TPrec(g,p)TRec(g,p)}{TPrec(g,p) + TRec(g,p)}
    \]
    where
\end{theorem}
\begin{proof}
    \item
    Let $g\colon [n] \rightarrow \{0,1\}^n$ be the ground truth and $p,\hat{p} \colon [n] \rightarrow \{0,1\}^n$ predictions.
       \paragraph{\cref{prop:detection}}
       Let $W \in A(g)$, $\hat{p}_{[n] \setminus W} = p_{[n] \setminus W}$, $p_W = 0$, and $|\hat{p}_W^{-1}(1)| > 0$.
       \begin{align*}
        TPrec\bigl(g,\hat{p}\bigr)
        &\leq 
        TPrec\bigl(g,p\bigr),
        \\
        TRec\bigl(g,\hat{p}\bigr)
        &> 
        TRec\bigl(g,p\bigr),
        \\
        TF1\bigl(g,p\bigr)
        &= 
        2\,\frac{
          TPrec\bigl(g,p\bigr)
          \times
          TRec\bigl(g,p\bigr)
        }{
          TPrec\bigl(g,p\bigr)
          +
          TRec\bigl(g,p\bigr)
        },
        \\
        \text{from the above it follows }\;
        TF1\bigl(g,\hat{p}\bigr)
        &> 
        TF1\bigl(g,p\bigr).
        \end{align*}    
    
    \paragraph{\cref{prop:alarms}} 
    Consider the following example:
    \begin{align*}
        g &= 111111111111\\
        p &= 100000000000\\
        \hat{p}&= 110011111111
    \end{align*}
    Then it holds $\displaystyle TF1(g,p) = 0.2667 < 0.5488 = TF1(g,\hat{p})$.    
    
    \paragraph{\cref{prop:false_positives}}
    Consider the following example:
    \begin{align*}
        g &= 000 001 000 000\\
        p &= 000 010 000 000\\
        \hat{p}&= 000 110 000 000
    \end{align*}
    Then it holds $\displaystyle TF1(g,p) = 0.0000 = TF1(g,\hat{p})$.

    \paragraph{\cref{prop:false_positive_alarms}}
    Consider the following example:
    \begin{align*}
        g &= 000 010 000 000\\
        p &= 000 001 000 000\\
        \hat{p}&= 000 101 000 000
    \end{align*}
    Then it holds $\displaystyle TF1(g,p) = 0 = TF1(g,\hat{p})$.
    
    \paragraph{\cref{prop:permutations}}
    Consider the following example:
    \begin{align*}
        g &= 000 001 000 000\\
        p &= 000 101 000 000\\
        \hat{p}&= 000 011 000 000
    \end{align*}
    
    Then it holds $\displaystyle TF1(g,p) = 0.6667 > 0.5000 = TF1(g,\hat{p})$.

    \paragraph{\cref{prop:trust}}
    Consider the following example:
    \begin{align*}
        g &= 000 010 000 000\\
        p &= 000 001 000 000\\
        \hat{p}&= 000 101 000 000
    \end{align*}
    Then it holds $\displaystyle TF1(g,p) = 0 = TF1(g,\hat{p})$.
    
    \paragraph{\cref{prop:prediction_size}}
    Let $W \in A(g)$, $\hat{p}_{[n] \setminus \{i^*\}} = p_{[n] \setminus \{i^*\}}$ for some $i^* \in W$, $\hat{p}(i^*) \neq p(i^*) = 0$, and $|A(\hat{p}_W)| \leq |A(p_W)|$.
        \begin{align*}
        TPrec\bigl(g,\hat{p}\bigr)
        &= 
        TPrec\bigl(g,p\bigr),
        \\
        TRec\bigl(g,\hat{p}\bigr)
        &> 
        TRec\bigl(g,p\bigr),
        \\
        TF1\bigl(g,p\bigr)
        &= 
        2\,\frac{
          TPrec\bigl(g,p\bigr)
          \times
          TRec\bigl(g,p\bigr)
        }{
          TPrec\bigl(g,p\bigr)
          +
          TRec\bigl(g,p\bigr)
        },
        \\
        \text{from the above it follows }\;
        TF1\bigl(g,\hat{p}\bigr)
        &> 
        TF1\bigl(g,p\bigr).
    \end{align*}

    \paragraph{\cref{prop:temporal_order}}
    Consider the following example:
    \begin{align*}
        g &= 111111111111\\
        p &= 100000000001\\
        \hat{p}&= 010100000000
    \end{align*}
    Then it holds $TF1(g,p) = 0.1538 < 0.2273 = TF1(g,\hat{p})$.

    \paragraph{\cref{prop:early_bias}}
     \begin{align*}
        g &= 000000000011\\
        p &= 111111111110\\
        \hat{p}&= 111111111101
    \end{align*}
    Then it holds $TF1(g,p) = 0.0296 < 0.4000 = TF1(g,\hat{p})$.
\end{proof}

\begin{theorem}\label{proof:nab}
    \textbf{NAB-Score} defined as \[
        NAB(g,p) = 100 \frac{s_{NAB}(g,p) - s_{NAB}(g,0)}{s_{NAB}(g,g) - s_{NAB}(g,0)}
    \]
    \begin{align*}
        s_{NAB}(g, p) = &A_{FN}|\{W \in A(g)\colon W \cap p^{-1}(1) = \emptyset\}|
        + \sum\limits_{\stackrel{W \in A(g)\colon}{W \cap p^{-1}(1) \neq \emptyset}} A_{TP}\left( \frac{2}{1 + e^{5 (\min W\cap p^{-1}(1) - \max W)}} - 1 \right)\\
        + &\sum\limits_{\stackrel{i \in p^{-1}(1) \cap g^{-1}(0) \colon}{i > \min g^{-1}(1)} } A_{FP}\left( \frac{2}{1 + e^{5 (\min\limits_{\stackrel{W \in A(g)\colon}{\max W < i}} (i - \max W))}} - 1 \right)
        -  A_{FP}|\{i \in p^{-1}(1) \cap g^{-1}(0) \colon i < \min g^{-1}(1)\}|
    \end{align*}
    where $A_{TP} \geq 0$, $A_{FP} \geq 0$, $A_{FN} \leq 0$
    satisfies 
\end{theorem}
\begin{proof}
    \item
    For all properties it suffices to show the relation of $s_{NAB}$, as the normalization does not change the order.
    Let $g\colon [n] \rightarrow \{0,1\}^n$ be the ground truth and $p,\hat{p} \colon [n] \rightarrow \{0,1\}^n$ predictions.
    \paragraph{\cref{prop:detection}} 
    Let $W \in A(g)$, $\hat{p}_{[n] \setminus W} = p_{[n] \setminus W}$, $p_W = 0$, and $|\hat{p}_W^{-1}(1)| > 0$.
    Then $s_{NAB}(g, p)$ and $s_{NAB}(g, \hat{p})$ only differ in the terms
    \begin{align*}
        \sum\limits_{\stackrel{W_1 \in A(g)\colon}{W_1 \cap \hat{p}^{-1}(1) \neq \emptyset}} A_{TP}\left( \frac{2}{1 + e^{5 (\min W_1\cap \hat{p}^{-1}(1) - \max W_1)}} - 1 \right)
        = &\sum\limits_{\stackrel{W_1 \in A(g)\colon}{W_1 \cap p^{-1}(1) \neq \emptyset}} A_{TP}\left( \frac{2}{1 + e^{5 (\min W_1\cap p^{-1}(1) - \max W_1)}} - 1 \right)\\
        &+ \underbrace{A_{TP}\left( \frac{2}{1 + e^{5 (\min W\cap \hat{p}^{-1}(1) - \max W)}} - 1 \right)}_{> 0}.
    \end{align*}
    and
    \[
        A_{FN}|\{W \in A(g)\colon W \cap \hat{p}^{-1}(1) = \emptyset\}|
        = A_{FN}|\{W \in A(g)\colon W \cap p^{-1}(1) = \emptyset\}| - A_{FN}
    \]
    Thus, it follows
    \[
        s_{NAB}(g, \hat{p}) - s_{NAB}(g, p) = A_{TP}\left( \frac{2}{1 + e^{5 (\min W\cap \hat{p}^{-1}(1) - \max W)}} - 1 \right) - A_{FN} \geq 0
    \]
    Where equality is achieved only, if $\hat{p}^{-1}(1) = \{\max W\}$ or $A_{TP} = 0$, and $A_{FN} = 0$.
    Thus, if $A_{TP} > 0$ and $A_{FN} < 0$, it holds $s_{NAB}(g, p) < s_{NAB}(g, \hat{p})$.
    
    \paragraph{\cref{prop:alarms}} 
    Consider the following example:
    \begin{align*}
        g &= 000 111 111 000\\
        p &= 000 111 000 000\\
        \hat{p}&= 000 111 011 000
    \end{align*}
    Then it holds $s_{NAB}(g,p) = A_{TP}\left( \frac{2}{1 + e^{-30}} - 1 \right) = s_{NAB}(g,\hat{p})$.
    
    \paragraph{\cref{prop:false_positives}} 
    Let $W \in A(|g-1|)$, $\hat{p}_{[n] \setminus \{i^*\}} = p_{[n] \setminus \{i^*\}}$ for some $i^* \in W$, $\hat{p}(i^*) \neq p(i^*) = 0$, and $|A(\hat{p}_W)| = |A(p_W)|$.
    \\
    \textbf{Case 1: $i^* < \min g^{-1}(1)$} 
    Then $s_{NAB}(g, p)$ and $s_{NAB}(g, \hat{p})$ only differ in the term
    \[
        |\{i \in \hat{p}^{-1}(1) \cap g^{-1}(0) \colon i < \min g^{-1}(1)\}|
        = |\{i \in p^{-1}(1) \cap g^{-1}(0) \colon i < \min g^{-1}(1)\}| + 1
    \]
    Thus, it follows $s_{NAB}(g, p) > s_{NAB}(g, \hat{p})$, if $A_{FP} > 0$.
    \\
    \textbf{Case 2: $i^* > \min g^{-1}(1)$} 
    Then $s_{NAB}(g, p)$ and $s_{NAB}(g, \hat{p})$ only differ in the third term by
    \[
        \underbrace{\left( \frac{2}{1 + e^{5 (\min\limits_{\stackrel{W \in A(g)\colon}{\max W < i^*}} (i^* - \max W))}} - 1 \right)}_{< 0}.
    \]
    Thus, it follows $s_{NAB}(g, p) > s_{NAB}(g, \hat{p})$, if $A_{FP} > 0$.
    
    \paragraph{\cref{prop:false_positive_alarms}}
    Consider the following example:
    \begin{align*}
        g &= 000 110 000 000\\
        p &= 000 000 000 011\\
        \hat{p}&= 000 001 000 010
    \end{align*}
    Then it holds $s_{NAB}(g,p) = A_{FP}\left( \frac{2}{1 + e^{30}} - 1  + \frac{2}{1 + e^{35}} - 1\right) \leq A_{FP}\left( \frac{2}{1 + e^{30}} - 1  + \frac{2}{1 + e^{5}} - 1\right) = s_{NAB}(g,\hat{p})$.
    
    \paragraph{\cref{prop:permutations}}
    Consider the following example:
    \begin{align*}
        g &= 000 110 000 000\\
        p &= 000 000 000 001\\
        \hat{p}&= 000 001 000 000
    \end{align*}
    Then it holds $s_{NAB}(g,p) = A_{FP}\left(\frac{2}{1 + e^{35}} - 1\right) < A_{FP}\left(\frac{2}{1 + e^{5}} - 1\right) = s_{NAB}(g,\hat{p})$, if $A_{FP} > 0$.
    
    \paragraph{\cref{prop:trust}}
    Let $W \in A(g)$, $\hat{W} \in A(|g-1|)$, $|A(\hat{p}_{W})| = |A(p_{W})|$, $\hat{p}_{[n] \setminus (W \cup \hat{W})} = p_{[n] \setminus (W \cup \hat{W})}$, $p_{\hat{W}} = 0$, and $\hat{p}_{\hat{W}} = \mathds{1}_{\{i^*\}}$ for some $i^* \in \hat{W}$.
    Then $s_{NAB}(g,p)$ and $s_{NAB}(g, \hat{p})$ only differ in the terms
    \[
        0 \leq \underbrace{\left( \frac{2}{1 + e^{5 (\min W\cap p^{-1}(1) - \max W)}} - 1 \right)}_{x_p} \leq 1
    \]
    and 
    \[
         -1\leq \underbrace{\left( \frac{2}{1 + e^{5 (\min\limits_{\stackrel{W_1 \in A(g)\colon}{\max W_1 < i^*}} (i^* - \max W_1))}} - 1 \right) - \mathds{1}(i^* < \min g^{-1}(1))}_{y_p} 
         \leq \left( \frac{2}{1 + e^{5}} - 1 \right) < 0
    \]
    For the property to be satisfied, we need 
    \begin{align*}
        A_{TP}x_{\hat{p}} + A_{FP}y_{\hat{p}} < A_{TP}x_{p}\\
        A_{TP}(x_{\hat{p}} - x_{p}) < A_{FP}y_{\hat{p}}
    \end{align*}
    Since, $A_{TP}x_{p} \geq 0$ and $A_{FP}y_{\hat{p}} \leq 0$, this does not holds if $x_{\hat{p}} - x_{p} > 0$, i.e. $\min \hat{p}_W^{-1}(1) < \min p_W^{-1}(1)$.
    
    \paragraph{\cref{prop:prediction_size}}
    Consider the following example:
    \begin{align*}
        g &= 000 111 111 000\\
        p &= 000 100 000 000\\
        \hat{p}&= 000 111 111 000
    \end{align*}
    Then it holds $s_{NAB}(g,p) = A_{TP}\left( \frac{2}{1 + e^{-30}} - 1 \right) = s_{NAB}(g,\hat{p})$.
    
    \paragraph{\cref{prop:temporal_order}}
    Let $W \in A(g)$, $\hat{p}_{[n] \setminus W} = p_{[n] \setminus W}$, $|A(\hat{p}_W)| = |A(p_W)|$, $|\hat{p}^{-1}(1)| = |p^{-1}(1)|$, and $\min\limits_{i \in W\colon p(i) = 1} i < \min\limits_{i \in W\colon \hat{p}(i) = 1} i$. 
    Then it holds
    \begin{align*}
        s_{NAB}(g,p) - s_{NAB}(g,\hat{p}) &= A_{TP}\left( \frac{2}{1 + e^{5 (\min W\cap p^{-1}(1) - \max W)}} - 1 \right) - A_{TP}\left( \frac{2}{1 + e^{5 (\min W\cap \hat{p}^{-1}(1) - \max W)}} - 1 \right)\\
        &= A_{TP}\left( \frac{2}{1 + e^{5 (\min W\cap p^{-1}(1) - \max W)}} - \frac{2}{1 + e^{5 (\min W\cap \hat{p}^{-1}(1) - \max W)}} \right) > 0.
    \end{align*}
    Thus, it holds $s_{NAB}(g,p) > s_{NAB}(g,\hat{p})$.
    
    \paragraph{\cref{prop:early_bias}}
    Consider the following example:
    \begin{align*}
        g &= 000 111 111 000\\
        p &= 000 100 001 000\\
        \hat{p}&= 000 110 000 000
    \end{align*}
    Then it holds $s_{NAB}(g,p) = A_{TP}\left( \frac{2}{1 + e^{-30}} - 1 \right) = s_{NAB}(g,\hat{p})$.

\end{proof}

\begin{theorem}\label{proof:tap}
    \textbf{Time-Series Aware Precision} defined as
    \[
        TaP(g,p) = 
        \alpha \frac{|\{W \in A(p) \colon \frac{1}{|W|}\sum\limits_{W_g \in A(g)} O(W_g, W) \geq \theta\}|}{|A(p)|} 
        + 
        (1-\alpha) \frac{1}{|A(p)|}\sum\limits_{W \in A(p)} \frac{\sum\limits_{W_g \in A(g)} O(W_g, W)}{|W|}
    \]
    where $\alpha \in [0,1]$, $\delta \geq 0$, $\theta > 0$, and \[
        O(W, W_p) = |W \cap W_p| + \sum_{j \in [\max W + 1, \max W + \delta] \cap W_p} \frac{1}{1 + e^{\frac{12(j - 2 - \max W)}{\delta - 1} -6}}
    \]
    satisfies no properties.
\end{theorem}
\begin{proof}
    \item
    Let g$\colon [n] \rightarrow \{0,1\}^n$ be the ground truth and $p,\hat{p} \colon [n] \rightarrow \{0,1\}^n$ predictions.
    \paragraph{\cref{prop:detection}} 
    Consider the example $g = \mathds{1}_{[b, b+1]}$, $p = \mathds{1}_{[a, b-1]} + \mathds{1}_{\{b + 2\}}$, and $\hat{p} = \mathds{1}_{[a, b+2]}$ with $a < b$.
    Then, the second term for $p$ is
    \[
        (1-\alpha) \frac{1}{|A(p)|}\sum\limits_{W \in A(p)} \frac{\sum\limits_{W_g \in A(g)} O(W_g, W)}{|W|} = (1 - \alpha)\frac{1}{2} \frac{1}{1 + e^{\frac{-12}{\delta - 1} -6}}
    \]
    and for $\hat{p}$
    \[
        (1-\alpha) \frac{1}{|A(\hat{p})|}\sum\limits_{W \in A(\hat{p})} \frac{\sum\limits_{W_g \in A(g)} O(W_g, W)}{|W|} = (1-\alpha) \frac{1}{|b + 2 - a|}\left( 1 + \frac{1}{1 + e^{\frac{-12}{\delta - 1} -6}}\right)
    \]
    For all $\delta$, we can select $a$ small enough and $b$ larger enough, such that 
    \[
        \frac{1}{2}\frac{1}{1 + e^{\frac{-12}{\delta - 1} -6}} > \frac{1}{|b + 2 - a|}\left( 1 + \frac{1}{1 + e^{\frac{-12}{\delta - 1} -6}}\right)
    \]
    which implies the second term of $TaP$ is smaller for $\hat{p}$ than for $p$.
    Additionally,
    \[
        \frac{1}{1 + e^{\frac{-12}{\delta - 1} -6}} > \frac{1}{|b + 2 - a|}\left( 1 + \frac{1}{1 + e^{\frac{-12}{\delta - 1} -6}}\right)
    \]
    implies that the first term of $TaP$ for $\hat{p}$ can never be larger than for $p$.
    Thus, it holds $TaP(g,p) > TaP(g,\hat{p})$.
    
    \paragraph{\cref{prop:alarms}} 
    Consider the following example:
    \begin{align*}
        g &= 000 111 111 000\\
        p &= 000 110 000 000\\
        \hat{p}&= 000 110 011 000
    \end{align*}
    The first term is equal to $\alpha$ for predictions, $1 \geq \delta$, end $0$ otherwise. 
    The second term is equal to $(1-\alpha)$ for both predictions.
    Thus, it holds $TaP(g,p) = TaP(g,\hat{p})$.
    
    \paragraph{\cref{prop:false_positives}} 
    Consider the following example:
    \begin{align*}
        g &= 000 000 111 000\\
        p &= 000 100 000 000\\
        \hat{p}&= 000 111 000 000
    \end{align*}
    Then it holds $TaP(g,p) = 0 = TaP(g,\hat{p})$.
    
    \paragraph{\cref{prop:false_positive_alarms}}
    Consider the following example:
    \begin{align*}
        g &= 000 000 111 000\\
        p &= 000 100 000 000\\
        \hat{p}&= 000 101 000 000
    \end{align*}
    Then it holds $TaP(g,p) = 0 = TaP(g,\hat{p})$.
    
    \paragraph{\cref{prop:permutations}}
    Consider the example $g = \mathds{1}_{\{a\}}$, $p = \mathds{1}_{\{a+1\}}$, and $\hat{p} = \mathds{1}_{\{a + \delta + 1\}}$.
    Then it holds $TaP(g,p) > \alpha \mathds{1}(\theta = 0) = TaP(g,\hat{p})$.
    
    \paragraph{\cref{prop:trust}}
    Consider the following example:
    \begin{align*}
        g &= 000 000 111 000\\
        p &= 000 000 000 100\\
        \hat{p}&= 000 100 000 100
    \end{align*}
    Then it holds $TaP(g,p) = 0 = TaP(g,\hat{p})$.
    
    \paragraph{\cref{prop:prediction_size}}
    Consider the following example:
    \begin{align*}
        g &= 000 111 111 000\\
        p &= 000 110 000 000\\
        \hat{p}&= 000 111 111 000
    \end{align*}
    Then it holds $TaP(g,p) = \alpha \mathds{1}(1 \geq \theta) + (1 - \alpha) = TaP(g,\hat{p})$.
    
    \paragraph{\cref{prop:temporal_order}}
    Consider the following example:
    \begin{align*}
        g &= 000 111 111 000\\
        p &= 000 111 000 000\\
        \hat{p}&= 000 000 111 000
    \end{align*}
    Then it holds $TaP(g,p) = \alpha \mathds{1}(1 \geq \theta) + (1 - \alpha) = TaP(g,\hat{p})$.
    
    \paragraph{\cref{prop:early_bias}}
    Consider the following example:
    \begin{align*}
        g &= 000 111 111 000\\
        p &= 000 000 111 000\\
        \hat{p}&= 000 001 110 000
    \end{align*}
    Then it holds $TaP(g,p) = \alpha \mathds{1}(1 \geq \theta) + (1 - \alpha) = TaP(g,\hat{p})$.
\end{proof}

\begin{theorem}\label{proof:tar}
    \textbf{Time-Series Aware Recall} defined as
    \[
        TaR(g,p) = \alpha \frac{|\{W \in A(g) \colon \frac{1}{|W|}\sum\limits_{W_p \in A(p)} O(W, W_p) \geq \theta\}|}{|A(g)|} + (1-\alpha) \frac{1}{|A(g)|}\sum\limits_{W \in A(g)} \min \left\{ 1, \frac{\sum\limits_{W_p \in A(p)} O(W, W_p)}{|W|} \right\}
    \]
    where $\alpha \in [0,1]$, $\delta \geq 0$, $\theta > 0$, and \[
        O(W, W_p) = |W \cap W_p| + \sum_{j \in [\max W + 1, \max W + \delta] \cap W_p} \frac{1}{1 + e^{-6 \frac{12(j - 2 - \max W)}{\delta - 1}}}
    \]
    satisfies \cref{prop:detection} and \cref{prop:prediction_size} for many but not all parameter combinations.
\end{theorem}
\begin{proof}
    \item
    Let g$\colon [n] \rightarrow \{0,1\}^n$ be the ground truth and $p,\hat{p} \colon [n] \rightarrow \{0,1\}^n$ predictions.

    \paragraph{\cref{prop:detection}} 
    Let $W \in A(g)$, $\hat{p}_{[n] \setminus W} = p_{[n] \setminus W}$, $p_W = 0$, and $|\hat{p}_W^{-1}(1)| > 0$.

    In general, it holds
    \[
        |\{W \in A(g) \colon \frac{1}{|W|}\sum\limits_{W_p \in A(\hat{p})} O(W, W_p) \geq \theta\}| \geq |\{W \in A(g) \colon \frac{1}{|W|}\sum\limits_{W_p \in A(p)} O(W, W_p) \geq \theta\}|
    \]
    and
    \[
        \sum\limits_{W \in A(g)} \min \left\{ 1, \frac{\sum\limits_{W_p \in A(\hat{p})} O(W, W_p)}{|W|} \right\} \geq \sum\limits_{W \in A(g)} \min \left\{ 1, \frac{\sum\limits_{W_p \in A(p)} O(W, W_p)}{|W|} \right\},
    \]
    If the first term depending on $\theta$ does not change, we can achieve equality in the second term for some parameterizations, e.g. $\delta = 1$.
    
    \paragraph{\cref{prop:alarms}} 
    Let $W \in A(g)$, and let $p_W(i) = 0$ for all $i \in I \subset W$, $|A(\hat{p}_W)| = |A(p_W)| + 1$, $\min\limits_{i \in I} i > \max\limits_{j \in w\colon p(j) = 1} j$, and $\hat{p}(i) = \mathds{1}_{I}(i) + p(i)\mathds{1}_{[n] \setminus I}(i)$.
    
    \paragraph{\cref{prop:false_positives}} 
    Consider the following example:
    \begin{align*}
        g &= 000 000 111 000\\
        p &= 000 100 000 000\\
        \hat{p}&= 000 111 000 000
    \end{align*}
    Then it holds $TaR(g,p) = 0 = TaR(g,\hat{p})$.
    
    \paragraph{\cref{prop:false_positive_alarms}}
    Consider the following example:
    \begin{align*}
        g &= 000 000 111 000\\
        p &= 000 100 000 000\\
        \hat{p}&= 000 101 000 000
    \end{align*}
    Then it holds $TaR(g,p) = 0 = TaR(g,\hat{p})$.
    
    \paragraph{\cref{prop:permutations}}
    Consider the example $g = \mathds{1}_{\{a\}}$, $p = \mathds{1}_{\{a+1\}}$, and $\hat{p} = \mathds{1}_{\{a + \delta + 1\}}$.
    Then it holds $TaR(g,p) > \alpha \mathds{1}(\theta = 0) = TaR(g,\hat{p})$.

    \paragraph{\cref{prop:trust}}
    Consider the following example:
    \begin{align*}
        g &= 000 000 111 000\\
        p &= 000 000 000 000\\
        \hat{p}&= 000 100 000 000
    \end{align*}
    Then it holds $TaR(g,p) = 0 = TaR(g,\hat{p})$.
    
    \paragraph{\cref{prop:prediction_size}}
    Let $W \in A(g)$, $\hat{p}_{[n] \setminus \{i^*\}} = p_{[n] \setminus \{i^*\}}$ for some $i^* \in W$, $\hat{p}(i^*) \neq p(i^*) = 0$, and $|A(\hat{p}_W)| \leq |A(p_W)|$.

    In general, it holds
    \[
        |\{W \in A(g) \colon \frac{1}{|W|}\sum\limits_{W_p \in A(\hat{p})} O(W, W_p) \geq \theta\}| \geq |\{W \in A(g) \colon \frac{1}{|W|}\sum\limits_{W_p \in A(p)} O(W, W_p) \geq \theta\}|
    \]
    and
    \[
        \sum\limits_{W \in A(g)} \min \left\{ 1, \frac{\sum\limits_{W_p \in A(\hat{p})} O(W, W_p)}{|W|} \right\} \geq \sum\limits_{W \in A(g)} \min \left\{ 1, \frac{\sum\limits_{W_p \in A(p)} O(W, W_p)}{|W|} \right\},
    \]
    where equality holds only, if 
    \[
        \sum_{j \in [\max W + 1, \max W + \delta] \cap p^{-1}(1)} \frac{1}{1 + e^{-6 \frac{12(j - 2 - \max W)}{\delta - 1}}} \geq |(g \cdot p)_W^{-1}(0)|
    \]
    There are parameterizations for which this is true, e.g. $\delta = 1$, but also parameterizations for which it is not true, e.g. $\delta = 0$.
    
    \paragraph{\cref{prop:temporal_order}}
    Consider the following example:
    \begin{align*}
        g &= 000 111 111 000\\
        p &= 000 011 000 000\\
        \hat{p}&= 000 110 000 000
    \end{align*}
    Then it holds $TaR(g,p) = TaR(g,\hat{p})$.

    \paragraph{\cref{prop:early_bias}}
    Consider the following example:
    \begin{align*}
        g &= 000 111 111 000\\
        p &= 000 010 100 000\\
        \hat{p}&= 000 110 000 000
    \end{align*}
    Then it holds $TaR(g,p) = TaR(g,\hat{p})$.

\end{proof}

\begin{theorem}\label{proof:p_delta}
    \textbf{Time-Tolerant Precision} \[
        P_{\delta}(g,p) = \frac{|\{i \in p^{-1}(1) \colon [i - \delta, i +\delta] \cap g^{-1}(1) \neq \emptyset\}|}{|p^{-1}(1)|}
    \] where $\delta \in \mathbb{Z}_{\geq 0}$ satisfies property \cref{prop:permutations} for $\delta=0$.
\end{theorem}
\begin{proof}
    \item
    Let g$\colon [n] \rightarrow \{0,1\}^n$ be the ground truth and $p,\hat{p} \colon [n] \rightarrow \{0,1\}^n$ predictions.
    
    \paragraph{\cref{prop:detection}} 
    % Let $W \in A(g)$, $\hat{p}_{[n] \setminus W} = p_{[n] \setminus W}$, $p_W = 0$, and $|\hat{p}_W^{-1}(1)| > 0$.
    Consider the following example:
        \begin{align*}
            g &= 000 110 011 000 \\
            p &= 000 110 000 000 \\
            \hat{p}&= 000 110 001 000 
        \end{align*}
    Then $\forall \delta: P_{\delta}(g,p) = 1 = P_{\delta}(g,\hat{p})$.

    \paragraph{\cref{prop:alarms}} 
    Consider the following example:
    \begin{align*}
        g &= 000 111 111 000\\
        p &= 000 111 000 000\\
        \hat{p}&= 000 111 011 000
    \end{align*}
    Then $\forall \delta: P_{\delta}(g,p) = 1 = P_{\delta}(g,\hat{p})$.
    
    \paragraph{\cref{prop:false_positives}} 
    Consider the following example:
        \begin{align*}
            g &= 000 000 000 000\\
            p &= 100 000 000 000\\
            \hat{p}&= 110 000 000 000
        \end{align*}
    Then it holds $P_{\delta}(g,p) = 0 = P_{\delta}(g,\hat{p}).$
    
    \paragraph{\cref{prop:false_positive_alarms}}
    Consider the following example:
    \begin{align*}
        g &= 000 000 000 000 \\
        p &= 000 110 000 000 \\
        \hat{p}&= 000 101 000 000
    \end{align*}
    Then it holds $P_{\delta}(g,p) = 0 = P_{\delta}(g,\hat{p}).$
    
    \paragraph{\cref{prop:permutations}}
    \textbf{Case $\delta=0$.} 
        Then $P_{\delta}(g,p)=Precision(g,p)$ and this property is fulfilled following \cref{proof:point-wise_precision}.
    
    \textbf{Case $\delta>0$.}
        Consider the following example: $g=\mathds{1}_{[a, a+1]}$ with $\delta+1<a<n$, $p=\mathds{1}_{\{a-\delta,a+1\}}$, and $\hat{p}=\mathds{1}_{\{a-\delta-1,a+1\}}$. Then it holds $P_{\delta}(g,p) = 1 > \frac{1}{2}= P_{\delta}(g,\hat{p})$.
    
    For example, let $\delta=1$:
    \begin{align*}
        g &= 000 000 110 000 \\
        p &= 000 001 010 000 \\
   \hat{p}&= 000 010 010 000
    \end{align*}
    Then it holds $P_{\delta}(g,p) = 1 > \frac{1}{2}= P_{\delta}(g,\hat{p})$.
    
    \paragraph{\cref{prop:trust}}
    Consider the following example:
    \begin{align*}
        g &= 000 111 111 000 \\
        p &= 000 010 000 100 \\
        \hat{p}&= 001 011 110 100 
    \end{align*}
    \begin{itemize}
        \item For $\delta=0$: $P_{\delta}(g,p) = \frac{1}{2}< \frac{2}{3} = P_{\delta}(g,\hat{p})$
        \item For $\delta > 0$: $P_{\delta}(g,p) = 1 = P_{\delta}(g,\hat{p})$
    \end{itemize}
    Regardless of the choice of $\delta$ this property does not hold.
    
    \paragraph{\cref{prop:prediction_size}}
    Consider the following example:
    \begin{align*}
        g &= 000 111 111 000 \\
        p &= 000 010 000 000 \\
        \hat{p}&= 000 111 111 000 
    \end{align*}
    Then $\forall \delta: P_{\delta}(g,p) = 1 = P_{\delta}(g,\hat{p})$.
    
    \paragraph{\cref{prop:temporal_order}}
    Consider the following example:
    \begin{align*}
        g &= 000 111 111 000 \\
        p &= 000 010 000 000 \\
        \hat{p}&= 000 000 010 000 
    \end{align*}
    Then $\forall \delta: P_{\delta}(g,p) = 1 = P_{\delta}(g,\hat{p})$.
    
    \paragraph{\cref{prop:early_bias}}
    Consider the following example:
    \begin{align*}
        g &= 000 111 111 000 \\
        p &= 000 010 010 000 \\
        \hat{p}&= 000 110 000 000 
    \end{align*}
    Then $\forall \delta: P_{\delta}(g,p) = 1 = P_{\delta}(g,\hat{p})$.
\end{proof}
 
\begin{theorem}\label{proof:r_delta}
    \textbf{Time-Tolerant Recall} \[
        R_{\delta}(g,p) = \frac{|\{i \in g^{-1}(1) \colon [i - \delta, i +\delta] \cap p^{-1}(1) \neq \emptyset\}|}{|g^{-1}(1)|}
    \] where $\delta \in \mathbb{Z}_{\geq 0}$ satisfies properties \cref{prop:detection}, \cref{prop:permutations}, and \cref{prop:prediction_size} only for $\delta=0$.
\end{theorem}
\begin{proof}
    \item
    Let g$\colon [n] \rightarrow \{0,1\}^n$ be the ground truth and $p,\hat{p} \colon [n] \rightarrow \{0,1\}^n$ predictions.
    
    \paragraph{\cref{prop:detection}} 
    \textbf{Case $\delta=0$.} Then $R_{\delta}(g,p)=Recall(g,p)$ and this property is fulfilled following \cref{proof:point-wise_recall}.
    
    \textbf{Case $\delta>0$.} 
    Consider the following example: 
    \begin{align*}
        g &= 000 000 100 000 \\
        p &= 000 001 100 000 \\
        \hat{p}&= 000 001 000 000
    \end{align*}
    Then it holds $R_{\delta}(g,p) = 1 = R_{\delta}(g,\hat{p})$.
    
    \paragraph{\cref{prop:alarms}} 
    Consider the following example:
    \begin{align*}
        g &= 000 111 111 000\\
        p &= 000 110 000 000\\
        \hat{p}&= 000 110 011 000
    \end{align*}
    Then $\forall \delta: R_{\delta}(g,p) = \frac{1}{3} < \frac{2}{3} = R_{\delta}(g,\hat{p})$.
    
    \paragraph{\cref{prop:false_positives}} 
    Consider the following example:
        \begin{align*}
            g &= 000 000 111 000\\
            p &= 000 010 000 000\\
            \hat{p}&= 000 011 000 000
        \end{align*}
    Then it holds
    \begin{itemize}
        \item For $\delta=0$: $R_{\delta}(g,p) = 0 = R_{\delta}(g,\hat{p})$
        \item For $\delta=1$: $R_{\delta}(g,p) = 0 < \frac{1}{3} = R_{\delta}(g,\hat{p})$
        \item For $\delta>1$: $R_{\delta}(g,p) = \frac{1}{3} < \frac{2}{3} = R_{\delta}(g,\hat{p})$
    \end{itemize}
    Regardless of the choice of $\delta$ this property does not hold.

    \paragraph{\cref{prop:false_positive_alarms}}
    Consider the following example:
    \begin{align*}
        g &= 000 000 111 000 \\
        p &= 000 110 010 000 \\
        \hat{p}&= 000 101 010 000
    \end{align*}
    Then it holds
    \begin{itemize}
        \item For $\delta=0$: $R_{\delta}(g,p) = \frac{1}{3} = R_{\delta}(g,\hat{p})$
        \item For $\delta>0$: $R_{\delta}(g,p) = 1 = R_{\delta}(g,\hat{p})$
    \end{itemize}
    Regardless of the choice of $\delta$ this property does not hold.
    
    \paragraph{\cref{prop:permutations}}
    \textbf{Case $\delta=0$.} Then $R_{\delta}(g,p)=Recall(g,p)$ and this property is fulfilled following \cref{proof:point-wise_recall}.
    
    \textbf{Case $\delta>0$.} 
        Consider the following example: $g=\mathds{1}_{[a, a+\delta+1]}$ with $\delta+1<a<a+\delta<n$, $p=\mathds{1}_{\{a-\delta,a+\delta+1\}}$, and $\hat{p}=\mathds{1}_{\{a-\delta-1,a+\delta+1\}}$. Then it holds $R_{\delta}(g,p) = 1 > \frac{\delta+1}{\delta+2} = R_{\delta}(g,\hat{p})$.
    
    For example, let $\delta=1$:
    \begin{align*}
        g &= 000 000 111 000 \\
        p &= 000 001 001 000 \\
   \hat{p}&= 000 010 001 000
    \end{align*}
    Then it holds $R_{\delta}(g,p) = 1 > \frac{\delta+1}{\delta+2} = \frac{2}{3} = R_{\delta}(g,\hat{p})$.

    \paragraph{\cref{prop:trust}}
    Consider the following example:
    \begin{align*}
        g &= 000 000 111 000 \\
        p &= 000 000 010 000 \\
   \hat{p}&= 000 100 110 000
    \end{align*}
    Then it holds
    \begin{itemize}
        \item For $\delta=0$: $R_{\delta}(g,p) = \frac{1}{3} < \frac{2}{3} = R_{\delta}(g,\hat{p})$
        \item For $\delta>0$: $R_{\delta}(g,p) = 1 = R_{\delta}(g,\hat{p})$
    \end{itemize}
    Regardless of the choice of $\delta$ this property does not hold.

    \paragraph{\cref{prop:prediction_size}}
    \textbf{Case $\delta=0$.} Then $R_{\delta}(g,p)=Recall(g,p)$ and this property is fulfilled following \cref{proof:point-wise_recall}.

    \textbf{Case $\delta>0$.}
    Consider the following example: $g=\mathds{1}_{[0, \delta+1]}$, $p=\mathds{1}_{\{0\}}+\mathds{1}_{\{\delta+2\}}$, and $p=\mathds{1}_{[0,1]}+\mathds{1}_{\{\delta+2\}}$. Then it holds $R_{\delta}(g,p) = 1 = R_{\delta}(g,\hat{p})$.
    
    For example for $\delta = 1$:
    \begin{align*}
        g &= 111 000 000 000 \\
        p &= 100 100 000 000 \\
   \hat{p}&= 110 100 000 000
    \end{align*}
    Then it follows $R_{\delta}(g,p) = 1 = R_{\delta}(g,\hat{p})$
    
    \paragraph{\cref{prop:temporal_order}}
    Consider the following example:
    \begin{align*}
        g &= 000 001 100 000 \\
        p &= 000 001 000 000 \\
   \hat{p}&= 000 000 100 000
    \end{align*}
    Then it holds
    \begin{itemize}
        \item For $\delta=0$: $R_{\delta}(g,p) = \frac{1}{2} = R_{\delta}(g,\hat{p})$
        \item For $\delta>0$: $R_{\delta}(g,p) = 1 = R_{\delta}(g,\hat{p})$
    \end{itemize}
    Regardless of the choice of $\delta$ this property does not hold.

    \paragraph{\cref{prop:early_bias}}
    Consider the following example:
    \begin{align*}
        g &= 0000 1111 0000\\
        p &= 0000 1100 0000\\
   \hat{p}&= 0000 0101 0000
    \end{align*}
    Then it holds
    \begin{itemize}
        \item For $\delta=0$: $R_{\delta}(g,p) = \frac{1}{2} = R_{\delta}(g,\hat{p})$
        \item For $\delta=1$: $R_{\delta}(g,p) = \frac{3}{4} < 1 = R_{\delta}(g,\hat{p})$
        \item For $\delta>1$: $R_{\delta}(g,p) = 1 = R_{\delta}(g,\hat{p})$
    \end{itemize}
    Regardless of the choice of $\delta$ this property does not hold.
\end{proof}
 
\begin{theorem}\label{proof:p_precision}
    \textbf{Average Precision Probability} defined as 
    \begin{align*}
        &P_{Precision}(g,p) \\
        &= \frac{
            \sum\limits_{\stackrel{W \in A(g)\colon}{I_W \cap p^{-1}(1) \neq \emptyset}} 
            \frac{1}{|I_W \cap p^{-1}(1)|}
            \sum\limits_{i \in I_W \cap p^{-1}(1)}\left(1 - \frac{|W| + \min\{\min\limits_{j \in W} |i - j|, \min\{\min W - \min I_W, \max I_W - \max W\}\} + \min\limits_{j \in W} |i - j|}{|I_W|}\right)
        }{
            |\{W \in A(g) \colon I_W \cap p^{-1}(1) \neq \emptyset\}|
        }
    \end{align*}
    where $I_W = \{i \in [n] \colon \min\limits_{j \in W} |i - \epsilon - j| < \min\limits_{\hat{W} \in A(g) \setminus \{W\}} \min\limits_{j \in \hat{W}} |i - \epsilon - j|\}$
    satisfies none of the properties.
    \\
    \textit{
        Note: We introduce $\epsilon \in (-\frac{1}{2}, \frac{1}{2})$ here to potentially break ties. The original paper \citep{huet2022local} considers the case $\epsilon = 0$, but does not specify how ties are broken. 
        Furthermore, the paper does not specify the case $|\{W \in A(g) \colon I_W \cap p^{-1}(1) \neq \emptyset\}| = 0$. This can happen, if the ground truth or the prediction does not contain a single anomaly window. In that case, the metric as stated is undefined. Thus, we assume $|g^{-1}(1)| > 0$ and $|p^{-1}(1)| > 0$.
    }
\end{theorem}
\begin{proof}
    \item
    Let g$\colon [n] \rightarrow \{0,1\}^n$ be the ground truth and $p,\hat{p} \colon [n] \rightarrow \{0,1\}^n$ predictions.
    
    \paragraph{\cref{prop:detection}} 
    Consider the following example:
    \begin{align*}
        g &= 010 000 001 000\\
        p &= 000 000 001 000\\
        \hat{p}&= 010 000 001 000
    \end{align*}
    Then it holds, $P_{Precision}(g, p) = \frac{30}{35} > \frac{29}{35} = P_{Precision}(g, \hat{p})$.

    \paragraph{\cref{prop:alarms}} 
    Consider the following example:
    \begin{align*}
        g &= 000 111 111 000\\
        p &= 000 111 000 000\\
        \hat{p}&= 000 111 011 000
    \end{align*}
    Then it holds, $P_{Precision}(g, p) = \frac{1}{2} = P_{Precision}(g, \hat{p})$.
    
    \paragraph{\cref{prop:false_positives}}
    Consider the following example:
    \begin{align*}
        g &= 010 000 100 000\\
        p &= 001 100 000 000\\
        \hat{p}&= 001 110 000 000
    \end{align*}
    Then it holds, $P_{Precision}(g, p) = \frac{1}{8} < \frac{1}{4} = P_{Precision}(g, \hat{p})$.
    
    \paragraph{\cref{prop:false_positive_alarms}}
    Consider the following example:
    \begin{align*}
        g &= 010 000 001 000\\
        p &= 000 011 101 000\\
        \hat{p}&= 000 010 101 000
    \end{align*}
    Then it holds, $P_{Precision}(g, p) = \frac{3}{35} < \frac{3}{28} = P_{Precision}(g, \hat{p})$.
    
    \paragraph{\cref{prop:permutations}}
    Consider the following example:
    \begin{align*}
        g &= 000 000 111 000\\
        p &= 010 000 100 000\\
        \hat{p}&= 000 010 100 000
    \end{align*}
    Then it holds, $P_{Precision}(g, p) = \frac{5}{12} < \frac{7}{12} = P_{Precision}(g, \hat{p})$.
    
    \paragraph{\cref{prop:trust}}
    Consider the following example:
    \begin{align*}
        g &= 000 111 111 000\\
        p &= 100 000 001 000\\
        \hat{p}&= 100 111 100 010
    \end{align*}
    Then it holds, $P_{Precision}(g, p) = \frac{1}{4} < \frac{13}{36} = P_{Precision}(g, \hat{p})$.
    
    \paragraph{\cref{prop:prediction_size}}
    Consider the following example:
    \begin{align*}
        g &= 000 111 111 000\\
        p &= 000 011 111 000\\
        \hat{p}&= 000 111 111 000
    \end{align*}
    Then it holds, $P_{Precision}(g, p) = \frac{1}{2} = P_{Precision}(g, \hat{p})$.
    
    \paragraph{\cref{prop:temporal_order}}
    Consider the following example:
    \begin{align*}
        g &= 000 111 111 000\\
        p &= 000 010 000 000\\
        \hat{p}&= 000 000 010 000
    \end{align*}
    Then it holds, $P_{Precision}(g, p) = \frac{1}{2} = P_{Precision}(g, \hat{p})$.
    
    \paragraph{\cref{prop:early_bias}}
    Consider the following example:
    \begin{align*}
        g &= 000 111 111 000\\
        p &= 000 010 010 000\\
        \hat{p}&= 000 110 000 000
    \end{align*}
    Then it holds, $P_{Precision}(g, p) = \frac{1}{2} = P_{Precision}(g, \hat{p})$.
\end{proof}
 
\begin{theorem}\label{proof:p_recall}
    \textbf{Average Recall Probability} defined as 
    \begin{align*}
        &P_{Recall}(g,p) = \\
        &\frac{1}{|A(g)|}
        \sum\limits_{W \in A(g)}
        \frac{1}{|W|}
        \sum\limits_{i \in W}
        \left(
            1 - \frac{\min\{d(i, I_W \cap p^{-1}(1)), \min\{i - \min I_W, \max I_W - i\}\} + d(i, I_W \cap p^{-1}(1))}{|I_W|}
        \right)
    \end{align*}
    where $I_W = \{i \in [n] \colon \min\limits_{j \in W} |i - \epsilon - j| < \min\limits_{\hat{W} \in A(g) \setminus \{W\}} \min\limits_{j \in \hat{W}} |i - \epsilon - j|\}$ and \[
        d(i,I \cap p^{-1}(1)) = \begin{cases}
            \min\limits_{j \in I \cap p^{-1}(1)} |i - j| &, I \cap p^{-1}(1) \neq \emptyset\\
            \infty &, I \cap p^{-1}(1) = \emptyset
        \end{cases}
    \]
    satisfies none of the properties.
    \\
    \textit{
        Note: We introduce $\epsilon \in (-\frac{1}{2}, \frac{1}{2})$ here to potentially break ties. The original paper \citep{huet2022local} considers the case $\epsilon = 0$, but does not specify how ties are broken. 
    }
\end{theorem}
\begin{proof}
    \item
    Let g$\colon [n] \rightarrow \{0,1\}^n$ be the ground truth and $p,\hat{p} \colon [n] \rightarrow \{0,1\}^n$ predictions.

    Consider the following example:
    \begin{align*}
        g &= 000 110 011 000\\
        p &= xxx xxx 000 000\\
    \end{align*}
    The term in the inner sum of $P_{Recall}$ for the second window is equal to $-\infty$ since $I_W \cap p^{-1}(1) = \emptyset$ for any predictions that differ only in the $x$'s.
    Thus, none of the properties except for \cref{prop:permutations} can hold for any two predictions that do not predict one of the ground truth alarms.
    
    \paragraph{\cref{prop:permutations}}
    Consider the following example:
    \begin{align*}
        g &= 000 000 010 000\\
        p &= 000 100 000 000\\
        \hat{p}&= 000 010 000 000
    \end{align*}
    Then it holds $P_{Recall}(g,p) = \frac{1}{3} \neq \frac{1}{2} = P_{Recall}(g,\hat{p})$.
\end{proof}
 
\begin{theorem}\label{proof:f1_affiliation}
    \textbf{Affiliation F1-Score} defined as
    \[
        F1_{affiliation}(g,p) = 2\frac{P_{Precision} \cdot P_{Recall}}{P_{Precision} + P_{Recall}}
    \]
    satisfies none of the properties.
    \\
    \textit{
        Note: $P_{Precision}$ is not defined, when either the ground truth or the prediction are zero.
        Additionally, $P_{Recall}$ equals $-\infty$ if at least one ground truth alarm does not have any affiliated predictions.
        Thus, we assume $|g^{-1}(1)| > 0$, $|p^{-1}(1)| > 0$, and $I_W \cap p^{-1}(1) \neq \emptyset$ for all $W \in A(g)$.
    }
\end{theorem}
\begin{proof}
    \item
    Let g$\colon [n] \rightarrow \{0,1\}^n$ be the ground truth and $p,\hat{p} \colon [n] \rightarrow \{0,1\}^n$ predictions.
    Consider \cref{proof:p_recall}. Similarly here, we can construct counterexamples for all properties since recall is $-\infty$. Affiliation F1-Score is undefined until there is at least one prediction in each window. Thus, none of the properties except for \cref{prop:permutations} can hold for any two predictions that do not predict one of the ground truth alarms.
    \paragraph{\cref{prop:permutations}}
    Consider the following example:
    \begin{align*}
        g &= 000 000 111 000\\
        p &= 010 000 100 000\\
        \hat{p}&= 000 010 100 000
    \end{align*}
    \[
    P_{\text{Precision}}(g, p) = \frac{5}{12} \quad P_{\text{Precision}}(g, \hat{p}) = \frac{7}{12}
    \]
    \[
    P_{\text{Recall}}(g, p) = \frac{1}{6} \quad P_{\text{Recall}}(g, \hat{p}) = \frac{19}{36}
    \]
    Then it holds, $F1_{\text{affiliation}}(g, p) = \frac{5}{21} < \frac{133}{240} = F1_{\text{affiliation}}(g, \hat{p})$.
\end{proof}

\begin{theorem}\label{proof:sTaP}
    \textbf{Enhanced Time-Series Aware Precision} defined as
    \[
        eTaP(g,p) = \frac{1}{2}\sum\limits_{W \in P} 
        \left(
            1 + \sum\limits_{W_g \in A}\frac{|W \cap W_g|}{|W|}
        \right)
        \frac{\sqrt{|W|}}{\sum\limits_{W_p \in A(p)} \sqrt{|W_p|}}
    \]
    where $\theta_r>0$, $\theta_p>0$, and
    \begin{align*}
        A &= \{W \in A(g) \colon \sum\limits_{W_p \in P}\frac{|W \cap W_p|}{|W|} \geq \theta_r\}\\
        P &= \{W \in A(p) \colon \sum\limits_{W_g \in A}\frac{|W \cap W_g|}{|W|} \geq \theta_p\}
    \end{align*}
    satisfies \cref{prop:permutations}.
\end{theorem}
\begin{proof}
    \item
    Let $g\colon [n] \rightarrow \{0,1\}^n$ be the ground truth and $p,\hat{p} \colon [n] \rightarrow \{0,1\}^n$ predictions.
    
    \paragraph{\cref{prop:detection}} 
    Consider the following example:
    Let $\mathbb N\ni a>1/\theta_r$, $g=\mathds{1}_{[1,a]}$, $p=0$, and $\hat{p} = \mathds{1}_{\{1\}}.$
    Then it holds $eTaP(g,p) = 0 = eTaP(g,\hat{p})$.
    
    \paragraph{\cref{prop:alarms}} 
    Consider the following example:
    Let $\mathbb N\ni a>2/\theta_r$, $g=\mathds{1}_{[1,a]}$, $p=\mathds{1}_{\{1\}}$, and $\hat{p} = \mathds{1}_{\{1, a\}}.$
    Then it holds $eTaP(g,p) = 0 = eTaP(g,\hat{p})$.
    
    \paragraph{\cref{prop:false_positives}} 
    Consider the following example:
    \begin{align*}
        g &= 000 000 000 000\\
        p &= 010 000 000 000\\
        \hat{p}&= 011 000 000 000
    \end{align*}
    Then it holds $eTaP(g,p) = 0 = eTaP(g,\hat{p})$.
    
    \paragraph{\cref{prop:false_positive_alarms}}
    Consider the following example:
    \begin{align*}
        g &= 000 000 000 000\\
        p &= 010 000 000 000\\
        \hat{p}&= 010 100 000 000
    \end{align*}
    Then it holds $eTaP(g,p) = 0 = eTaP(g,\hat{p})$.
    
    \paragraph{\cref{prop:permutations}}
    Let $W \in A(|g-1|)$, $\hat{p}_{[n] \setminus W} = p_{[n] \setminus W}$, $|\hat{p}_W^{-1}(1)| = |p_W^{-1}(1)|$, and $|A(\hat{p}_W)| = |A(p_W)|$. Since $\hat{p}_{[n] \setminus W} = p_{[n] \setminus W}$, then $A(\hat{p})=A(p)$, and hence $eTaP(g,\hat{p})=eTaP(g,p)$.
    
    \paragraph{\cref{prop:trust}}
    Consider the following example:
    \begin{align*}
        g &= 010 000 000 000\\
        p &= 100 000 000 000\\
        \hat{p}&= 101 000 000 000
    \end{align*}
    Then it holds $eTaP(g,p) = 0 = eTaP(g,\hat{p})$.
    
    \paragraph{\cref{prop:prediction_size}}
    Consider the following example:
    Let $\mathbb N\ni a>2/\theta_r$, $g=\mathds{1}_{[1,a]}$, $p=\mathds{1}_{\{1\}}$, and $\hat{p} = \mathds{1}_{\{1,2\}}.$
    Then it holds $eTaP(g,p) = 0 = eTaP(g,\hat{p})$.
    
    \paragraph{\cref{prop:temporal_order}}
    Consider the following example:
    Let $\mathbb N\ni a>2/\theta_r$, $g=\mathds{1}_{[1,a]}$, $p=\mathds{1}_{\{1\}}$, and $\hat{p} = \mathds{1}_{\{2\}}.$
    Then it holds $eTaP(g,p) = 0 = eTaP(g,\hat{p})$.
    
    \paragraph{\cref{prop:early_bias}}
    Consider the following example:
    Let $\mathbb N\ni a>2/\theta_r$, $g=\mathds{1}_{[1,a]}$, $p=\mathds{1}_{\{1\}}$, and $\hat{p} = \mathds{1}_{\{2\}}.$
    Then it holds $eTaP(g,p) = 0 = eTaP(g,\hat{p})$.
\end{proof}

\begin{theorem}\label{proof:eTaR}
    \textbf{Enhanced Time-Series Aware Recall} defined as 
    \[
        eTaR(g,p) = \frac{1}{2|A(g)|} \sum\limits_{W \in A}
        \left(
            1 + \sum\limits_{W_p \in P}\frac{|W \cap W_p|}{|W|}
        \right)
    \]
    where $\theta_r>0$, $\theta_p>0$, and
    \begin{align*}
        A &= \{W \in A(g) \colon \sum\limits_{W_p \in P}\frac{|W \cap W_p|}{|W|} \geq \theta_r\}\\
        P &= \{W \in A(p) \colon \sum\limits_{W_g \in A}\frac{|W \cap W_g|}{|W|} \geq \theta_p\}
    \end{align*}
    satisfies \cref{prop:permutations}.
\end{theorem}
\begin{proof}
    \item
    Let $g\colon [n] \rightarrow \{0,1\}^n$ be the ground truth and $p,\hat{p} \colon [n] \rightarrow \{0,1\}^n$ predictions.
    
    \paragraph{\cref{prop:detection}} 
   % Let $W \in A(g)$, $\hat{p}_{[n] \setminus W} = p_{[n] \setminus W}$, $p_W = 0$, and $|\hat{p}_W^{-1}(1)| > 0$.
    Consider the following example:
    Let $\mathbb N\ni a>1/\theta_r$, $g=\mathds{1}_{[1,a]}$, $p=0$, and $\hat{p} = \mathds{1}_{\{1\}}.$
    Then it holds $eTaR(g,p) = 0 = eTaR(g,\hat{p})$.
    
    \paragraph{\cref{prop:alarms}}
    Consider the following example:
    Let $\mathbb N\ni a>2/\theta_r$, $g=\mathds{1}_{[1,a]}$, $p=\mathds{1}_{\{1\}}$, and $\hat{p} = \mathds{1}_{\{1, a\}}.$
    Then it holds $eTaR(g,p) = 0 = eTaR(g,\hat{p})$.
    
    \paragraph{\cref{prop:false_positives}} 
    Consider the following example:
    \begin{align*}
        g &= 100 000 000 000\\
        p &= 010 000 000 000\\
        \hat{p}&= 011 000 000 000
    \end{align*}
    Then it holds $eTaR(g,p) = 0 = eTaR(g,\hat{p})$.
    
    \paragraph{\cref{prop:false_positive_alarms}}
    Consider the following example:
    \begin{align*}
        g &= 100 000 000 000\\
        p &= 010 000 000 000\\
        \hat{p}&= 010 100 000 000
    \end{align*}
    Then it holds $eTaR(g,p) = 0 = eTaR(g,\hat{p})$.
    
    \paragraph{\cref{prop:permutations}}
    Let $W \in A(|g-1|)$, $\hat{p}_{[n] \setminus W} = p_{[n] \setminus W}$, $|\hat{p}_W^{-1}(1)| = |p_W^{-1}(1)|$, and $|A(\hat{p}_W)| = |A(p_W)|$. Since $\hat{p}_{[n] \setminus W} = p_{[n] \setminus W}$, then $A(\hat{p})=A(p)$, and hence $eTaR(g,\hat{p})=eTaR(g,p)$.
    
    \paragraph{\cref{prop:trust}}
    Consider the following example:
    \begin{align*}
        g &= 100 000 000 000\\
        p &= 100 000 000 000\\
        \hat{p}&= 101 000 000 000
    \end{align*}
    Then it holds $eTaR(g,p) = 1 = eTaR(g,\hat{p})$.
    
    \paragraph{\cref{prop:prediction_size}}
    Consider the following example:
    Let $\mathbb N\ni a>2/\theta_r$, $g=\mathds{1}_{[1,a]}$, $p=\mathds{1}_{\{1\}}$, and $\hat{p} = \mathds{1}_{\{1,2\}}.$
    Then it holds $eTaR(g,p) = 0 = eTaR(g,\hat{p})$.

    \paragraph{\cref{prop:temporal_order}}
    Consider the following example:
    Let $\mathbb N\ni a>2/\theta_r$, $g=\mathds{1}_{[1,a]}$, $p=\mathds{1}_{\{1\}}$, and $\hat{p} = \mathds{1}_{\{2\}}.$
    Then it holds $eTaR(g,p) = 0 = eTaR(g,\hat{p})$.

    \paragraph{\cref{prop:early_bias}}
    Consider the following example:
    Let $\mathbb N\ni a>2/\theta_r$, $g=\mathds{1}_{[1,a]}$, $p=\mathds{1}_{\{1\}}$, and $\hat{p} = \mathds{1}_{\{2\}}.$
    Then it holds $eTaR(g,p) = 0 = eTaR(g,\hat{p})$.
\end{proof}

\begin{theorem}\label{proof:temporal_distance}
    \textbf{Temporal Distance} is defined as
    \[TD(g, p) = f_{\text{closest}}(g, p) + f_{\text{closest}}(p, g)\]
    \[f_{\text{closest}}(a, b) = \sum_{i \in a^{-1}(1)} \min_{j \in b^{-1}(1)} |i - j|.\]
    A Temporal Distance-based metric $m_{TD}(g, p) = f(TD(g, p))$, where $f$ is a strictly decreasing function, satisfies properties \cref{prop:detection}, \cref{prop:prediction_size}.
\end{theorem}

\begin{proof}
    \item
    Let g$\colon [n] \rightarrow \{0,1\}^n$ be the ground truth and $p,\hat{p} \colon [n] \rightarrow \{0,1\}^n$ predictions. 

    \begin{lemma}\label{proof:temporal_distance_lemma}
    Let $W \in A(g)$, $\hat{p}_{[n] \setminus \{i^*\}} = p_{[n] \setminus \{i^*\}}$ for some $i^* \in W$, and $\hat{p}(i^*) \neq p(i^*) = 0$. Then $m_{TD}(g, \hat{p}) > m_{TD}(g, p)$.
    \end{lemma}
    
    \begin{proof}
    
    Consider the two $f_{closest}$ terms of TD separately:

    \begin{enumerate}
        \item Since \( i^* \in \hat{p}^{-1}(1) \), we have \( \min_{j \in \hat{p}^{-1}(1)} |i^* - j| = 0 \). For any $i' \in g^{-1}(1) \setminus \{i^*\}$ if \( |i' - i^*| < \min_{j \in p^{-1}(1)} |i' - j| \), then 
        \[
        \min_{j \in \hat{p}^{-1}(1)} |i' - j| = |i' - i^*| < \min_{j \in p^{-1}(1)} |i' - j|.
        \]
        Thus: \[f_{\text{closest}}(g, \hat{p}) < f_{\text{closest}}(g, p)\]

        \item Note, that for $I\subset[n]$, the following holds: \[f_{\text{closest}}(a, b) = f_{\text{closest}}(a_I, b) + f_{\text{closest}}(a_{[n] \setminus I}, b).\]
        Since \( i^* \in g^{-1}(1) \), we have \( \min_{j \in g^{-1}(1)} |i^* - j| = 0 \), thus: \[f_{\text{closest}}(\hat{p}_W, g) = f_{\text{closest}}(p_W, g) = 0 \implies f_{\text{closest}}(p, g) = f_{\text{closest}}(\hat{p}, g)\]
    \end{enumerate}
    Then:
    \[TD(g, \hat{p}) < TD(g, p) \implies m_{TD}(g, \hat{p}) > m_{TD}(g, p).\]
    \end{proof}
    
    \paragraph{\cref{prop:detection}} 
    Let $W \in A(g)$, $\hat{p}_{[n] \setminus W} = p_{[n] \setminus W}$, $p_W = 0$, and $|\hat{p}_W^{-1}(1)| > 0$. 
    By \cref{proof:temporal_distance_lemma} , it directly follows that $m_{TD}(g, \hat{p}) > m_{TD}(g, p)$.

    \paragraph{\cref{prop:alarms}} 
    Consider the following example:
    \begin{align*}
        g &= 000 111 111 000\\
        p &= 000 111 000 000\\
        \hat{p}&= 000 111 011 000
    \end{align*}
    Then it holds $TD(g,p)=6>1=TD(g,\hat{p})\implies m_{TD}(g, p)<m_{TD}(g, \hat{p})$.
    
    \paragraph{\cref{prop:false_positives}} 
    Consider the following example:

    \begin{align*}
        g &= 000 000 011 000\\
        p &= 000 010 000 000\\
        \hat{p}&= 000 011 000 000
    \end{align*}
    Then it holds $TD(g,p)=10=TD(g,\hat{p}) \implies m_{TD}(g, p)=m_{TD}(g, \hat{p})$.

    \paragraph{\cref{prop:false_positive_alarms}}
    Consider the following example:
    \begin{align*}
        g &= 000 000 111 000\\
        p &= 001 110 110 000\\
        \hat{p}&= 001 010 110 000
    \end{align*}
    Then it holds $TD(g,p)=10>7=TD(g,\hat{p}) \implies m_{TD}(g, p)<m_{TD}(g, \hat{p})$.
    
    \paragraph{\cref{prop:permutations}}
    Consider the following example:
    \begin{align*}
        g &= 000 000 011 110\\
        p &= 011 001 011 000\\
        \hat{p}&= 001 011 011 000
    \end{align*}
    Then it holds $TD(g,p)=16>13=TD(g,\hat{p}) \implies m_{TD}(g, p)<m_{TD}(g, \hat{p})$.
    
    \paragraph{\cref{prop:trust}}
    \begin{align*}
        g &= 000 011 110 000\\
        p &= 000 010 000 000\\
        \hat{p}&= 001 011 110 000
    \end{align*}
    Then it holds $TD(g,p)=6>2=TD(g,\hat{p}) \implies m_{TD}(g, p)<m_{TD}(g, \hat{p})$.

    \paragraph{\cref{prop:prediction_size}}
    Let $W \in A(g)$, $\hat{p}_{[n] \setminus \{i^*\}} = p_{[n] \setminus \{i^*\}}$ for some $i^* \in W$, $\hat{p}(i^*) \neq p(i^*) = 0$, and $|A(\hat{p}_W)| \leq |A(p_W)|$. By \cref{proof:temporal_distance_lemma} , it directly follows that $m_{TD}(g, \hat{p}) > m_{TD}(g, p)$.
    
    \paragraph{\cref{prop:temporal_order}}
    \begin{align*}
        g &= 000 011 111 000\\
        p &= 000 011 000 000\\
        \hat{p}&= 000 000 011 000
    \end{align*}
    Then it holds $TD(g,p)=TD(g,\hat{p})=6 \implies m_{TD}(g, p)=m_{TD}(g, \hat{p})$.
    
    \paragraph{\cref{prop:early_bias}}
    Consider the following example:
    \begin{align*}
        g &= 000 111 111 000\\
        p &= 000 010 010 000\\
        \hat{p}&= 000 011 000 000
    \end{align*}
    Then it holds $TD(g,p)=4<7=TD(g,\hat{p}) \implies m_{TD}(g, p)>m_{TD}(g, \hat{p})$.
\end{proof}
 
\begin{theorem}\label{proof:average_alert_delay}
    \textbf{Average Alert Delay} is defined as \[
        AAD(g,p) = \frac{1}{|\{ W \in A(g) \colon W \cap p^{-1}(1) \neq \emptyset\}|}\sum\limits_{\stackrel{W \in A(g)\colon}{W \cap p^{-1}(1) \neq \emptyset}} \min_{i\in W \colon p(i) = 1} i - \min_{i\in W} i.
    \]
     An Average Alert Delay-based metric $m_{AAD}(g, p) = f(AAD(g, p))$, where $f$ is a strictly decreasing function, satisfies \cref{prop:permutations} and \cref{prop:temporal_order}.
\end{theorem}

\begin{proof}
    \item
    Let $\colon [n] \rightarrow \{0,1\}^n$ be the ground truth and $p,\hat{p} \colon [n] \rightarrow \{0,1\}^n$ predictions.
    
    \paragraph{\cref{prop:detection}} 
    Consider the following example:
    \begin{align*}
        g &= 000 110 011 000\\
        p &= 000 110 010 000\\
        \hat{p}&= 000 110 000 000
    \end{align*}
    Then it holds $AAD(g,p) = 0  = AAD(g,\hat{p}) \implies m_{AAD}(g, p) = m_{AAD}(g, \hat{p})$.
    
    \paragraph{\cref{prop:alarms}} 
    Consider the following example:
    \begin{align*}
        g &= 000 111 111 000\\
        p &= 000 010 000 000\\
        \hat{p}&= 000 010 010 000
    \end{align*}
    Then it holds $AAD(g,p) = 1  = AAD(g,\hat{p}) \implies m_{AAD}(g, p) = m_{AAD}(g, \hat{p})$.
    
    \paragraph{\cref{prop:false_positives}} 
    Consider the following example:
    \begin{align*}
        g &= 000 000 111 000\\
        p &= 000 010 010 000\\
        \hat{p}&= 000 110 010 000
    \end{align*}
    Then it holds $AAD(g,p) = 1  = AAD(g,\hat{p}) \implies m_{AAD}(g, p) = m_{AAD}(g, \hat{p})$.
    
    \paragraph{\cref{prop:false_positive_alarms}}
    Consider the following example:
    \begin{align*}
        g &= 000 000 111 000\\
        p &= 010 010 010 000\\
        \hat{p}&= 000 010 010 000
    \end{align*}
    Then it holds $AAD(g,p) = 1 = AAD(g,\hat{p}) \implies m_{AAD}(g, p) = m_{AAD}(g, \hat{p})$.
    
    \paragraph{\cref{prop:permutations}}
    Let $W \subset W' \in A(|g-1|)$, $\hat{p}_{[n] \setminus W} = p_{[n] \setminus W}$, $|\hat{p}_W^{-1}(1)| = |p_W^{-1}(1)|$, and $|A(\hat{p}_W)| = |A(p_W)|$.
    Then $\{ W \in A(g) \colon W \cap \hat{p}^{-1}(1) \neq \emptyset\} = \{ W \in A(g) \colon W \cap p^{-1}(1) \neq \emptyset\}$.
    Thus, $AAD(g,p) = AAD(g,\hat{p}) \implies m_{AAD}(g, p) = m_{AAD}(g, \hat{p})$.
    
    \paragraph{\cref{prop:trust}}
    Consider the following example:
    \begin{align*}
        g &= 000 000 111 000\\
        p &= 000 000 010 000\\
        \hat{p}&= 000 010 010 000
    \end{align*}
    Then it holds $AAD(g,p) = 1 = AAD(g,\hat{p})\implies m_{AAD}(g, p) = m_{AAD}(g, \hat{p})$.
    
    \paragraph{\cref{prop:prediction_size}}
    Consider the following example:
    \begin{align*}
        g &= 000 111 111 000\\
        p &= 000 010 000 000\\
        \hat{p}&= 000 011 000 000
    \end{align*}
    Then it holds $AAD(g,p) = 1 = AAD(g,\hat{p}) \implies m_{AAD}(g, p) = m_{AAD}(g, \hat{p})$.
    
    \paragraph{\cref{prop:temporal_order}}
    Let $W \in A(g)$, $\hat{p}_{[n] \setminus W} = p_{[n] \setminus W}$, $|A(\hat{p}_W)| = |A(p_W)|$, $|\hat{p}^{-1}(1)| = |p^{-1}(1)|$, and $\min\limits_{i \in W\colon p(i) = 1} i < \min\limits_{i \in W\colon \hat{p}(i) = 1} i$. Then $\min\limits_{i\in W \colon p(i) = 1} i - \min\limits_{i\in W} i < \min\limits_{i\in W \colon \hat{p}(i) = 1} i - \min\limits_{i\in W} i$, and hence $AAD(g,p) < AAD(g,\hat{p}) \implies m_{AAD}(g, p) > m_{AAD}(g, \hat{p})$.
    % Consider the following example:
    % \begin{align*}
    %     g &= 000 111 111 000\\
    %     p &= 000 010 000 000\\
    %     \hat{p}&= 000 001 000 000
    % \end{align*}
    % Then it holds $AAD(g,p) = 1 < 2 = AAD(g,\hat{p})$.
    
    \paragraph{\cref{prop:early_bias}}
    Consider the following example:
    \begin{align*}
        g &= 000 111 111 000\\
        p &= 000 010 010 000\\
        \hat{p}&= 000 011 000 000
    \end{align*}
    Then it holds $AAD(g,p) = 1 = AAD(g,\hat{p}) \implies m_{AAD}(g, p) = m_{AAD}(g, \hat{p})$.
\end{proof}

\end{document}